\documentclass[a4]{article}
\usepackage[a4paper,left=0.5in,right=0.5in,top=0.5in,bottom=0.5in,%
footskip=.25in]{geometry}
\usepackage{titling}
\usepackage[T1]{fontenc}
\usepackage{helvet}
\usepackage{chemmacros}
\usepackage{makecell}
\usepackage[percent]{overpic}
\usepackage{tikz}
\usepackage[normalem]{ulem}
\usetikzlibrary{shapes.geometric, arrows, calc, decorations.pathreplacing}
\usepackage[hyphens]{url}
\usepackage{graphicx}
\usepackage{subfiles}
\usepackage{caption}
\usepackage[superscript]{cite}
\usepackage{xspace}
\usepackage{enumitem}%
\usepackage[justification=centering]{caption}
\usepackage{subcaption}
\usepackage{graphicx}
\usepackage{stfloats}
\usepackage[hang,flushmargin, norule]{footmisc}
\usepackage{marvosym}
\usepackage{nameref}
\usepackage{seqsplit}
\usepackage[hidelinks]{hyperref}
\usepackage[dvipsnames]{xcolor}
\PassOptionsToPackage{table}{xcolor}
\usepackage{xcolor}
\usepackage{longtable}
\usepackage{booktabs}
\usepackage{array}
\usepackage{titlesec}
\usepackage{multicol}
\usepackage{cite}
\usepackage{amsmath,amssymb,amsfonts}
\usepackage{algcompatible}
\usepackage{algorithm,algpseudocode}
\usepackage{graphicx}
\usepackage{booktabs, multirow, soul}
\usepackage[flushleft]{threeparttable}
\usepackage{xspace, mathtools, comment}
\usepackage{caption}
\usepackage{arydshln}
\usepackage[resetlabels,labeled]{multibib}
\usepackage{etoolbox}
\usepackage{multicol}
\usepackage{lipsum}
\usepackage{textcase}
\usepackage{collcell}
\usepackage{lineno}

\usepackage{subfiles}
\usepackage{hyperref}
\usepackage{xr-hyper}

\titlespacing*{\section}{0pt}{1\baselineskip}{0.1\baselineskip}
\titlespacing*{\subsection}{0pt}{1\baselineskip}{0.1\baselineskip}
\titlespacing*{\subsubsection}{0pt}{1\baselineskip}{0.1\baselineskip}

\titleformat*{\section}{\normalfont\sffamily\large\bfseries}
\titleformat*{\subsection}{\normalfont\sffamily\normalsize\bfseries}
\titleformat*{\subsubsection}{\normalfont\sffamily\small\bfseries}

\algnewcommand{\LineComment}[1]{\Statex \(\triangleright\) #1}
\algnewcommand{\ShortLineComment}[1]{\Statex \hspace{1.8em}\(\triangleright\) #1}
\algnewcommand{\ShortShortLineComment}[1]{\Statex \hspace{3.1em}\(\triangleright\) #1}

\DeclareCaptionLabelSeparator{bar}{\space\textbar\space}

\makeatletter
\renewcommand\@biblabel[1]{#1.}
\makeatother

\makeatletter
\renewenvironment{table*}%
{\renewcommand\familydefault\sfdefault
	\@float{table}}
{\end@float}
\makeatother

\makeatletter
\renewcommand{\scriptsize}{\@setfontsize\scriptsize{7}{7}}
\renewcommand{\footnotesize}{\@setfontsize\footnotesize{8}{9}}
\makeatother

\patchcmd{\thebibliography}
{\settowidth}
{\setlength{\parsep}{0pt}\setlength{\itemsep}{0pt plus 0pt}\settowidth}
{}{}
\apptocmd{\thebibliography}
{\small}
{}{}

\renewcommand{\footnoterule}{%
	\kern -3pt
	\hrule width 0.5\textwidth height 0.3pt
	\kern 2pt
}

\makeatletter
\newcommand\footnoteref[1]{\protected@xdef\@thefnmark{\ref{#1}}\@footnotemark}
\makeatother

\newcommand{\etal}{\mbox{\emph{et al.}}\xspace}

\newcommand{\grey}[1]{\textcolor{gray}{#1}}

\newcommand{\method}
{\mbox{$\mathop{\mathsf{conDitar}}\limits$}\xspace}
\newcommand{\methodopt}{\mbox{$\mathop{\method\text{-}{\mathsf{dev}}}\limits$}\xspace}
\newcommand{\opt}{\mbox{$\mathop{\mathsf{paOPT}}\limits$}\xspace}

\newcommand{\compound}[1]{compound~#1\xspace}

\newcommand{\cmpTen}{\compound{{CL-10}}}
\newcommand{\cmpEleven}{\compound{{CL-11}}}

\newcommand{\cmpFifteen}{\compound{{PL-2}}}
\newcommand{\cmpSixteen}{\compound{{PL-1}}}
\newcommand{\cmpSeventeen}{\compound{{PL-4}}}
\newcommand{\cmpEighteen}{\compound{{PL-3}}}

\newcommand{\CmpFifteen}{\compound{{PL-2}}}
\newcommand{\CmpSixteen}{\compound{{PL-1}}}

\newcommand{\CmpEighteen}{\compound{{PL-3}}}

\newcommand{\methodforward}{\mbox{$\mathop{\mathsf{conDitar}\text{-}\mathsf{forward}}\limits$}\xspace}
\newcommand{\methodbackward}{\mbox{$\mathop{\mathsf{conDitar}\text{-}\mathsf{reverse}}\limits$}\xspace}
\newcommand{\zograd}{\mbox{$\mathop{\mathsf{ZOGrad}}\limits$}\xspace}
\newcommand{\noiseopt}{\mbox{$\mathop{\mathsf{NoiseOpt}}\limits$}\xspace}
\newcommand{\pocketenc}{\mbox{$\mathop{\mathsf{\PRL}\text{-}\mathsf{enc}}\limits$}\xspace}
\newcommand{\pocketdec}{\mbox{$\mathop{\mathsf{\PRL}\text{-}\mathsf{dec}}\limits$}\xspace}

\newcommand{\predictor}{\mbox{$\mathop{\mathsf{pcLG}}\limits$}\xspace}
\newcommand{\diff}{\mbox{$\mathop{\predictor}\limits$}\xspace}

\newcommand{\PRL}{\mbox{$\mathop{\mathsf{msPRL}}\limits$}\xspace}

\newcommand{\pocketset}{\mbox{$\mathop{\mathcal{P}}\limits$}\xspace}
\newcommand{\pocketatomset}{\mbox{$\mathop{{A}_p}\limits$}\xspace}
\newcommand{\pocketresidueset}{\mbox{$\mathop{{R}_p}\limits$}\xspace}

\newcommand{\pocketatom}{\mathsf{a}^p}
\newcommand{\pocketatompos}{\mathbf{x}^p}
\newcommand{\pocketatomtype}{\mathbf{v}^p}

\newcommand{\mol}{\mbox{$\mathop{\mathcal{D}}\limits$}\xspace}

\newcommand{\molneighbour}{\mbox{$\mathop{\mathbf{N}}\limits$}\xspace}
\newcommand{\pocketneighbour}{\mbox{$\mathop{\mathbf{N}}\limits$}\xspace}

\newcommand{\interactionneighbour}{\mbox{$\mathop{\mathbf{N}}\limits$}\xspace}

\newcommand{\molatom}{\mathsf{a}^g}
\newcommand{\molatompos}{\mathbf{x}^g}
\newcommand{\molatomtype}{\mathbf{v}^g}

\newcommand{\molpocketatom}{\mathsf{a}}
\newcommand{\molpocketatompos}{\mathbf{x}}
\newcommand{\molpocketatomtype}{\mathbf{v}}

\newcommand{\residue}{\mathsf{r}^p}
\newcommand{\residuepos}{\mathbf{x}^r}
\newcommand{\residuetype}{\mathbf{v}^r}

\newcommand{\pascalar}{\mathbf{s}^p}
\newcommand{\pavec}{\mathcal{H}^p}

\newcommand{\pamolscalar}{\mathbf{s}}
\newcommand{\pamolvec}{\mathcal{H}}

\newcommand{\prscalar}{\mathbf{s}^r}
\newcommand{\prvec}{\mathcal{H}^r}

\newcommand{\pcscalar}{\mathbf{s}^{c}}
\newcommand{\pcvec}{\mathcal{H}^{c}}

\newcommand{\rcscalar}{\mathbf{s}^{a}}
\newcommand{\rcvec}{\mathcal{H}^{a}}

\newcommand{\messscalar}{\mathbf{m}^p}
\newcommand{\messvec}{M^p}
\newcommand{\messatomscalar}{\mathbf{m}^g}
\newcommand{\messatomvec}{M^g}

\newcommand{\messatomscalarhat}{\hat{\mathbf{m}}^g}
\newcommand{\messatomvechat}{\hat{M}^g}

\newcommand{\cumalpha}{\mbox{$\mathop{\bar{\alpha}}\limits$}\xspace}

\newcommand{\atomvec}{\mathcal{H}^g}
\newcommand{\atomscalar}{\mathbf{s}^g}

\newcommand{\bondpred}{\mbox{$\mathop{\mathbf{e}}\limits$}\xspace}
\newcommand{\bondtype}{\mbox{$\mathop{\mathbf{b}}\limits$}\xspace}

\NewDocumentCommand{\distg}{m m}{
  \ensuremath{d(\mathsf{a}^g_{#1}, \mathsf{a}^g_{#2})}
}

\NewDocumentCommand{\distp}{m m}{
  \ensuremath{d(\mathsf{a}^p_{#1}, \mathsf{a}^p_{#2})}
}

\NewDocumentCommand{\dist}{m m}{
  \ensuremath{d(\mathsf{a}^g_{#1}, \mathsf{a}^p_{#2})}
}

\NewDocumentCommand{\distnotation}{m m}{
  \ensuremath{d(\mathsf{a}_{#1}, \mathsf{a}_{#2})}
}

\newcommand{\xtilg}{\tilde{\mathbf{x}}^g\xspace}
\newcommand{\vtilg}{\tilde{\mathbf{x}}^g\xspace}

\newcommand{\noisepos}{\mathbf{z}^{\mathtt{x}}}
\newcommand{\noisetype}{\mathbf{z}^{\mathtt{v}}}

\newcommand{\pockettwomol}{\mbox{$\mathop{\mathsf{Pocket2Mol}}\limits$}\xspace}
\newcommand{\targetdiff}{\mbox{$\mathop{\mathsf{TargetDiff}}\limits$}\xspace}
\newcommand{\ipdiff}{\mbox{$\mathop{\mathsf{IPDiff}}\limits$}\xspace}
\newcommand{\alidiff}{\mbox{$\mathop{\mathsf{ALIDiff}}\limits$}\xspace}
\newcommand{\decompdiff}
{\mbox{$\mathop{\mathsf{DecompDiff}}\limits$}\xspace}

\newcommand{\decompopt}
{\mbox{$\mathop{\mathsf{DecompOpt}}\limits$}\xspace}

\newcommand{\AR}{\mbox{$\mathop{\mathsf{AR}}\limits$}\xspace}

\newcommand{\diffsbdd}{\mbox{$\mathop{\mathsf{DiffSBDD}}\limits$}\xspace}

\newcommand{\flowr}{\textsc{FLOWR}}

\newcommand{\DatasetsSection}{{Datasets}\xspace}

\newcommand{\PESection}{{Pocket Encoder (\pocketenc)}\xspace}

\newcommand{\DiffSubsection}{{Diffusion Process}\xspace}

\newcommand{\PredictorSubsection}{{Pocket-conditioned Ligand Generator (\predictor)}\xspace}

\newcommand{\TrainSubsection}{{Model Training}\xspace}

\newcommand{\DistributionalAnalysisSection}{{Analysis of Atom and Bond Composition on \dataset}\xspace}

\newcommand{\dataset}{\mbox{$\mathop{\mathsf{CDH}}\limits$}\xspace}

\newcommand{\CD}{\mbox{$\mathop{\mathsf{CD}}\limits$}\xspace}

\newcommand{\PDLone}{{PD-L1}\xspace}

\newcommand{\CSFoneR}{CSF1R\xspace}

\title{Generating Developable 3D Molecules via Pocket-Conditioned Diffusion and Property-Aware Optimization}
\date{\vspace{-5ex}}

 \author{
 	Ruoxi Gao\textsuperscript{\rm 1}, 
 	Jiangweizhi Peng\textsuperscript{\rm 2}, 
    Ziqi Chen\textsuperscript{\rm 3}\thanks{Work done while at The Ohio State University},
 	Frazier N. Baker\textsuperscript{\rm 1},
    David C. Kombo\textsuperscript{\rm 4},
    John L. Kane, Jr.\textsuperscript{\rm 4},  \\
    Andrew A. Scholte\textsuperscript{\rm 4}, 
    Yi Li\textsuperscript{\rm 4},
    Matthew J. LaMarche\textsuperscript{\rm 4},
    Luigi I. Iconaru\textsuperscript{\rm 5}, 
    Hans-Peter Biemann\textsuperscript{\rm 5}, \\
	Mingyi Hong\textsuperscript{\rm 6},
 	Xia Ning\textsuperscript{\rm 1,7,8,9 \Letter}
 }
 \newcommand{\Address}{
    \centering
 	\textsuperscript{\rm 1}Computer Science and Engineering, The Ohio State University, Columbus, OH 43210.\\
 	\textsuperscript{\rm 2}Industrial and System Engineering, University of Minnesota, Minneapolis, MN 55455.\\
    \textsuperscript{\rm 3}Google Cloud, Google LLC, Mountain View, CA 94043. \\
    \textsuperscript{\rm 4}Medicinal Chemistry Department, Integrated Drug Discovery, Sanofi, Cambridge, MA 02141. \\
    \textsuperscript{\rm 5}In-Vitro Biology Department, Integrated Drug Discovery, Sanofi, Cambridge, MA 02141. \\
    \textsuperscript{\rm 6}Electrical and Computer Engineering, University of Minnesota, Minneapolis, MN 55455. \\
 	\textsuperscript{\rm 7}Biomedical Informatics, The Ohio State University, Columbus, OH 43210.\\
 	\textsuperscript{\rm 8}Medicinal Chemistry and Pharmacognosy, The Ohio State University, Columbus, OH 43210.\\
 	\textsuperscript{\rm 9}Translational Data Analytics Institute, The Ohio State University, Columbus, OH 43210. \\
 	\textsuperscript{\Letter}ning.104@osu.edu
 }

\newcommand{\myabstract}[2][1]{%
	\renewcommand\maketitlehookd{
		\Address
		\vspace{10pt}
		\mbox{}\medskip\par
		\centering
		\begin{minipage}{#1\textwidth}
			\textbf{
			{\fontfamily{phv}\selectfont
				#2
				}
			}
		\end{minipage}
	}
}

\usepackage{bibentry}

\begin{document}

\myabstract[]{Drug discovery and development is complex,
time-consuming, and resource-intensive,
motivating the use of
computational methods, such as generative diffusion models,
for \emph{de novo} drug design.
Many of these generative models follow the structure-based drug design (SBDD) paradigm using diffusion, 
where molecules are designed to fit a target pocket.
However, existing diffusion-based SBDD methods typically
couple pocket and ligand representation learning,
model interactions on a single, atom-level scale,
and prioritize binding affinity while deferring other developability properties.
Here, we introduce \methodopt, a {c}onditional {d}iffusion-based {SBDD} framework 
capable of generating ligands with strong binding affinities and favorable ADMET properties. 
\methodopt consists of three modules: 
\PRL, a pretrained multi-scale pocket representation learning module 
that encodes the binding pocket
into expressive representations;
\method, a pocket-conditioned diffusion model for molecule generation guided by \PRL representations and a multi-scale ligand generator \predictor; and
\opt, a training-free, generation-time method for optimizing \method's ligand generation for developability properties.
For predicted binding affinity, \method outperforms the state-of-the-art SBDD baselines, generating ligands with an
average score of -8.85 kcal/mol and a 7.4\% improvement over the best baseline on a newly-curated benchmark of human disease targets.
Across
five ADMET properties, \methodopt delivers improvements of up to 73\% over \method while retaining similar predicted binding affinity.
To further validate 
the abilities of \method and \methodopt to generate developable molecules, 
we have applied them 
to design ligands for two validated druggable targets: the programmed death-ligand 1 (\PDLone), and colony-stimulating factor-1 receptor (\CSFoneR) proteins. 
Top-ranked generatively designed molecules and their analogs have been 
experimentally synthesized and biologically tested.
Two molecules generated directly by \methodopt for \PDLone exhibited SPR-derived
K$_D$ of 3.49 and 3.75 $\mu$M, respectively. 
Hit expansion based on \methodopt-designed molecules identified selective \CSFoneR inhibitors
with IC$_{50}$ values as low as 200 nM, while also uncovering opportunities for 
drug repositioning. 
}

\maketitle

\section*{Introduction}

Drug discovery and development is complex, time-consuming, and resource-intensive -- new drugs typically take 12-15 years 
to develop~\cite{singh2023drug} at costs of \$378 million to \$1.76 billion~\cite{sertkaya2024costs}.
To accelerate this process and reduce costs, computational methods,
particularly recent generative models (e.g., diffusion~\cite{hoogeboom22diff,guan2023decompdiff}, variational autoencoders~\cite{jin18jtvae}, large language models~\cite{Yu2024}),
have been employed for \emph{de novo} drug design.
Rather than conducting expensive searches over large libraries to identify potential binding molecules, 
these models directly generate molecules \emph{in silico} based on the chemical knowledge learned 
from a vast amount of experimental data.
These models generally follow one of the two conventional drug design frameworks:
\textbf{(1)} structure-based drug design (SBDD), where molecules are designed to fit into the binding pocket 
structure of a given target; and
\textbf{(2)} ligand-based drug design (LBDD), where molecules are designed to resemble a known binding ligand.
Compared to LBDD, SBDD offers a more direct and mechanistic entry point to molecular discovery 
by explicitly exploiting pocket-ligand interactions, thereby providing a principled starting point for 
subsequent ligand-based modeling and optimization.
Recent development of generative models for SBDD~\cite{chen2025generating, guan2023decompdiff, guan2023targetdiff} features diffusion~\cite{ho2020ddpm}, 
a process that gradually transforms noise into molecular structures, as the leading paradigm. 
By learning the underlying distributions of chemically and structurally feasible molecules from data 
and leveraging the distributions to generate novel molecules, 
diffusion holds substantial promise to revolutionize \emph{de novo} drug design 
by rapidly producing high-quality, functionally relevant molecular candidates.
Still, existing diffusion-based models for SBDD leave substantial room for improvement.
Typically, diffusion-based SBDD models tend to couple pocket and ligand representation learning, 
using a shared representation to reflect the complex pocket-ligand interactions.
However, such coupling could amplify early generation errors, leading to progressively degraded 
representations during diffusion's iterative refinement process. 
Instead, dedicated pocket representation learning may better capture  
binding pocket structures and properties, and thus, ligand generation can be conditioned 
on robust pocket features that favor strong binding.
Furthermore, many existing models do not yet fully leverage interaction signals across multiple scales 
-- ranging from atom-level to residue-level -- in modeling pocket-ligand interactions.
Incorporating rich, multi-scale interaction signals could allow models to better represent 
the biochemical environment of the binding site, capturing both local atomic interactions 
and broader residue-level context that jointly govern ligand recognition and binding.
In addition, generative SBDD models typically prioritize the generation of ligands with optimized binding affinity, 
leaving other critical drug properties, such as Absorption, Distribution, Metabolism, Excretion, and Toxicity (ADMET), 
to be addressed in downstream lead optimization phase. 
While this aligns with the conventional drug development process~\cite{hughes2011principles}, 
it also underscores the opportunity for a more proactive strategy that considers a broader set of 
drug-relevant properties at the stage of initial molecule generation, thereby increasing the likelihood of 
ultimate success.
These insights motivate the design of a new, decoupled, multi-scale, and developability-aware diffusion-based 
framework for SBDD.

\begin{figure*}[htbp]
\centering
 \includegraphics[width=\textwidth]{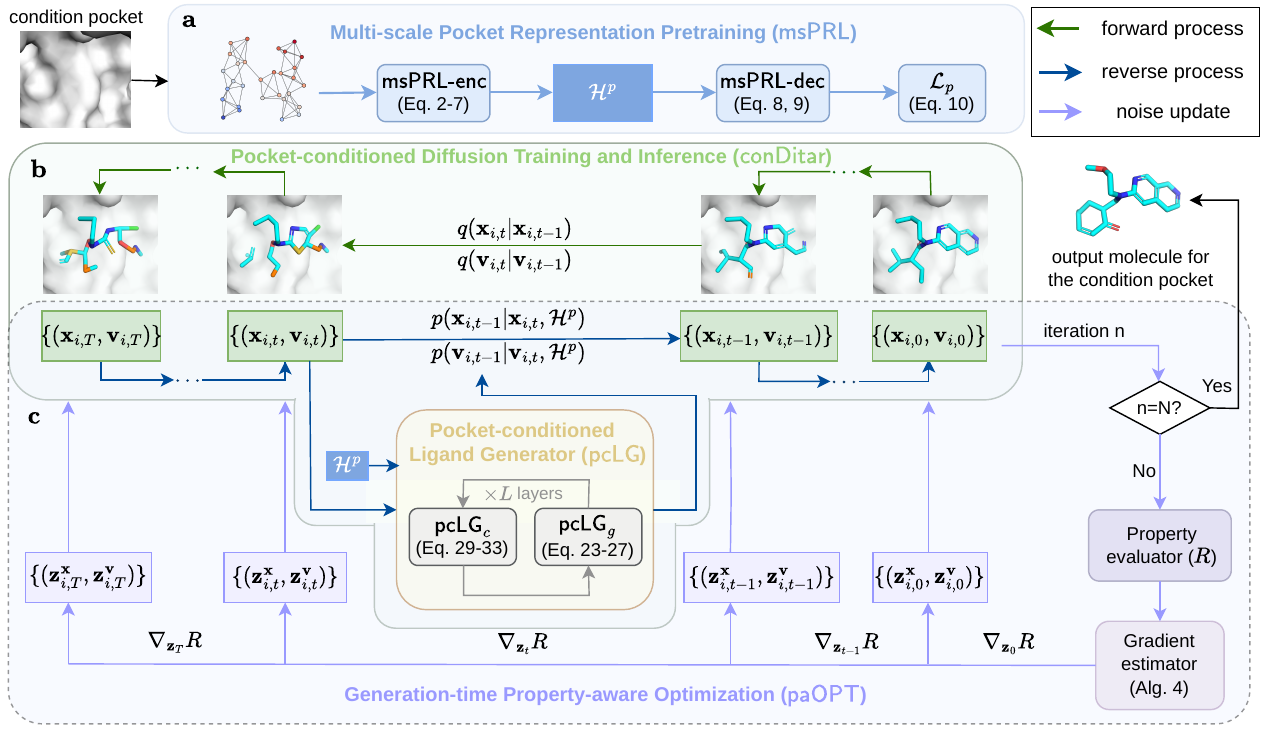}
\caption{The overall schematic diagram of \methodopt.
\textbf{a,} Pocket representation pretraining, \PRL. A pocket representation learning module is pretrained to encode protein pockets into multi-scale atom-level and residue-level features.
\textbf{b,} \method model training and inference. Conditioned on these fixed pocket features, \method generates 3D ligands for target pockets through a pocket-conditioned diffusion model.
\textbf{c,} Generation-time molecule optimization, \opt. During inference, \methodopt optimizes the original trajectory in a training-free manner to steer generated molecules toward improved ADMET properties.}
\label{fig:model_architecture}
\end{figure*}

Here, we introduce \methodopt, a {c}onditional {d}iffusion-based {SBDD} framework 
capable of generating ligands with strong binding affinities and favorable {ADMET properties}. 
\methodopt consists of three modules: 
\textbf{(1)} a pretrained \ul{m}ulti-\ul{s}cale \ul{p}ocket \ul{r}epresentation \ul{l}earning module, referred to as \PRL, 
which encodes the binding pocket's atomic and residue-level composition and structure
into expressive representations that will guide ligand generation;
\textbf{(2)} a pocket-\ul{con}ditioned \ul{di}ffusion model that generates ligands for strong \ul{tar}get binding, 
referred to as \method; and
\textbf{(3)}  a generation-time, \ul{p}roperty-\ul{a}ware \ul{opt}imization mechanism for drug developability built upon \method, referred to as \opt.
\method employs a \ul{p}ocket-\ul{c}onditioned \ul{l}igand \ul{g}enerator, referred to as \predictor, to predict 
atom types and structures of a potential binding ligand from the pocket representation learned from \PRL
and pocket-ligand interactions at both the atom and residue levels.
The {\predictor's predicted atom positions and types} are used to direct the diffusion process of \method towards the final  
generated ligands of high binding affinities.  
Beyond binding affinity, the training-free, plug-and-play \opt steers ligand generation towards both high binding affinities and 
favorable developability in terms of ADMET profiles.
In summary, \methodopt presents a new diffusion-based paradigm for generating ligands with high binding affinity and favorable drug developability profiles, bridging binding-driven design and developability optimization in real-world drug discovery and development settings.
Fig.~\ref{fig:model_architecture} presents the overview
architecture of \methodopt.

To evaluate \methodopt, as well as other SBDD methods in the literature~\cite{luo2021sbdd, peng22pocket2mol, guan2023targetdiff, guan2023decompdiff}, 
we carefully curate a new dataset, denoted as \dataset, on top of the widely-used benchmark dataset CrossDocked2020~\cite{Francoeur2020}, denoted as \CD. 
While well adopted, \CD includes both human and non-human protein targets 
(33\% and 67\%, respectively).
This poses risks to translational efficacy and safety, 
as even for conserved targets, sequence, structural, dynamics, and conformational differences
between human and non-human proteins can alter ligand binding and target engagement~\cite{marshall2023poor}.
To address this issue, \dataset is curated to consist exclusively of human targets,  
including human targets from \CD, the human orthologs of non-human targets in \CD, 
and some additional, carefully-selected, experimentally validated 
human targets from the Protein Database Bank (PDB)~\cite{berman2000protein} that are of 
high therapeutic interest in life-threatening conditions (e.g., cancers).
Over \dataset and \CD, 
we extensively compare \methodopt against seven state-of-the-art baselines in generating high-quality ligands. 
Computational results demonstrate that {\method} generates ligands with superior predicted binding affinity, achieving average scores of {$-8.85$} kcal/mol {and a 7.4\% improvement over the best baseline} on \dataset,
and {\methodopt} further improves ADMET properties on \dataset with an average improvement of {up to 73\%}
across five ADMET properties while maintaining comparable predicted binding affinities.

To further demonstrate that \methodopt can propose developable molecules,
we conduct case studies on the programmed death-ligand 1 (\PDLone), 
and colony-stimulating factor-1 receptor (\CSFoneR) proteins, which are two therapeutically important and validated targets with available structures.
Top-ranked \methodopt-generated ligands and their analogs are synthesized and tested in biological assays.
Across the two targets, the selected molecules show promising biological activities.  
In particular, two ligands generated by \methodopt for \PDLone show SPR-derived binding activities in the low micromolar range, with K$_\text{D}$ values below 3.80~$\mu$M. 
For \CSFoneR, two analogs derived from \methodopt-generated ligands show strong kinase activities and high selectivity, with IC$_{50}$ values below 0.31~$\mu$M and promiscuity hit rates of at most 2.4\%.
These biological testing results suggest that \methodopt is not limited to generating ligands with computationally predicted binding, 
but can also produce biologically active and developable molecules \emph{in vitro}.

In summary, we highlight the advantages of \methodopt as follows:

\begin{itemize}[noitemsep, topsep=0pt]
    \item \methodopt unifies pocket-conditioned generation and property optimization within a single framework, 
    incorporating developability considerations directly into the initial drug design process.
    \item \method produces ligands with strong binding affinities leveraging decoupled pocket representation learning.
    This decoupling allows \PRL to better capture binding pocket information, thus 
    providing \predictor with robust pocket conditioning.
    \item \PRL and \predictor encode multi-scale features of the binding pocket and pocket-ligand interactions, allowing molecule generation to be informed by both local atomic details and broader residue-level context that jointly govern ligand recognition and
    binding.
    \item \methodopt achieves favorable ADMET profiles by incorporating the training-free \opt to directly optimize those properties during the generation process of \method. 
    This allows external ADMET signals and pocket conditioning to jointly guide ligand generation.
    \item Case studies with extensive \emph{in silico} analyses on \PDLone and \CSFoneR
    demonstrate that \methodopt can generate developable molecules with binding modes similar to those of known binding 
    ligands of these two targets.
    \item Fifteen top-ranked 
    \methodopt-designed molecules for \PDLone and \CSFoneR, and their derivatives and analogs,
    have been experimentally synthesized and tested. Two hits have been identified with target inhibition at nanomolar concentrations, 
    and seven hits
    have target inhibition at low micromolar concentrations. 
\end{itemize}

\captionsetup[figure]{justification=justified,singlelinecheck=true}

\section*{Related Work}

\paragraph{Generative Models for SBDD}

Structure-based drug design (SBDD) has leveraged various generative modeling approaches to design ligands tailored to specific protein pockets. Peng \etal~\cite{peng22pocket2mol} constructed an encoder-predictor framework \pockettwomol that can generate ligands in an autoregressive way.
However, \pockettwomol relies on the autoregressive sampling process, which tends to violate geometric constraints and produce ligands with limited interactions with the pocket, resulting in low binding affinity.
More recently, target-aware diffusion models\cite{guan2023targetdiff,guan2023decompdiff,zhoudecompopt,gu2024aligning,huang2024protein} for ligand 
generation have largely addressed these limitations,
offering better geometric consistency and binding affinities.
Guan \etal~\cite{guan2023targetdiff} pioneered this direction by introducing an E(3)-equivariant conditional diffusion 
model~\cite{fuchs2020se,satorras2021n,ho2020ddpm,hoogeboom2021argmax} \targetdiff that jointly generates atomic coordinates of ligands in 3D and types conditioned on protein pocket structures.  
It achieves this through careful design of equivariant score networks
and denoising processes that respect geometric symmetries, thereby establishing a strong baseline for diffusion-based SBDD methods.
To further improve conformational stability and molecular validity, Guan \etal~\cite{guan2023decompdiff} proposed \decompdiff. 
Inspired by medicinal chemistry practice of scaffold-and-arm design~\cite{schneider1999scaffold}, \decompdiff decomposes ligand molecules into scaffolds and arms and learns separate diffusion priors for each.
This decomposed design facilitates more structured generation and enhances efficiency in exploring chemical space.
Building upon the idea of a decomposed prior, Zhou \etal~\cite{zhoudecompopt} introduced \decompopt, which decomposes pocket-ligand interactions into local subpocket-arm interactions 
and employs iterative optimization over the arms.
However, \targetdiff, \decompdiff and \decompopt primarily rely on geometric features of the pocket and ligand-only priors
without directly encoding protein-ligand interaction signals into the diffusion dynamics.
To explicitly leverage such interactions,
Huang \etal~\cite{huang2024protein} proposed \ipdiff, which redesigns both forward and reverse processes to be informed by a learnable binding-affinity signal.
This adaptation guides the generation toward ligands with improved binding properties. 
Whereas \ipdiff steers trajectories using a specific supervised affinity signal, 
Gu \etal~\cite{gu2024aligning} introduced a more general framework \alidiff, which controls generation by aligning the reverse process of a pretrained diffusion model with desired preferences via Diffusion-DPO~\cite{wallace2024diffusion} adapted to molecular space.
Beyond diffusion-based methods, recent work has also explored flow matching as an alternative generative paradigm for SBDD.
For example, Cremer \etal~\cite{cremer2026flowr} introduced \flowr, which combines continuous and categorical flow matching with equivariant optimal transport for ligand generation.
Despite the different designs of current generative models, existing methods entangle pocket encoding with ligand denoising through a single end-to-end generation objective, 
which can blur interaction signals. 
This gap motivates our design of decoupling the two: obtain a stable pretrained pocket summary first, 
then use it as a fixed context, enabling denoising to focus solely on pocket-ligand interactions.
Moreover, prior works process pocket information only at the atom level, which may miss higher-level structural patterns encoded by residues.
In contrast, our approach combines both atom- and residue-level pocket information to capture pocket characteristics and protein-ligand interactions at multiple scales.
Furthermore, most prior works focus on improving binding affinity while ignoring ADMET profiles.
Our approach seeks to address this issue by explicitly incorporating ADMET-aware optimization into the generation process.

\paragraph{Alignment for Diffusion Models}
There {is} a growing body of research on adapting pretrained generative diffusion models to specific downstream objectives. These approaches can be broadly categorized into training-based and training-free methods, depending on whether the model parameters are updated. In training-based methods, Prabhudesai \etal~\cite{prabhudesai2023aligning} proposed directly backpropagating through differentiable reward functions into diffusion model parameters. While effective when such reward functions exist, this approach is prone to reward hacking in the absence of strong regularization. Black \etal~\cite{blacktraining} applied policy gradient methods such as PPO to finetune diffusion models. Although more stable than direct backpropagation, reinforcement learning finetuning remains resource- and time-intensive, limiting its flexibility for diverse downstream tasks. By contrast, training-free methods operate in a plug-and-play fashion, requiring {minimal} or no parameter updates. Song \etal~\cite{song2023loss} introduced reward-guided sampling, incorporating estimated reward gradients as guidance terms during generation. Li \etal~\cite{li2024derivative} proposed a value-based decoding strategy that selects denoised samples according to estimated value functions. More recently, Ma \etal~\cite{ma2025inference} framed alignment as a search problem and introduced a search-over-paths algorithm to exploit generation-time scaling behaviors. While these methods advance training-free alignment, they share a key limitation: 
it is difficult to reliably assess the quality of generated samples before the generative process is complete. 

To address the above-mentioned limitations, a recent emerging line of work formulates the entire generation process as an optimization problem,
where the objective is to directly optimize task-specific downstream performance of the final generated sample. 
Concretely, in standard diffusion sampling, each denoising step draws the next state from the model-predicted posterior by combining a deterministic update (the posterior mean) with stochastic noise injected by the sampler.
By treating the injected noise variables as optimization variables, one can perturb the denoising trajectory and thereby steer the final sample toward preferred outcomes while keeping the diffusion model fixed. Eyring \etal~\cite{eyring2024reno} optimized the noise variable at the initial step of diffusion generation against external reward functions, achieving improved alignment at inference. Tang \etal~\cite{tanginference} extended this by optimizing injected noise across the entire denoising trajectory, propagating reward signals on the final outputs back to noise inputs while keeping the model parameters fixed. 
While technically sound and effective in the field of image generation, these methods could not be readily applied for generating new chemical entities (NCEs). In particular, they are typically designed for continuous Gaussian noise and differentiable objectives, making it nontrivial to handle discrete (categorical) distribution over atom types and to incorporate molecular property evaluators, which are often black-box and non-differentiable.
Our proposed optimization scheme builds upon the idea of noise optimization with several key novelties.
First, whereas prior methods operate only on continuous latent noise, we extend the optimization framework to handle the categorical sampling process for discrete atom types. 
Second, while existing works mainly consider differentiable reward functions, we investigate practical solutions for optimization under non-differentiable, black-box evaluators so that the framework can be applied to real-world scenarios. Finally, to the best of our knowledge, this work is the first to explore noise optimization in the context of drug design, and we demonstrate that this can be an effective mechanism for aligning diffusion-based generation of NCEs with specific objectives (e.g., higher absorption and lower toxicity).

\section*{Materials}
%
\subsection*{Datasets} %

We use the training and testing sets of the widely-adopted 
CrossDocked2020~\cite{Francoeur2020} benchmark, referred to as \CD, for model training and testing, respectively.
In addition, we also use a new, manually annotated dataset, \dataset, for testing.

\subsubsection*{Training data}
%
We follow the data preparation and splitting process of Luo \etal~\cite{luo2021sbdd}, refining 22.5 million docked binding complexes to high-quality docking poses.
For each protein in \CD, there are multiple ligand-binding complexes: 
\textbf{(1)} determined experimentally and reported in the Protein Data Bank (PDB)~\cite{berman2000protein},
and \textbf{(2)} obtained through simulation
by redocking ligands into non-cognate receptors. 
We retain high-quality poses, including experimentally determined and docked poses whose
heavy-atom RMSD < 1\;\AA\; relative to the experimentally-observed ligands pose from the PDB, and select diverse proteins with pairwise sequence identity below 30\%.
After filtering, the dataset consists of 100,000 protein-ligand pairs for training.
As \CD does not provide explicit pocket regions, following Guan \etal~\cite{guan2023targetdiff}, 
we define the binding pocket as all residues and their atoms within a 10\;\AA \;radius of its ligand.

\subsubsection*{Testing data}

\paragraph*{\CD testing data} 

The \CD test set is consistent with the data preparation and splitting process described in {\CD}’s training data. We use all 100 protein-ligand complexes 
from {\CD}'s test set for testing. 
We refer to the known ligand provided in the test set as the {reference ligand}.
The latter is used only to identify the pocket region in the protein structure.
Once the pocket is identified, it is provided to \method and \methodopt 
as input, while the reference ligand itself is excluded from the generation process.
The test set includes human proteins associated with oncological, neurodegenerative, infectious, cardiovascular, and metabolic diseases, 
as well as non-human proteins such as bacterial, fungal, parasitic, viral, and plant proteins.

\paragraph*{\dataset testing data} 

While \CD's test set has been widely used in the research community, we discovered that it contains 
protein targets for non-human species, such as plants and bacteria, which limits its utility 
for human therapeutics discovery. 
To address this limitation, we manually curated a new benchmarking dataset based on \CD, referred to as \dataset, 
which contains only human targets.  
\dataset retains all 33 human targets from \CD's test set, and includes an additional
39 human targets (9 orthologs, 30 new targets) from the PDB.
These additional PDB targets are carefully selected to cover high-burden diseases such as cancers, neurological 
disorders, and autoimmune and inflammatory diseases, making \dataset more relevant to human diseases. 
Table~\ref{tbl:target_property} in Appendix~\ref{appendix:dataset_info} presents the \dataset information.

For all the targets in \dataset, we consider their relevant 
key
ADMET properties, including: 
\textbf{(1)} carcinogenicity (Carci), the propensity of a molecule to cause cancer, 
\textbf{(2)} Ames mutagenicity (Ames),  
the property of a molecule likely to mutate DNA, 
\textbf{(3)} hERG inhibition (hERG), the potential cardiac side effects 
due to inhibition of the cardiac hERG (KCNH2) potassium channel,  
\textbf{(4)} human intestinal absorption (HIA), the ability of an oral drug to be absorbed through the intestinal lining, 
and 
\textbf{(5)} blood-brain barrier permeability (BBBP), the ability to cross the blood-brain barrier and enter 
the central nervous system. 
Assessment of carcinogenicity, Ames mutagenicity, and hERG inhibition is essential to ensure drug safety 
by mitigating risks of tumorigenicity, genotoxicity, and cardiotoxicity. 
Evaluation of human intestinal absorption and blood-brain barrier permeability is critical to 
establish pharmacokinetic suitability and therapeutic accessibility of drug candidates.
These ADMET properties represent key determinants in drug discovery, 
guiding the identification and optimization of viable therapeutic candidates.
Unfortunately, current generative SBDD models do not proactively evaluate or optimize 
these properties, largely missing opportunities 
to increase the overall success rate of the drug discovery and development process. 
\dataset is constructed to enable such ADMET evaluation.

\subsection*{Baselines} %

To evaluate the effectiveness of \method and \methodopt in generating ligands that bind to target protein pockets,
we compare them against the following state-of-the-art baselines for SBDD: 
{\AR}~{\cite{luo2021sbdd}}, {\pockettwomol}~{\cite{peng22pocket2mol}}, {\diffsbdd}~{\cite{schneuing2022structure}}, {\targetdiff}~{\cite{guan2023targetdiff}}, {\decompdiff}~{\cite{guan2023decompdiff}}, {\decompopt}~{\cite{zhoudecompopt}},
{\ipdiff}~{\cite{huang2024protein}} and
{\alidiff}~{\cite{gu2024aligning}}, where \AR and \pockettwomol
are non-diffusion methods, and the rest are all diffusion-based methods. 
These methods generate 3D binding ligands conditioned on the pockets of protein targets. 
We choose these baselines because they are well-established SBDD methods with strong performance on \CD. 
We used their author-provided implementations with released checkpoints and default inference settings 
for all baselines to conduct computational experiments on \dataset (except \alidiff, which does not provide checkpoints). 
%

\subsection*{Model Training and Evaluation} %

All the models, including \method, \methodopt, and the baselines, are trained over \CD training data, 
and tested on the \CD test set and \dataset.
In \method, \pocketenc and \diff have different training objectives, thus,  
\pocketenc is trained using only the protein pockets from \CD training set, and 
\diff is trained using the protein-ligand complexes.
In \methodopt, we focus on the targets in \dataset and their ADMET property optimization, and 
its generation-time optimization does not need model finetuning. 
For each target in \dataset, \methodopt optimizes one particular ADMET property key to that target 
during the generation time. The properties to optimize for different disease targets are presented in Table~\ref{tbl:target_property} in Appendix~\ref{appendix:dataset_info}. 
Thus, the evaluation is organized with respect to each ADMET property across multiple relevant targets.

\subsection*{Evaluation Metrics}

\subsubsection*{General ligand properties}

%
To predict binding affinity, following the literature~\cite{guan2023targetdiff, guan2023decompdiff},
we use Vina Scores (Vina S) calculated by AutoDock Vina~\cite{Eberhardt2021},
which evaluates the quality of the binding poses of generated ligands against protein targets. 
In addition, as suggested in the literature~\cite{guan2023targetdiff, guan2023decompdiff}, 
we optimize the poses of the generated 3D ligands using a local energy minimization algorithm 
and a docking algorithm, as implemented in AutoDock Vina~\cite{Eberhardt2021}.
We then evaluate the predicted binding affinities of these optimized poses via two metrics: 
Vina Minimization (Vina M) based on local energy minimization, and Vina Dock (Vina D) based on docking.
Lower Vina scores indicate stronger binding affinities.
Based on Vina D, following the literature~\cite{guan2023targetdiff, guan2023decompdiff}, 
we also measure the percentage of how many generated ligands across all the targets 
bind better than their reference ligand, 
referred to as High Affinity percentage (HA\%).
Higher HA\% indicates a better capacity to generate ligands above the reference affinities.

For drug-likeness, we evaluate whether the generated ligands are drug-like using the quantitative estimate of drug-likeness (QED)~\cite{Bickerton2012} and synthesizable using a synthetic accessibility (SA) score~\cite{Ertl2009}.
We also calculate the diversity among generated ligands of each target, which is a per-target measurement and defined as the average pairwise Tanimoto distances~\cite{bajusz2015tanimoto} derived using 2048-bit fingerprints as implemented within RDKit~\cite{rdkit}. Higher diversity indicates a better ability to explore broader chemical space.
To jointly consider predicted binding affinity, drug-likeness, and synthetic feasibility, following previous work~\cite{guan2023targetdiff, guan2023decompdiff}, 
we evaluate the success rate (SR\%) calculated as the percentage of all generated ligands across all targets
with Vina D $<$ -8.18, QED $>$ 0.25, and SA $>$ 0.59.
Higher SR\% indicates a better ability to generate high-quality ligands.
The Vina D score threshold -8.18, which corresponds to a binding affinity of less than 1 $\mu$M, 
is widely recognized in medicinal chemistry as an indicator of moderate biological activity~\cite{zhoudecompopt}.
The QED and SA score thresholds of 0.25 and 0.59 are set as the 10th percentile
of approved drugs in DrugCentral~\cite{ursu2016drugcentral}, enforcing reasonable lower bounds 
to maximize the consideration of potentially promising ligands. 
%

\subsubsection*{ADMET properties}

We evaluate the generated ligands 
on five ADMET properties: Carci, Ames, hERG, HIA, and BBBP. 
We use \mbox{ADMET-AI}~\cite{swanson2024admet} to estimate a probabilistic score for each property. 
For Carci, Ames, and hERG, lower scores are preferred, indicating lower toxicity risks, 
while for HIA and BBBP, higher scores are more desired, indicating higher human intestinal absorption and 
blood-brain barrier permeability, respectively. 
%

\subsubsection*{Ligand structures}
%
We also evaluate the quality of generated ligands by analyzing their stability and 3D structural quality.
Specifically, to evaluate stability, we calculate both atomic level and molecular level stability, as described in literature~\cite{hoogeboom22diff}.
Atomic level stability is defined by the percentage of atoms that maintain correct valency, 
while molecular level stability measures the percentage of molecules in which all the atoms are stable, 
both for all the generated ligands and for all the targets. 
We evaluate the 3D structures of generated ligands using the same metrics as in Peng \etal~\cite{peng2023moldiff}.
We calculate the 
Jensen-Shannon (JS) divergences for bond lengths, bond angles, and dihedral angles,
which measure how far the distributions of these properties of generated ligands are compared 
with those of real molecules (i.e., training molecules) regarding the 3D structures.

\section*{Results}

We assess the performance of the baselines and \method 
 on the 100 test pockets from \CD test set and the 72 curated test pockets from \dataset, 
using the evaluation metrics with 100 ligands generated per pocket by these methods.
We evaluate \methodopt using the \dataset, optimizing the 100 ligands per pocket 
with respect to the ADMET properties most relevant to its corresponding therapeutic indication.
Here, we present the results on \dataset; the results on \CD are discussed in Appendix~\ref{supp:cd}.
%

\subsection*{Overall Comparison on \dataset}

\begin{table}[H]
	\centering
		\caption{Comparison on \dataset}
	\label{tbl:overall_results_curated}
\begin{threeparttable}
	\begin{tabular}{
		@{\hspace{2pt}}l@{\hspace{2pt}}
		@{\hspace{2pt}}r@{\hspace{2pt}}
		@{\hspace{2pt}}r@{\hspace{2pt}}
		@{\hspace{4pt}}r@{\hspace{4pt}}
		@{\hspace{2pt}}r@{\hspace{2pt}}
		@{\hspace{2pt}}r@{\hspace{2pt}}
		@{\hspace{5pt}}r@{\hspace{5pt}}
		@{\hspace{2pt}}r@{\hspace{2pt}}
		@{\hspace{2pt}}r@{\hspace{2pt}}
		@{\hspace{5pt}}r@{\hspace{5pt}}
		@{\hspace{2pt}}r@{\hspace{2pt}}
	         @{\hspace{2pt}}r@{\hspace{2pt}}
		@{\hspace{5pt}}r@{\hspace{5pt}}
		@{\hspace{2pt}}r@{\hspace{2pt}}
		@{\hspace{2pt}}r@{\hspace{2pt}}
		@{\hspace{5pt}}r@{\hspace{5pt}}
		@{\hspace{2pt}}r@{\hspace{2pt}}
		@{\hspace{2pt}}r@{\hspace{2pt}}
		@{\hspace{5pt}}r@{\hspace{5pt}}
		@{\hspace{2pt}}r@{\hspace{2pt}}
		@{\hspace{2pt}}r@{\hspace{2pt}}
		@{\hspace{5pt}}r@{\hspace{5pt}}
		@{\hspace{2pt}}r@{\hspace{2pt}}
		@{\hspace{2pt}}r@{\hspace{2pt}}
		@{\hspace{2pt}}r@{\hspace{2pt}}
		%
		}
		\toprule
		\multirow{2}{*}{Method} & \multicolumn{2}{c}{Vina S$\downarrow$} & & \multicolumn{2}{c}{Vina M$\downarrow$} & & \multicolumn{2}{c}{Vina D$\downarrow$} & & \multicolumn{2}{c}{{{HA}\%$\uparrow$}}  & & \multicolumn{2}{c}{QED$\uparrow$} & & \multicolumn{2}{c}{SA$\uparrow$} & & \multicolumn{2}{c}{Div$\uparrow$} & & \multirow{2}{*}{SR\%$\uparrow$} & 
        \\
	    \cmidrule{2-3}\cmidrule{5-6} \cmidrule{8-9} \cmidrule{11-12} \cmidrule{14-15} \cmidrule{17-18} \cmidrule{20-21} 
		& Avg. & Med. &  & Avg. & Med. &  & Avg. & Med. & & Avg. & Med.  & & Avg. & Med.  & & Avg. & Med.  & & Avg. & Med. &  \\
		\midrule
		Reference    & -7.45 & -7.35 & & -7.89 & -7.63 & &  -8.01 & -7.71 & &  - & - & &  0.46 & 0.46 & &  0.73 & 0.76 & &  - & - & & 29.2 \\
		\midrule
		\AR & \underline{-6.63} & -6.51 & &  -6.95 & -6.69 & &  -7.43 & -7.21 & & 37.0 & 21.6 & & \underline{0.52} & \underline{0.53} & &  {0.63} & {0.63} & & {0.69} & {0.69}  & & 12.2   \\
		\pockettwomol   & -5.28 & -5.02 & &  -6.44 & -6.16 & &  -7.23 & -7.00 & & 33.1 & 15.7 & & \textbf{0.61} & \textbf{0.61} & & \textbf{0.79} & \textbf{0.80} & & \textbf{0.79} & \textbf{0.81}  & &  \textbf{29.7} \\
		\diffsbdd  & -3.24 & -5.12 & &  -5.27 & -5.94 & & -7.02 & -7.30 & &  34.1 & 21.9 & &  0.48 & 0.49 & &  0.63 & 0.61 & & \underline{0.78} & \underline{0.75} & &  11.3 \\
		\targetdiff     & -6.38 & \underline{-6.67} & &  \underline{-7.18} & \underline{-7.25} & & \underline{-8.24} & \underline{-8.19} & & 47.1 & 43.9 & &  0.46 & 0.47 & &  0.58 & 0.57 & & 0.71 & 0.70 & &  12.3   \\
        \decompdiff & -6.17 & -6.02 & & -6.96 & -6.76 & & -7.88 & -7.83  && \underline{49.8} & \underline{45.5} && 0.48 & 0.48 && 0.64 & 0.63 && 0.67 & 0.66 &&  18.3 \\
        \decompopt & -6.34 & -6.22 & &  -7.04 & -6.88 & &  -7.98 & -7.83 & &  51.7 &  52.9 & &  0.48 & 0.48 & & \underline{0.65} & \underline{0.64} & & 0.67 & 0.67 & &  20.5  \\
        \grey{\ipdiff} & \grey{-8.02} & \grey{-8.15} & & \grey{-8.43} & \grey{-8.30} & & \grey{-9.24} & \grey{-8.91} & & \grey{62.0} & \grey{66.3} & & \grey{0.46} & \grey{0.46} & & \grey{0.55} & \grey{0.54} & & \grey{0.72} & \grey{0.71} & & \grey{10.0} \\
		\midrule
		\method       &  \textbf{-6.80} & \textbf{-7.10} & & \textbf{-7.69} & \textbf{-7.78} & & \textbf{-8.85} & \textbf{-8.71} & &  \textbf{62.2} & \textbf{74.6} & &  0.46 & 0.45 & &  0.59 & 0.58 & & 0.60 & 0.58 & &  \underline{22.0} \\
		\bottomrule
	\end{tabular}%
	\begin{tablenotes}[normal,flushleft]
		\begin{footnotesize}
	\item 
\!\!Columns represent: ``Vina S'': the predicted binding affinities between the initially generated poses of ligands and the protein pockets; 
		``Vina M'': the predicted binding affinities between the poses after local structure minimization and the protein pockets;
		``Vina D'': the predicted binding affinities between the poses determined by AutoDock Vina and the protein pockets;
		``HA'': the percentage of generated ligands with Vina D lower than those of reference ligands;
		``QED'': the quantitative estimate of drug-likeness;
		``SA'': the synthetic accessibility score;
		``Div'': the diversity among generated ligands;
            ``SR'': the percentage of generated molecules with predicted binding affinities, QED and SA above certain thresholds. 
            The best values are in \textbf{bold}, and the second-best values are
            \underline{underlined}. Rows in \grey{gray} are excluded from the comparison. 
            $\uparrow$ / $\downarrow$ indicate that higher / lower values are better.
		\par
		\par
		\end{footnotesize}
	\end{tablenotes}
\end{threeparttable}    
\end{table}

Table~\ref{tbl:overall_results_curated} presents a comprehensive comparison between \method and 
all the baselines in terms of generating drug-like and diverse ligands that effectively bind to protein pockets 
of \dataset. 
\ipdiff is reported for completeness but excluded from best/second-best marking because ligands generated by this method exhibit simplified structures with excessive carbons and single bonds 
(Details are in the \DistributionalAnalysisSection section). 
In terms of Vina S (Avg/Med), \method outperforms all the baseline methods, and also achieves 
the highest HA\%, with a 2.6\% and 24.9\% improvement over the best baselines on Vina S and HA\%, respectively. 
This suggests that \method better guides ligand atoms toward regions where they can engage in interactions with the pocket.
After energy minimization and re-docking, \method remains superior in terms of Vina M and Vina D scores, with 
7.1\% and 7.4\% improvement over the best baselines, respectively. 
This indicates the generated initial poses offer strong starting points for further refinement.
\AR is the second-best model in terms of Vina S. 
\AR learns pocket-anchored atom type density maps, 
making it easier to place atoms and achieve high binding affinities.
In terms of Vina M and Vina D, \targetdiff achieves the second-best performance.
Unlike \AR, whose auto-regressive sampling strategy suffers from high variance in pose quality, 
\targetdiff generates more consistent poses, improving Vina M and Vina D over \AR but still underperforming \method.
Among the baseline methods, \decompopt specifically performs iterative, docking-guided arm optimization, 
aiming to achieve high binding affinities and thus high Vina scores.
Instead, \method does not explicitly conduct such optimization -- the fact that it still outperforms \decompopt
in binding affinity scores indicates \method encodes pocket information and pocket-ligand interactions better than baselines. 
Compared with diffusion models that do not perform any Vina-based optimization 
(\diffsbdd, \targetdiff, \decompdiff), \method attains the best binding affinity scores while being more sample efficient. 
The key difference is that \method conditions generation on a pretrained pocket embedding, 
so pocket information is available \emph{a priori} and does not need to be encoded every timestep during sampling, 
reducing overhead and improving pose quality. 
In addition, \diffsbdd shows substantially lower Vina S than other methods. 
A possible reason is coarse encoding of pocket structure, making \diffsbdd ignore some important structural information and potential interaction sites.
This leads {\diffsbdd} to struggle to place ligands correctly within the pocket, causing steric clashes, suboptimal functional-group orientations, and consequently higher Vina scores.
In contrast, \method avoids this issue through its dedicated pocket encoder.

The \method method achieves QED and SA scores comparable with those of other diffusion baselines. 
Notably, the two autoregressive models (\AR and \pockettwomol) exhibit QED and SA 
scores significantly higher than those of all diffusion-based models. 
These two methods tend to generate smaller and simpler molecules via an atom-by-atom autoregression, 
with an average heavy-atom count of 17 and 20, respectively, compared to 26
from other diffusion-based baselines and 28 from the references. 
While such molecules are favored by QED and SA measurement due to 
small size, simple structures and limited scaffolds~\cite{bickerton2012quantifying,ertl2009estimation},
\AR and \pockettwomol do not generally enjoy the same favor in terms of binding affinities. 
Among the diffusion-based methods, the decomposition baselines (\decompdiff and \decompopt)
achieve highest QED and SA scores. 
Their decomposed, multiple priors for different atom roles bias the generation of molecules toward those with more drug-like, synthesizable 
fragments, yielding higher QED and SA scores than those from other diffusion models of a single prior. 
\method emphasizes binding-quality during generation, and therefore, shows moderate QED and SA with 
high binding affinities. 

We observe that \method\!’s diversity slightly falls behind the baselines. 
However, \method\!’s diversity is still acceptable given its strengths in generating high-affinity molecules, and it is comparable with those of \decompdiff and \decompopt.
We hypothesize that the diversity is limited because the pocket is pre-encoded as a fixed condition for diffusion. 
A static pocket representation may sharpen the denoising landscape, limiting the chemical space that \method may visit.
In contrast, the baselines that jointly diffuse ligand and pocket allow the pocket representation to adapt to noisy states of ligands, and thus, achieve higher ligand diversity. 
In terms of SR\%, 
\method attains the second-best SR\%, notably above all diffusion baselines, indicating that it balances design objectives of drugs effectively and yields realistic ligand candidates that are both effective binders and chemically viable. 
\pockettwomol achieves the highest SR\%, consistent with its strong QED and SA performance.
However, \pockettwomol significantly underperforms \method in terms of Vina scores.

\subsection*{Overall Comparison after ADMET Optimization}

\paragraph{Overall Performance of \methodopt} 

\begin{table}[H]
	\centering
	\caption{Comparison on \dataset with optimized property Carci}
	\label{tbl:overall_results_carci}

\begin{threeparttable}
\resizebox{\textwidth}{!}{
\begin{minipage}{\textwidth}

	\begin{tabular}{
		@{\hspace{2pt}}l@{\hspace{2pt}}
		@{\hspace{2pt}}r@{\hspace{2pt}}
		@{\hspace{2pt}}r@{\hspace{2pt}}
		@{\hspace{4pt}}r@{\hspace{4pt}}
		@{\hspace{2pt}}r@{\hspace{2pt}}
		@{\hspace{2pt}}r@{\hspace{2pt}}
		@{\hspace{5pt}}r@{\hspace{5pt}}
		@{\hspace{2pt}}r@{\hspace{2pt}}
		@{\hspace{2pt}}r@{\hspace{2pt}}
		@{\hspace{5pt}}r@{\hspace{5pt}}
		@{\hspace{2pt}}r@{\hspace{2pt}}
		@{\hspace{2pt}}r@{\hspace{2pt}}
		@{\hspace{5pt}}r@{\hspace{5pt}}
		@{\hspace{2pt}}r@{\hspace{2pt}}
		@{\hspace{2pt}}r@{\hspace{2pt}}
		@{\hspace{5pt}}r@{\hspace{5pt}}
		@{\hspace{2pt}}r@{\hspace{2pt}}
		@{\hspace{2pt}}r@{\hspace{2pt}}
		@{\hspace{5pt}}r@{\hspace{5pt}}
		@{\hspace{2pt}}r@{\hspace{2pt}}
		@{\hspace{2pt}}r@{\hspace{2pt}}
		@{\hspace{5pt}}r@{\hspace{5pt}}
		@{\hspace{2pt}}r@{\hspace{2pt}}
		@{\hspace{2pt}}r@{\hspace{2pt}}
		@{\hspace{2pt}}r@{\hspace{2pt}}
		@{\hspace{2pt}}r@{\hspace{2pt}}
		@{\hspace{2pt}}r@{\hspace{2pt}}
	}

	\toprule
	\multirow{2}{*}{Method} 
	& \multicolumn{2}{c}{Vina S$\downarrow$} 
	& & \multicolumn{2}{c}{Vina M$\downarrow$} 
	& & \multicolumn{2}{c}{Vina D$\downarrow$} 
	& & \multicolumn{2}{c}{{HA}\%$\uparrow$}  
	& & \multicolumn{2}{c}{QED$\uparrow$} 
	& & \multicolumn{2}{c}{SA$\uparrow$} 
	& & \multicolumn{2}{c}{Div$\uparrow$} 
	& & \multirow{2}{*}{SR\%$\uparrow$} 
	& & \multicolumn{2}{c}{Carci$\downarrow$} 
	\\
	\cmidrule{2-3}\cmidrule{5-6}\cmidrule{8-9}
	\cmidrule{11-12}\cmidrule{14-15}\cmidrule{17-18}
	\cmidrule{20-21}\cmidrule{25-26}
	& Avg. & Med. &  & Avg. & Med. &  & Avg. & Med. &
	  & Avg. & Med. &  & Avg. & Med. &  & Avg. & Med. &
	  & Avg. & Med. &  & & & Avg. & Med. \\
	\midrule

	Reference & -7.53 & -7.45 && -7.93 & -7.68 && -8.00 & -7.71 && - & - && 0.45 & 0.43 && 0.73 & 0.77 && - & - && 25.0 && 0.19 & 0.14\\
	\midrule
	\AR & -6.52 & -6.44 && -6.87 & -6.61 && -7.37 & -7.09 && 36.7 & 21.0 && \underline{0.52} & \underline{0.53} && 0.63 & 0.63 && 0.68 & 0.68 && 11.2 && \underline{0.14} & \underline{0.07}\\
	\pockettwomol & -5.30 & -5.04 && -6.46 & -6.15 && -7.25 & -6.96 && 34.5 & 14.5 && \textbf{0.61} & \textbf{0.62} && \textbf{0.79} & \textbf{0.80} && \textbf{0.79} & \textbf{0.80} && \textbf{29.5} && 0.18 & 0.13\\
	\diffsbdd & -3.11 & -5.12 && -5.22 & -5.92 && -6.88 & -7.25 && 34.3 & 20.3 && 0.49 & 0.50 && 0.63 & 0.62 && \underline{0.78} & \underline{0.76} && 11.6 && 0.18 & 0.12\\
	\targetdiff & -6.37 & \underline{-6.60} && -7.17 & -7.23 && -8.20 & -8.15 && 47.9 & 41.1 && 0.46 & 0.47 && 0.59 & 0.58 && 0.71 & 0.69 && 12.3 && 0.20 & 0.15\\
	\decompdiff & -6.12 & -5.88 && -6.93 & -6.63 && -7.86 & -7.74 && 48.5 & 49.5 && 0.48 & 0.49 && 0.64 & 0.62 && 0.66 & 0.66 && 17.1 && 0.20 & 0.15\\
	\decompopt & -6.30 & -6.14 && -7.02 & -6.79 && -8.00 & -7.78 && 54.1 & 56.1 && 0.48 & 0.49 && \underline{0.64} & \underline{0.64} && 0.67 & 0.66 && 19.1 && 0.20 & 0.15\\
	\grey{\ipdiff} & \grey{-7.98} & \grey{-8.23} && \grey{-8.41} & \grey{-8.40} && \grey{-9.22} & \grey{-8.91} && \grey{62.1} & \grey{66.3} && \grey{0.46} & \grey{0.47} && \grey{0.55} & \grey{0.54} && \grey{0.72} & \grey{0.71} && \grey{10.2} && \grey{0.22} & \grey{0.18}\\
	\midrule
	\method    & \textbf{-6.66} & \textbf{-6.80} && \textbf{-7.46} & \textbf{-7.45} && \textbf{-8.48} & \textbf{-8.42} && \textbf{55.9} & \textbf{62.3} && 0.49 & 0.51 && 0.62 & 0.60 && 0.65 & 0.62 && \underline{23.0} && 0.22 & 0.22 \\
	\methodopt & \underline{-6.55} & \textbf{-6.80} && \underline{-7.35} & \underline{-7.38} && \underline{-8.37} & \underline{-8.36} && \underline{54.2} & \underline{56.6} && \underline{0.52} & \underline{0.53} && 0.60 & 0.58 && 0.63 & 0.61 && 18.2 && \textbf{0.06} & \textbf{0.06}\\
	\bottomrule
	\end{tabular}

	\vspace{2pt}
	\begin{tablenotes}[flushleft]
	\footnotesize
	\item 
    \!\!Columns represent: ``Vina S'': the predicted binding affinities between the initially generated poses of ligands and the protein pockets; 
		``Vina M'': the predicted binding affinities between the poses after local structure minimization and the protein pockets;
		``Vina D'': the predicted binding affinities between the poses determined by AutoDock Vina and the protein pockets;
		``HA'': the percentage of generated ligands with Vina D lower than those of reference ligands;
		``QED'': the quantitative estimate of drug-likeness;
		``SA'': the synthetic accessibility score;
		``Div'': the diversity among generated ligands;
            ``SR'': the percentage of generated molecules with binding affinities, QED, and SA above a certain threshold. 
            ``Carci'': the predicted carcinogenicity score for generated molecules;
            The best values are in \textbf{bold}, and the second-best values are
            \underline{underlined}. Rows in \grey{gray} are excluded from the comparison. 
            $\uparrow$ / $\downarrow$ indicate that higher / lower values are better.
	\end{tablenotes}

\end{minipage}
}
\end{threeparttable}

\end{table}

The \methodopt method maintains strong performance on general molecular metrics compared with all evaluated baselines, ranking among the top methods in predicted binding affinity (Vina S/M/D), drug-likeness (QED), and synthetic accessibility (SA).  %
To examine \methodopt's performance  
in more detail, in Table~\ref{tbl:overall_results_carci}, we take the carcinogenicity optimization setting as an example, and compare the general metrics of \methodopt ligands with those of \method ligands to ensure that the ADMET optimization procedure does not disrupt the desirable structures of initial \method ligands.
In general, the results show that ligands produced by \methodopt achieve Vina S/M/D, QED, and SA scores that are on par with, and in some cases exceed, those of \method.
As shown in Table~\ref{tbl:overall_results_carci}, \methodopt reduces the average Carci score from {0.22 to 0.06} (by {73}\%), which is the best among all methods, while the average Vina S/M/D scores change only slightly by less than 2.5\%. Despite this marginal degradation, the resulting Vina scores of \methodopt remain stronger than all other baselines (except \ipdiff).
The \methodopt and \method methods perform very similarly across QED and SA scores in Tables~\ref{tbl:overall_results_carci}, indicating that drug-likeness and synthetic accessibility are well preserved after optimization.
This preservation can be attributed to the noise optimization design in \methodopt: rather than directly modifying the molecular structures of ligands generated by \method, the optimization is performed in the diffusion noise space, refining the generation process while keeping ligands within \method's learned generative distribution.
We note that the diversity of \methodopt ligands remains similar to that of \method ligands, suggesting that \methodopt enhances ADMET properties without collapsing the generated molecules into a narrow structural family.
We also summarize the results of optimizing other ADMET properties (Ames, hERG, BBBP and HIA) in Tables~\ref{tbl:overall_results_ames}-\ref{tbl:overall_results_hia} in Appendix~\ref{supp:opt_results}, and observe a similar pattern to carcinogenicity optimization: \methodopt consistently improves the targeted ADMET score substantially while only modestly affecting binding affinity scores and other general molecular metrics.

\paragraph{ADMET Property Optimization}

%
Our results demonstrate that, leveraging noise optimization, the sampling process of \methodopt is guided towards a chemical subspace with more favorable ADMET properties, facilitating efficient discovery of safer and more effective drug candidates.
Table~\ref{tbl:overall_results_admet} presents a comparison of generated molecules across different ADMET properties of interest, where each property score is optimized in a separate experiment. The reported results of \methodopt are obtained by evaluating the best-performing outcomes across multiple optimization iterations. As the results show, \methodopt consistently generates molecules with substantially more favorable ADMET properties compared to those from the baselines.
Specifically, \methodopt produces ligands with the lowest predicted carcinogenicity and Ames mutagenicity, achieving roughly 50\% to 60\% better average Carci and Ames scores, compared to all evaluated baselines. This indicates \methodopt generates drug candidates with much lower predicted risks of cancer induction and DNA damage across evaluated protein targets.
Similarly, Table~\ref{tbl:overall_results_admet} demonstrates that \methodopt decreases the hERG property score by 32\%, relative to the next best baseline (\pockettwomol), 
indicating considerable reduction in the predicted cardiotoxicity of the generated ligands.
Finally, \methodopt enhances absorption-related properties, achieving notable improvements in both BBBP and HIA, compared to other baseline methods.
Taken together, these results highlight the effectiveness of \methodopt as an optimization method targeting desired ADMET objectives.

\begin{table}[H]
	\centering
	\caption{Comparison of ADMET property scores on \dataset}
	\label{tbl:overall_results_admet}
\begin{threeparttable}
\resizebox{0.60\textwidth}{!}{
\begin{minipage}{\textwidth}
	\begin{tabular}{
		@{\hspace{2pt}}l@{\hspace{2pt}}
		@{\hspace{2pt}}r@{\hspace{2pt}}
		@{\hspace{2pt}}r@{\hspace{2pt}}
		@{\hspace{4pt}}r@{\hspace{4pt}}
		@{\hspace{2pt}}r@{\hspace{2pt}}
		@{\hspace{2pt}}r@{\hspace{2pt}}
		@{\hspace{5pt}}r@{\hspace{5pt}}
		@{\hspace{2pt}}r@{\hspace{2pt}}
		@{\hspace{2pt}}r@{\hspace{2pt}}
		@{\hspace{5pt}}r@{\hspace{5pt}}
		@{\hspace{2pt}}r@{\hspace{2pt}}
	    @{\hspace{2pt}}r@{\hspace{2pt}}
		@{\hspace{5pt}}r@{\hspace{5pt}}
		@{\hspace{2pt}}r@{\hspace{2pt}}
	    @{\hspace{2pt}}r@{\hspace{2pt}}
	}
	\toprule
	\multirow{2}{*}{Method} 
	& \multicolumn{2}{c}{Carci$\downarrow$} 
	& & \multicolumn{2}{c}{Ames$\downarrow$} 
	& & \multicolumn{2}{c}{hERG$\downarrow$} 
	& & \multicolumn{2}{c}{BBBP$\uparrow$}
	& & \multicolumn{2}{c}{HIA$\uparrow$} 
	\\
	\cmidrule{2-3}\cmidrule{5-6}\cmidrule{8-9}\cmidrule{11-12}\cmidrule{14-15}
	& Avg. & Med. &  & Avg. & Med. &  & Avg. & Med. & & Avg. & Med. & & Avg. & Med. \\
	\midrule
	Reference & 0.19 & 0.14 && 0.33 & 0.28 && 0.44 & 0.38 && 0.66 & 0.70 && 0.73 & 0.98\\
	\midrule
	\AR & \underline{0.14} & \underline{0.07} && 0.48 & 0.47 && 0.31 & 0.21 && 0.70 & 0.77 && 0.88 & \underline{0.99} \\
	\pockettwomol & 0.18 & 0.13 && 0.42 & 0.40 && \underline{0.22} & \underline{0.11} && 0.76 & 0.84 && 0.94 & \textbf{1.00} \\
	\diffsbdd & 0.18 & 0.12 && 0.46 & 0.43 && 0.30 & 0.21 && 0.71 & 0.77 && 0.91 & \textbf{1.00}\\
	\targetdiff & 0.20 & 0.15 && 0.36 & 0.29 && 0.35 & 0.26 && 0.68 & 0.72 && 0.92 & \textbf{1.00} \\
	\decompdiff & 0.20 & 0.15 && 0.34 & 0.27 && 0.38 & 0.28 && 0.70 & 0.75 && 0.82 & \underline{0.99}\\
	\decompopt & 0.20 & 0.15 && 0.35 & 0.29 && 0.37 & 0.26 && 0.69 & 0.75 && 0.82 & \textbf{1.00} \\
	\ipdiff & 0.22 & 0.18 && \underline{0.24} & \underline{0.16} && 0.43 & 0.46 && \underline{0.80} & \underline{0.86} && \underline{0.97} & \textbf{1.00}\\
	\midrule
	\method    & 0.22 & 0.22 && 0.37 & 0.38 && 0.43 & {0.44} && 0.64 & 0.62 && 0.93 & {0.98}\\
	\methodopt & \textbf{0.06} & \textbf{0.06} && \textbf{0.10} & \textbf{0.09} && \textbf{0.15} & \textbf{0.12} && \textbf{0.89} & \textbf{0.90} && \textbf{0.99} & \textbf{1.00}\\
	\bottomrule
	\end{tabular}

	\begin{tablenotes}[flushleft]
	\footnotesize
	\item \hspace{-5pt} Columns represent: ``Carci'': the carcinogenicity;
    ``Ames'': Ames mutagenicity;
    ``hERG'': hERG inhibition;
    ``BBBP'': blood-brain barrier permeability;
    ``HIA'': human intestinal absorption. The best values are in \textbf{bold}, and the second-best values are \underline{underlined}. $\uparrow$ / $\downarrow$ indicate that higher / lower values are better. 
	\end{tablenotes}
\end{minipage}
}
\end{threeparttable}
\vspace{-10pt}
\end{table}

\begin{figure}[htbp]
  \centering
  \hspace{-8pt}
\includegraphics[width=\textwidth]{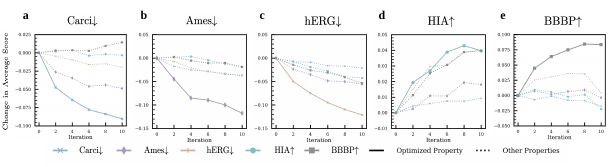}
\hfill
\vspace{-8pt}
  \caption{Changes in ADMET property scores over \methodopt optimization iterations. In each subfigure, the solid line indicates the property being optimized, while the dashed lines indicate the other properties not targeted for optimization. $\mathbf{a}$-$\mathbf{e}$ show the trends when optimizing Carci, Ames, hERG, HIA, and BBBP, respectively}.
  \label{fig:opt_trend}
\end{figure}

\paragraph{Change in ADMET Properties over Optimization Iterations}
To gain deeper insight into how the optimized as well as other non-optimized ADMET properties are affected during the optimization process, in Figure~\ref{fig:opt_trend}, we visualize the changes in ADMET property scores of \methodopt generated ligands across optimization iterations. 
Each property in the corresponding subplot is optimized independently, while the trajectories of non-optimized properties are tracked in parallel. For optimized properties, the scores consistently shift in the intended direction of improvement as the number of iterations increases -- carcinogenicity, Ames, and hERG scores gradually decline, 
while HIA and BBBP scores steadily increase. In contrast, the trajectories of non-optimized properties remain largely stable, indicating that, in \methodopt, optimizing one property does not substantially compromise other properties. Nevertheless, we observe mild tradeoffs across certain properties. For instance, reducing hERG tends to slightly lower BBBP and HIA scores, and conversely, improving absorption-related properties (HIA, BBBP) can increase hERG. 
This coupling reflects common physicochemical and structural factors underlying these ADMET properties.
A molecule's ability to permeate intestinal and blood-brain barriers is often influenced by its lipophilicity, polarity, and molecular size~\cite{weiss2024balanced}.
However, these same factors also impact a molecule's ability to bind the hERG potassium channel.
This highlights a key challenge in lead optimization: each structural modification has myriad effects that must be carefully monitored and controlled.
Disentangling such effects remains a difficult problem, which we leave for future exploration, possibly through multi-objective optimization.

\subsection*{Comparison on Molecular Structures of Generated Ligands}

\begin{table}[H]
    \centering
    \caption{Comparison on molecular structures of generated ligands on \dataset}
    \label{tbl:overall_docking_results_quality_10}
    \begin{threeparttable}
    \resizebox{\textwidth}{!}{
        \begin{tabular}{
            l@{\hspace{8pt}}
            l@{\hspace{2pt}}
            c
            c
            c
            c
            c
            c
            c
            c
            c
            c
        }
        \toprule
        Group & Metric & & \AR & \pockettwomol & {\diffsbdd} & \targetdiff & \decompdiff & {\ipdiff} & \decompopt & \method & \methodopt \\
        \midrule
        \multirow{2}{*}{Stability}
        & Atom stability ($\uparrow$) & & 0.922 & 0.871 & 0.889 & \textbf{0.949} & 0.906 & 0.939 & 0.916 & {0.935} & \underline{0.937}\\
        & Molecule stability ($\uparrow$) & & \textbf{0.449} & 0.171 & 0.220 & \underline{0.389} & 0.235 & 0.356 & 0.267 & 0.330 & 0.349 \\
        \midrule
        \multirow{3}{*}{3D structures} 
        & JS. bond lengths ($\downarrow$) & & 0.463 & 0.437 & 0.329 & 0.336 & 0.298 & 0.481 & \underline{0.285} & \textbf{0.272} & \textbf{0.272} \\
        & JS. bond angles ($\downarrow$) & & 0.372 & 0.244 & 0.298 & 0.234 & \underline{0.164} & 0.320 & \textbf{0.159} & 0.187 & 0.190 \\
        & JS. dihedral angles ($\downarrow$) & & 0.408 & 0.233 & 0.290 & 0.258 & 0.202 & 0.325 & {0.199} & \textbf{0.146} & \underline{0.147} \\
        \bottomrule
        \end{tabular}}
        \begin{tablenotes}[flushleft]
        \footnotesize
        \vspace{0.2em}
        \item[] 
        \begin{minipage}{\textwidth}
        Rows represent: 
        “atom stability”: the proportion of stable atoms that have the correct valency; 
        “molecule stability”: the proportion of generated ligands with all atoms stable;
        “JS. bond lengths/bond angles/dihedral angles”: the Jensen–Shannon (JS) divergences of bond lengths, bond angles, and dihedral angles between generated ligands and training ligands.
        The best values are in \textbf{bold}, and the second-best values are \underline{underlined}.
        $\uparrow$ / $\downarrow$ indicate that higher / lower values are better.
        \end{minipage}
        \end{tablenotes}
        
    \end{threeparttable}
\end{table}

Table~\ref{tbl:overall_docking_results_quality_10} presents stability and 3D structure metrics for generated ligands,
evaluating structural plausibility, such as proper atomic valences and realistic bond angles and dihedral angles geometry.
Notably, {\method} and {\methodopt} achieve atom-stability scores of 0.935 and 0.937, respectively,
indicating strong adherence to correct valency at the atom level.
The \method and \methodopt methods also deliver 
competitive molecule stability,
suggesting their ligands are chemically plausible.
Interestingly, \methodopt introduces additional gains over \method on both atom and molecule stability,
suggesting the guidance from ADMET properties also biases generation towards more realistic structures with fewer valence violations and lower bond strain.

Table~\ref{tbl:overall_docking_results_quality_10} also presents Jensen-Shannon (JS) divergences~\cite{lin2002divergence} between the generated ligands and 
the training data ligands for bond lengths, bond angles, and dihedral angles.
The \method and \methodopt methods generate ligands with the lowest divergences from the training ligand distribution for bond lengths and dihedral angles, indicating both methods can generate molecules with realistic chemical structures.
Furthermore, the bond angle distributions for \methodopt and \method are only slightly more divergent from the training distribution than the best baseline, \decompopt.
Overall, \method and \methodopt demonstrate a strong ability to generate ligands with high-quality chemical structures, mimicking the real-world structural qualities of the training ligands.

\begin{figure}[h]
    \centering
    \includegraphics[width=0.8\textwidth]{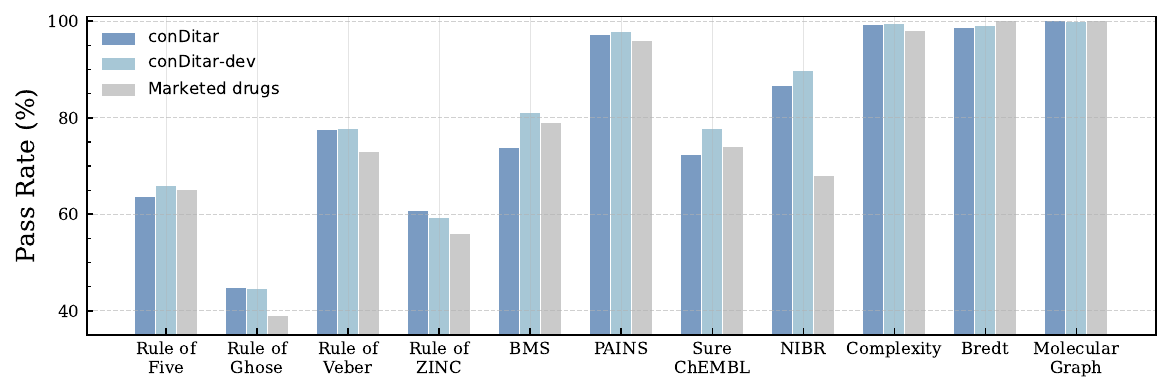}
    \vspace{-5pt}
    \caption{Comparison on filter pass rates of generated ligands to marketed drugs from ChemBL~\cite{gaulton2012chembl}.
    Rule of Five, Rule of Ghose, Rule of Veber, and Rule of ZINC are common drug-likeness rules,
    covering physicochemical properties, lipophilicity, hydrogen-bonding capacity, molecular size, flexibility, and bioavailability-related characteristics;
    BMS, PAINS, SureChEMBL, and NIBR are
    pharmaceutical structural alerts, covering undesirable, reactive, promiscuous, or assay-interfering substructures;
    Complexity, Bredt, and Molecular Graph are
    cheminformatics validity tests, covering molecular complexity, strained structural motifs, and chemically implausible or unstable molecular graphs.}
    \label{fig:filterpassingrate}
\end{figure}

Figure~\ref{fig:filterpassingrate} 
presents the performance of generated ligands from \method and \methodopt on several drug discovery filters,
applying common drug-likeness rules
(Rule of Five~\cite{lipinski2004lead}, Rule of Ghose~\cite{ghose1999knowledge}, Rule of Veber~\cite{veber2002molecular}, Rule of ZINC~\cite{irwin2005zinc,sterling2015zinc,irwin2020zinc20}),
pharmaceutical structural alerts (BMS~\cite{pearce2006empirical}, PAINS~\cite{baell2010new}, SureChEMBL~\cite{papadatos2016surechembl}, NIBR~\cite{schuffenhauer2020evolution}),
and cheminformatics validity tests (Complexity~\cite{bertz1981first}, Bredt~\cite{kobrich1973bredt}, Molecular Graph Validity~\cite{polykovskiy2020molecular}).
We select these filters as representative examples of commonly-used filters in conventional drug discovery~\cite{mignani2018present,kralj2023molecular},
removing undesirable chemical moieties and capturing a variety of perspectives on the properties of drug-like molecules collected by chemists over decades of research.
Please note, these filters are not absolute determinants of success in drug discovery,
but provide intuitive guidelines that reflect the real-world experience of medicinal chemists.
Filter pass rates of marketed drugs are also reported, providing a fair external benchmark for general medicinal chemistry quality.
Note that these filter pass rates differ from the property metrics of Table~\ref{tbl:overall_docking_results_quality_10}: 
Filter pass rates estimate rule-based drug-likeness and screen for known alert substructures, whereas Table~\ref{tbl:overall_docking_results_quality_10} verifies the chemical validity of molecules independent of drug-likeness requirements.

Ligands generated by \method and \methodopt pass each filter at rates competitive with or higher than marketed drugs,
indicating that they mirror the properties of approved drugs.
This suggests the generated ligands are largely compatible with conventional drug discovery wisdom, as reflected by these filters and rules.
Notably, \methodopt improves the pass rates over \method on pharmaceutical structural alerts (BMS, PAINS, SureChEMBL, NIBR),
demonstrating its capability to reduce undesirable toxicophoric and reactive motifs through ADMET-oriented optimization.
Overall, these filters highlight the promising nature of ligands generated from \method, while showing that \methodopt can generally improve or maintain this performance,
avoiding
undesirable chemical moieties without harming the overall quality of generated ligands.


\subsection*{Analysis of Atom and Bond Composition on \dataset}
\label{sec: distribution_analysis}

\begin{figure}[h]
\centering
\includegraphics[width=0.8\linewidth]{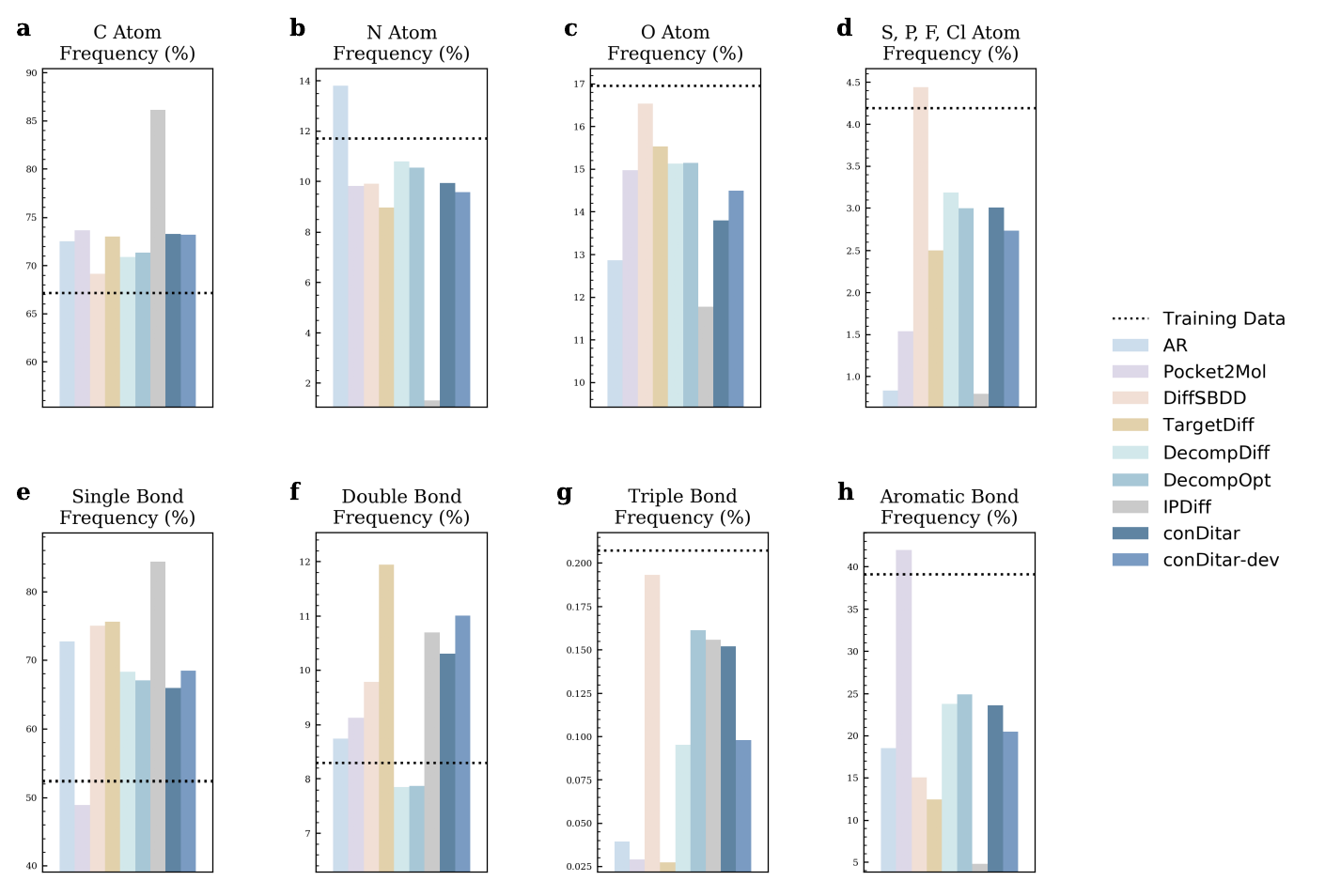}
   \caption{Comparison of atom (\textbf{a}-\textbf{d} for C, N, O, and others (F, S, Cl, P), respectively) and bond (\textbf{e}-\textbf{h} for Single, Double, Trible, Aromatic, respectively) type distributions across different models.
    The dashed line displays the percentages in the training data as a realistic reference for drug molecules.}
    \label{fig:category_analysis}
\end{figure}

To evaluate the capability of the models in generating realistic molecular structures, 
it is essential to analyze the composition of atoms and bonds in generated ligands for chemical feasibility.
As shown in Figure~\ref{fig:category_analysis}, in general, 
\method and \methodopt produce atom and bond compositions that are aligned with the reference distribution in training data of \CD.
Compared with the reference, all models show a slight overproduction of carbon.
This occurs because carbon dominates the training data (67.2\%), so the models trained on it tend to sample it more frequently.
Consequently, other atom types are slightly under sampled.
This pattern also appears in the distribution of bond types, where single bonds occur more often than in the training data (52.4\%).
Among all bond types, aromatic rings require the most strict topological and geometric constraints.
Diffusion-based models jointly denoise atom positions and types by focusing on local rather than global structures,
which makes them prone to losing fine-grained substructures, such as aromatic rings.
In contrast, \pockettwomol predicts atoms and bonds sequentially conditioned on previously sampled structure.
The bond prediction and the dependencies on previously generated structures allow it to better capture aromatic patterns.

\ipdiff achieves the best predicted binding affinity scores, as reported in Table \ref{tbl:overall_results_crossdock} and Table \ref{tbl:overall_results_curated}.
Its high affinity scores may stem from generating carbon-heavy, single-bond-dominated structures with few heteroatoms and limited aromaticity, as Figure~\ref{fig:category_analysis} shows.
Excessive carbons in the ligand scaffold render the ligand broadly hydrophobic and prone to bind the pocket in a non-specific way.
Although this may yield high Vina scores, 
it does not indicate a high intrinsic binding affinity.
The hydrophobic behavior also implies high lipophilicity, which often leads to poor ADMET properties and a higher risk of off-target interactions.
At the same time, the scarcity of heteroatoms further lowers solubility and selectivity.
Therefore, although the ligands generated by \ipdiff achieve high Vina scores, 
they are less attractive as drug candidates.

\subsection*{Case Studies} 
%

To demonstrate the practical utility of our methods, we use the 
programmed death-ligand 1 (\PDLone) and the colony-stimulating factor-1 receptor 
(\CSFoneR) targets for case studies.
\PDLone is a critical target for cancer immunotherapy due to its pivotal role in immune evasion by tumors~\cite{mandal2025overcoming}.
Current anti-\PDLone therapies are being hampered by several clinical limitations, such as 
modest efficacy and resistance~\cite{mandal2025overcoming}. 
Several high-resolution crystal structures of \PDLone-ligand complexes are publicly available in the PDB database~\cite{berman2000protein,Guzik2017}. 
\CSFoneR is a biologically important kinase involved in macrophage survival, proliferation, and differentiation~\cite{wen2023csf1r, cannarile2017colony}. 
Current \CSFoneR inhibitors on the market, for example, pexidartinib and vimseltinib, 
have emerged as safe and efficacious options for the treatment of tenosynovial giant cell tumor (TGCT), 
a non-malignant tumor of the joint, tendon sheath, or bursa driven by the overexpression of CSF1. 
However, approved \CSFoneR-targeting drugs still remain limited, motivating the development of additional candidate inhibitors. 
\CSFoneR has a publicly available high-quality co-crystal structure~\cite{kane2024identification}.
%

\subsubsection{Experimental Setup}

For both \PDLone and \CSFoneR, the case study pipeline consists of two sequential and complementary stages: 
\textbf{(1)} a structure-based design stage to generate candidate ligands by \methodopt, and 
\textbf{(2)} a ligand-based design stage to expand the \methodopt-generated 
ligand set  
through analog search in existing corporate compounds deck and to prioritize promising compounds, 
thereby mimicking a hit expansion campaign as routinely done in early drug discovery.
These two stages together provide a rigorous workflow for candidate generation, screening, and selection prior to synthesis and subsequent biological testing.

\begin{figure}[!htbp]
\centering

\begin{minipage}[t]{0.43\textwidth}
\centering

\includegraphics[width=\linewidth,height=0.20\textheight,keepaspectratio]{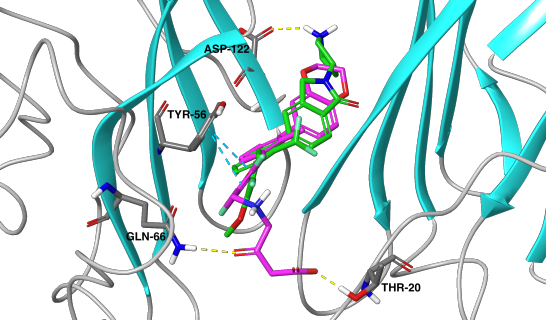}

\vspace{0.5em}

\caption{Docked poses of compound PL-0 (shown in magenta), a known cognate ligand of \PDLone, and compound PL-1 (shown in green), a \methodopt-generated novel ligand of \PDLone, suggest that both compounds exhibit similar binding modes.}
\label{fig:case_pdl1_docked}

\end{minipage}
\hspace{0.04\textwidth}
\begin{minipage}[t]{0.43\textwidth}
\centering

\includegraphics[width=\linewidth,height=0.20\textheight,keepaspectratio]{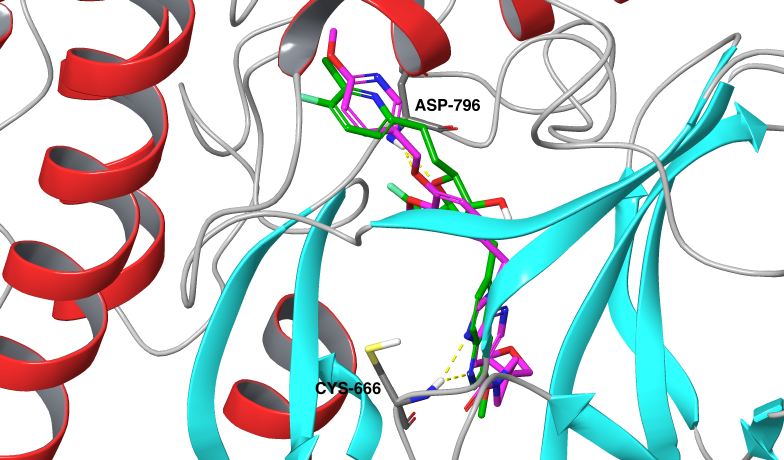}

\vspace{0.5em}

\caption{Docked poses of compound CL-0 (shown in magenta), a known cognate ligand of \CSFoneR, and compound CL-12 (shown in green), a \methodopt-generated novel ligand of \CSFoneR, suggest that both compounds exhibit similar binding modes.}
\label{fig:case_csf1r_docked}

\end{minipage}

\end{figure}

\paragraph{Structure-Based Design via \methodopt}
%
The co-crystal structures of \PDLone and \CSFoneR bound to one of their respective cognate ligands 
are retrieved from the PDB~\cite{berman2000protein} under code 5N2F~\cite{Guzik2017} and 8W1L~\cite{pdb8w1l, kane2024identification}, respectively. 
The 
targets are prepared at pH 7.4 using the protein wizard module as implemented in Maestro~\cite{maestro2026}.
Top-ranked ligands generated by \methodopt in both cases 
were further filtered to remove undesirable chemical moieties as previously described 
by Kombo \etal~\cite{kombo2024predictions}.
The remaining ligand structures are protonated at pH 7.4 and energy-minimized using Ligprep as implemented in Maestro~\cite{maestro2026}.
Flexible ligand docking was carried out using GLIDE in standard precision mode (SP), followed by extra precision mode (XP). 
%

\paragraph{Ligand-Based Design based on \methodopt Generation}
%
Following the structure-based design stage, ligand-based virtual screening of the Sanofi compound collection was carried out 
using FastROCS~\cite{grant1996fast, rush2005shape, hawkins2007comparison}, a GPU-accelerated tool for shape similarity search.
Ligands generated by \methodopt are used to build the shape query.
To further prioritize the identified virtual hits,
machine learning (ML) models developed using structure-activity relationship (SAR) datasets experimentally-determined at Sanofi~\cite{kane2024identification} are used to predict potency against \PDLone and \CSFoneR, respectively~\cite{kombo2024predictions}.
Derived analogs
are further scored and ranked using a weighted Pareto multi-parameter optimization approach combining docking, 
ML-predicted potency, and 3D similarity in shape and chemical features.
%

\subsubsection{Docking Results}
Binding modes of reference compounds, \methodopt-generated molecules, and their analogs are predicted by docking using GLIDE~\cite{maestro2026}.
Figure~\ref{fig:case_pdl1_docked} and~\ref{fig:case_csf1r_docked} show examples for \PDLone and \CSFoneR, respectively. 
In the case of \PDLone, the terminal phenyl group of the quintessential bi-phenyl moiety in both the reference ligand and the ligand generated by 
\methodopt
make $\pi$-$\pi$ interactions with TYR-56. 
The other end of both ligands interacts with ASP-122, one of the key residues driving molecular recognition at the \PDLone dimer interface.
In the case of \CSFoneR, both reference and 
\methodopt-generated ligand make hydrogen-bond interactions with CYS-666 and ASP-796, which constitute a hallmark of kinase inhibitors of this class. 
These examples indicate that \methodopt-generated ligands and their corresponding reference ligands have similar binding modes, further emphasizing the ability of
\methodopt
to learn from the binding site characteristics and 
accordingly generate high-quality binding ligands.

\subsubsection{Biological Testing Results}
\label{sec:results:case:experiments}

\begin{figure}[htbp]
    \centering
    \setlength{\tabcolsep}{2pt}
    \renewcommand{\arraystretch}{0.9}

    \newcommand{\molfig}[1]{%
        \begin{minipage}[c]{0.19\textwidth}
            \centering
            \includegraphics[width=\linewidth,height=0.09\textheight,keepaspectratio]
            {P2Diff_figures/two_cases/PDL1_structures/#1}
            
            \vspace{-2pt}
            {\scriptsize\textbf{#1}}
        \end{minipage}
    }

    \begin{tabular}{ccccc}
        \molfig{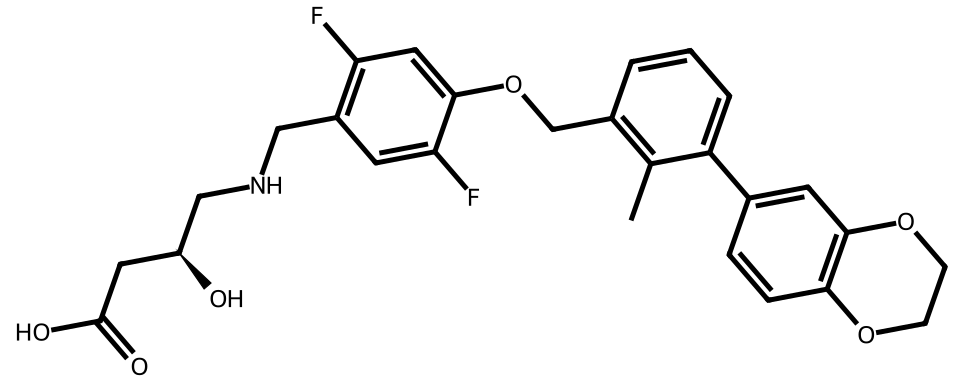} & \molfig{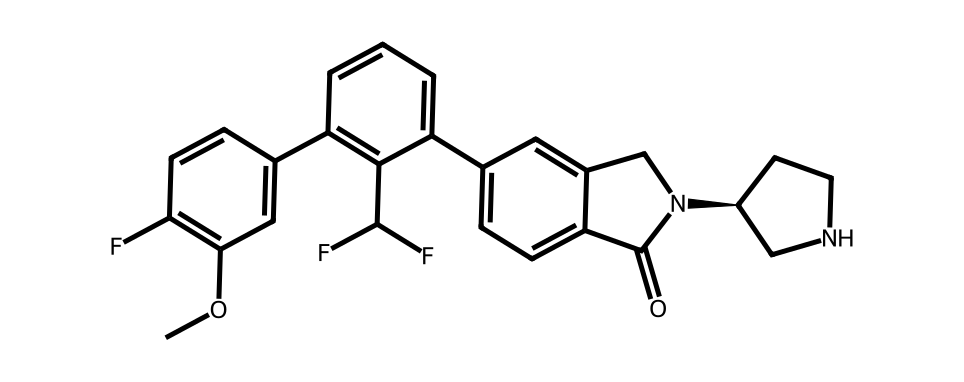} & \molfig{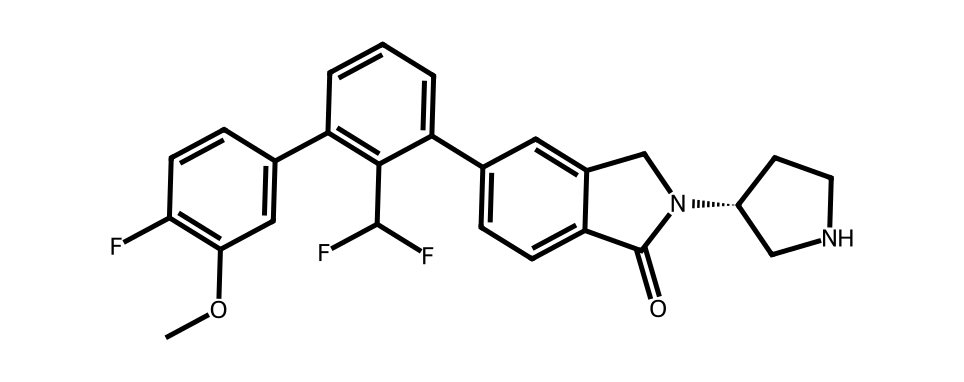} & \molfig{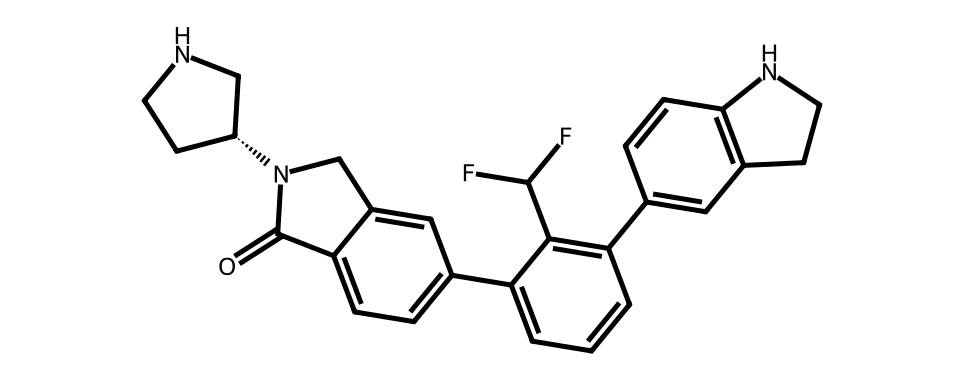} & \molfig{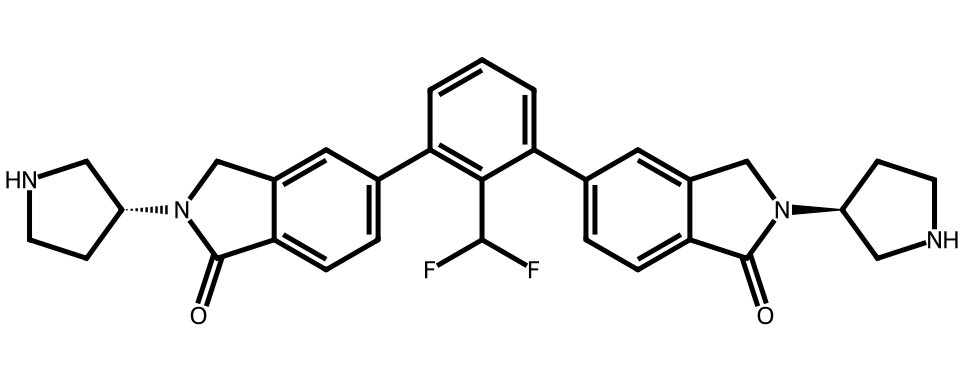}
    \end{tabular}

    \caption{\PDLone ligand structures. Each structure is labeled by its ID.}
    \label{fig:pdl1_structure}
\end{figure}
\begin{table}[ht]
\centering
\begin{threeparttable}
\caption{SPR-derived binding affinity and molecular properties of 
\methodopt-generated
\PDLone ligands and their analogs.}
\label{tab:pdl1_ligands}
\begin{tabular}{llrrrr}
\toprule
ID & Source & {K$_{\text{D}}$}($\mu$M)$\downarrow$ & LogD & QED$\uparrow$ & LipE$\uparrow$ \\
\midrule
{PL-1} &  \methodopt           & 3.49 & 3.01 & 0.65 & 2.45 \\
{PL-2} &  \methodopt           & 3.75 & 3.01 & 0.65 & 2.42 \\
{PL-3} &  \methodopt analog (modified)  & 
{4.00} & 2.70 & 0.64 & 2.70 \\
{PL-4} &  \methodopt analog (modified)  & 
{12.30} & 0.33 & 0.50 & 4.58 \\
{PL-0} & PDB: 5N2F        & ND   & 3.04 & 0.43 & ND   \\
\bottomrule
\end{tabular}
\begin{tablenotes}[flushleft]
\item \!\!\footnotesize{Columns represent: 
``Source'': the method used to obtain the molecule structure -- 
``modified'' indicates the analog was manually modified from \methodopt-generated molecules; 
``K$_{\text{D}}$'': SPR-derived equilibrium dissociation constant in $\mu$M; 
``LogD'': distribution coefficient, with a range between 1-3 generally considered optimal in drug discovery; 
``QED'': quantitative estimate of drug-likeness; 
``LipE'': lipophilic efficiency; 
$\uparrow$ / $\downarrow$ indicates that higher / lower values are better.}
\end{tablenotes}
\end{threeparttable}
\end{table}

\begin{figure}[htbp]
    \centering
    \setlength{\tabcolsep}{2pt}
    \renewcommand{\arraystretch}{0.9}

    \newcommand{\molfig}[1]{%
        \begin{minipage}[c]{0.19\textwidth}
            \centering

            \begin{minipage}[c][0.090\textheight][c]{\linewidth}
                \centering
                \includegraphics[
                    width=\linewidth,
                    height=0.085\textheight,
                    keepaspectratio
                ]{P2Diff_figures/two_cases/CSF1R_structures/#1}
            \end{minipage}

            \begin{minipage}[c][0.022\textheight][t]{\linewidth}
                \centering
                {\scriptsize\textbf{#1}}
            \end{minipage}
        \end{minipage}
    }

    \begin{tabular}{ccccc}
        \molfig{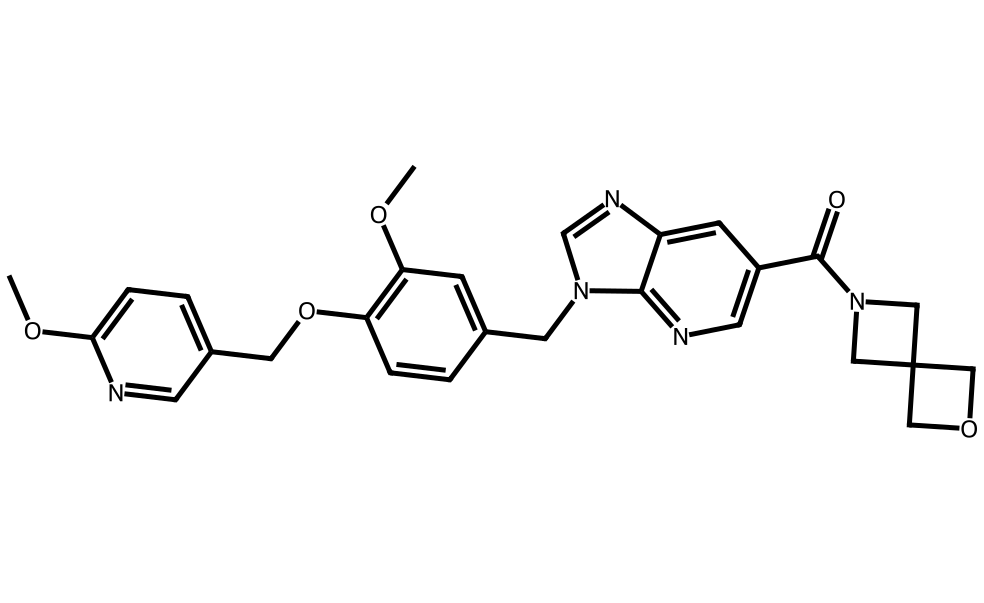}  & \molfig{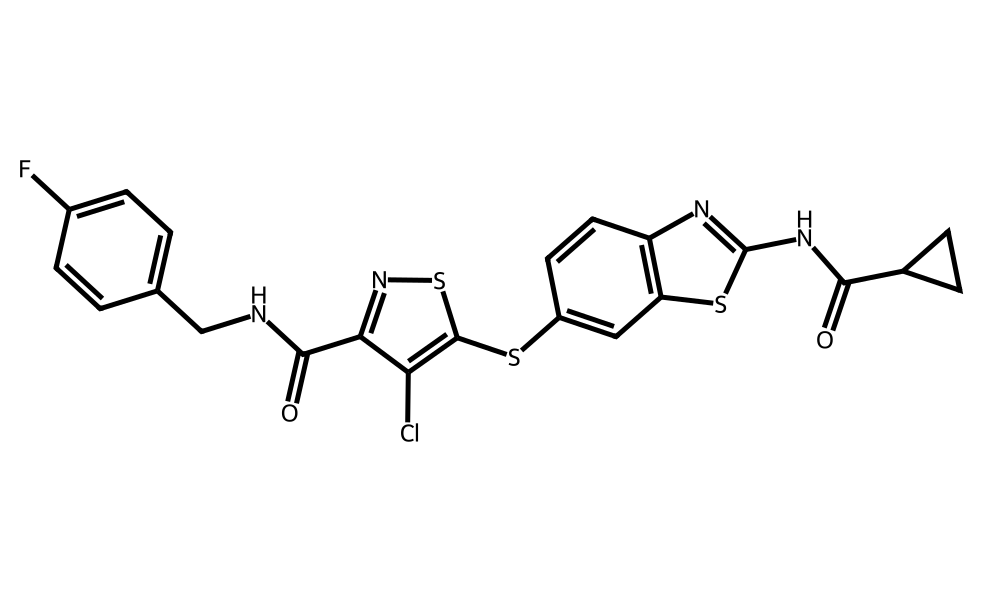}  & \molfig{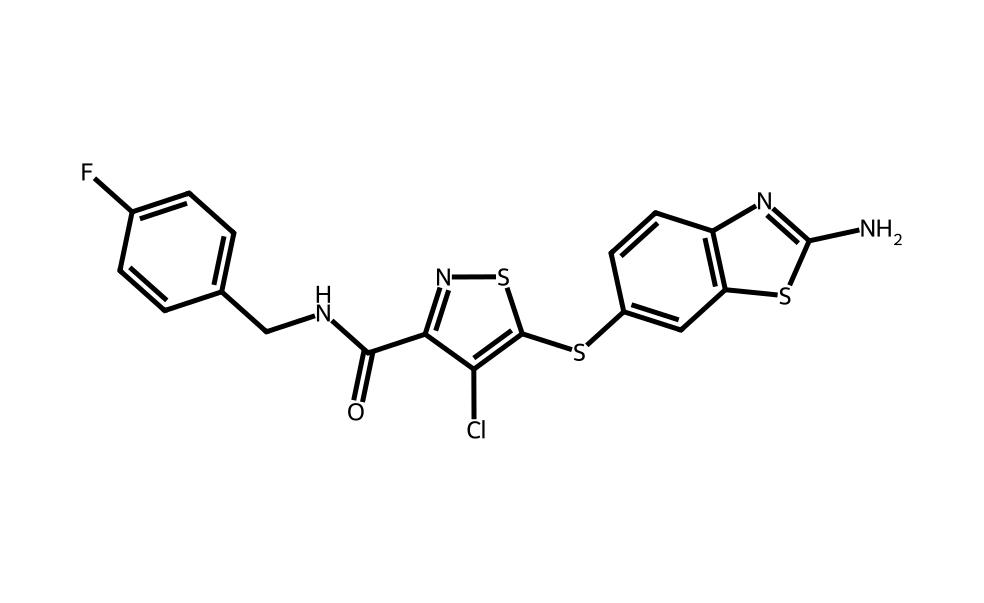}  & \molfig{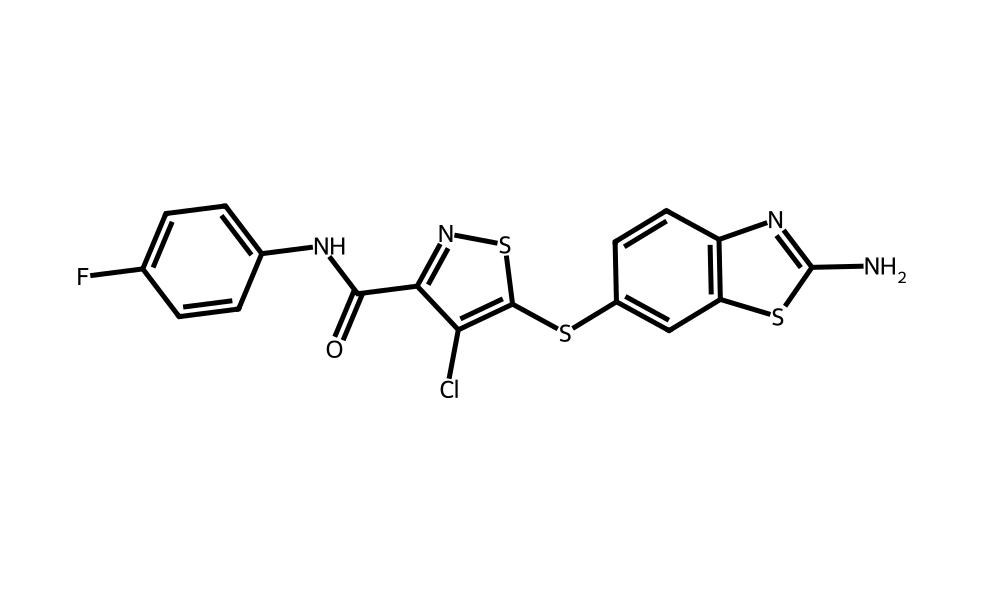}  & \molfig{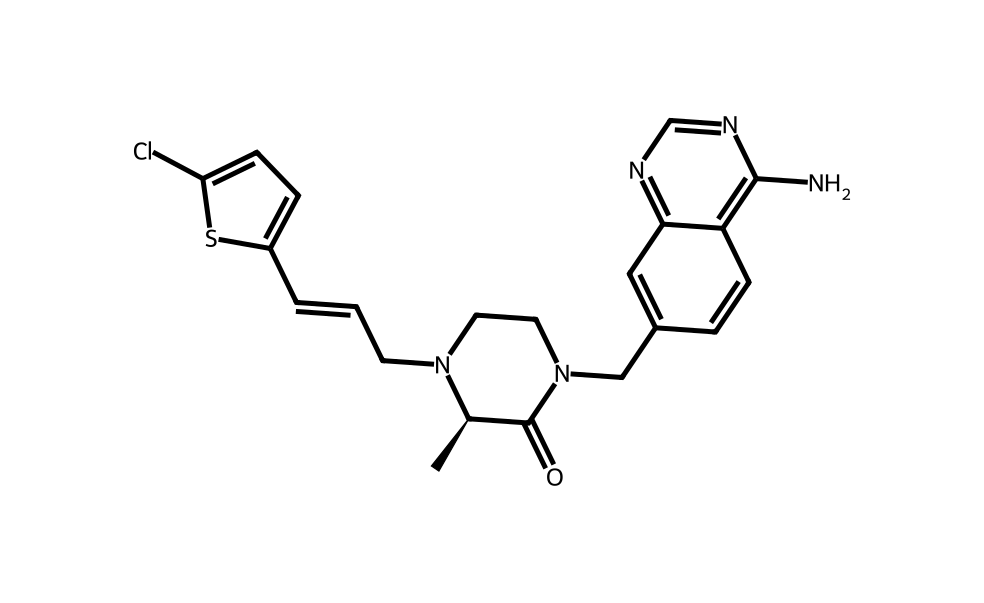}  \\
        \molfig{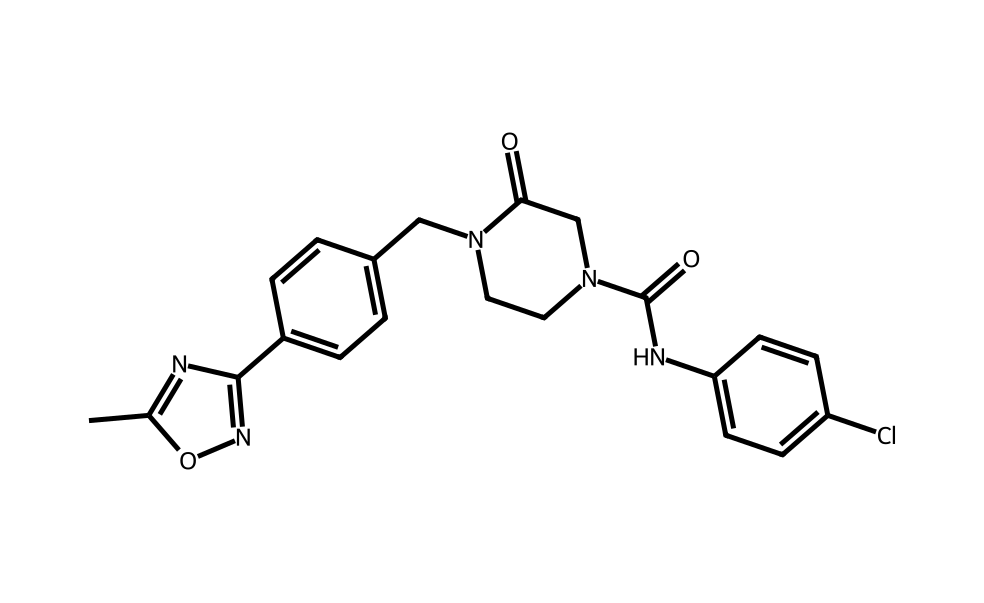}  & \molfig{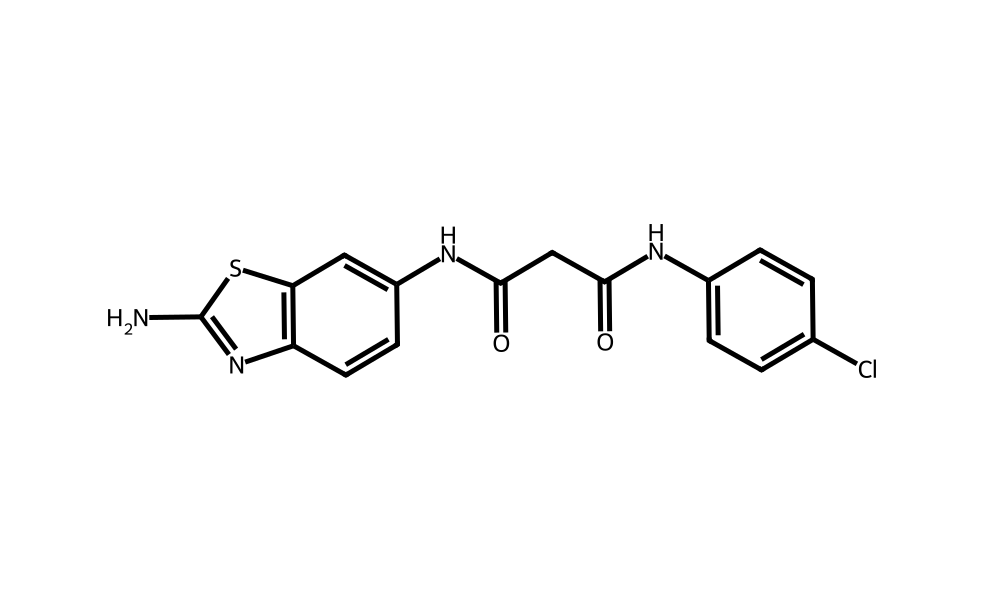}  & \molfig{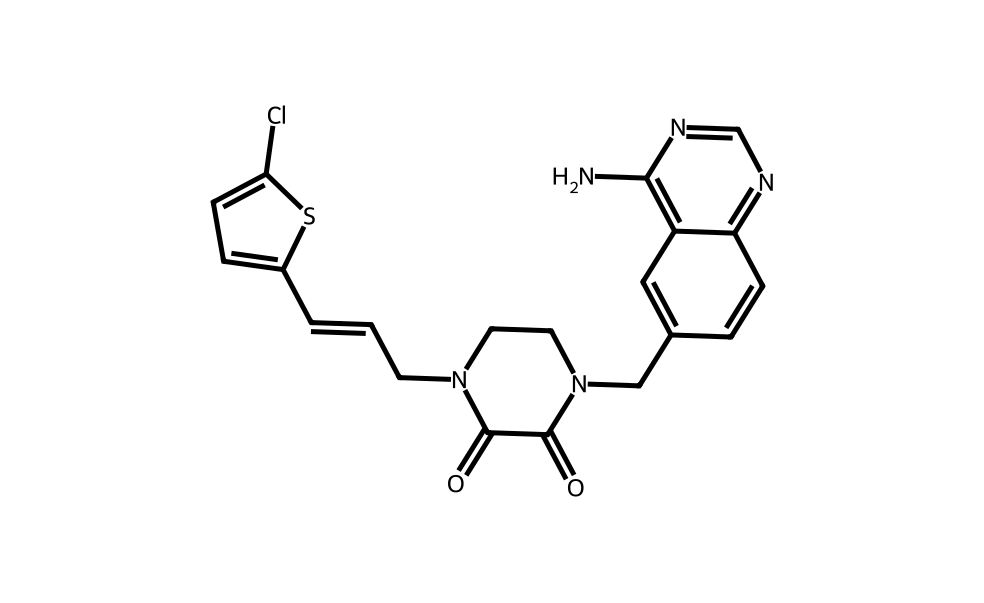}  & \molfig{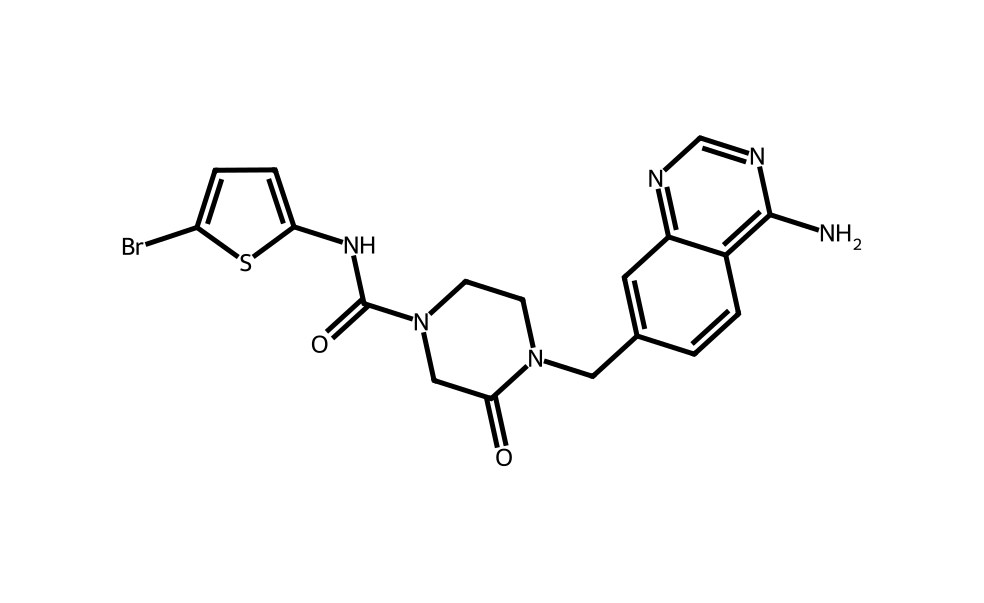}  & \molfig{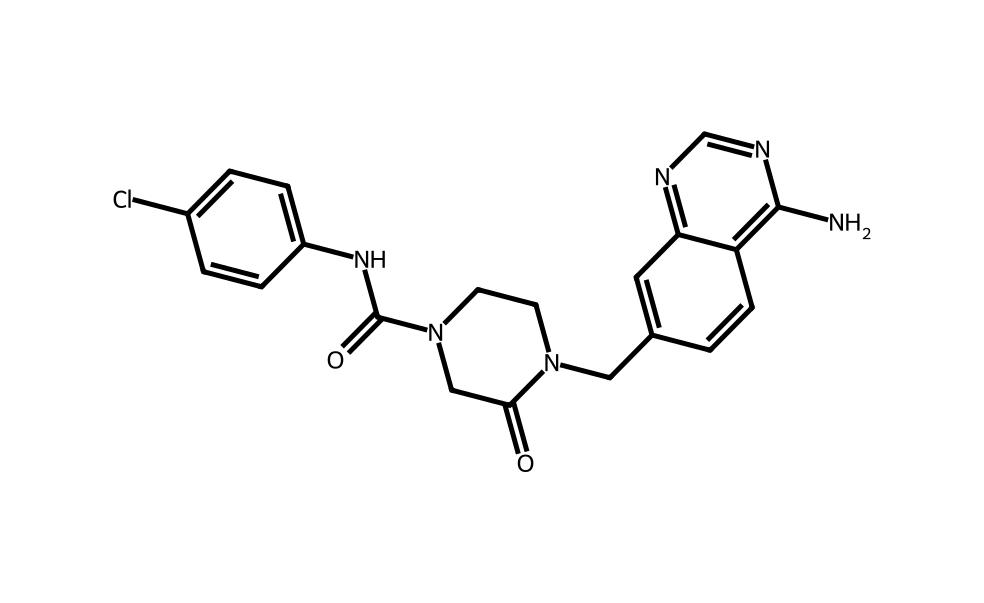}  \\
        \molfig{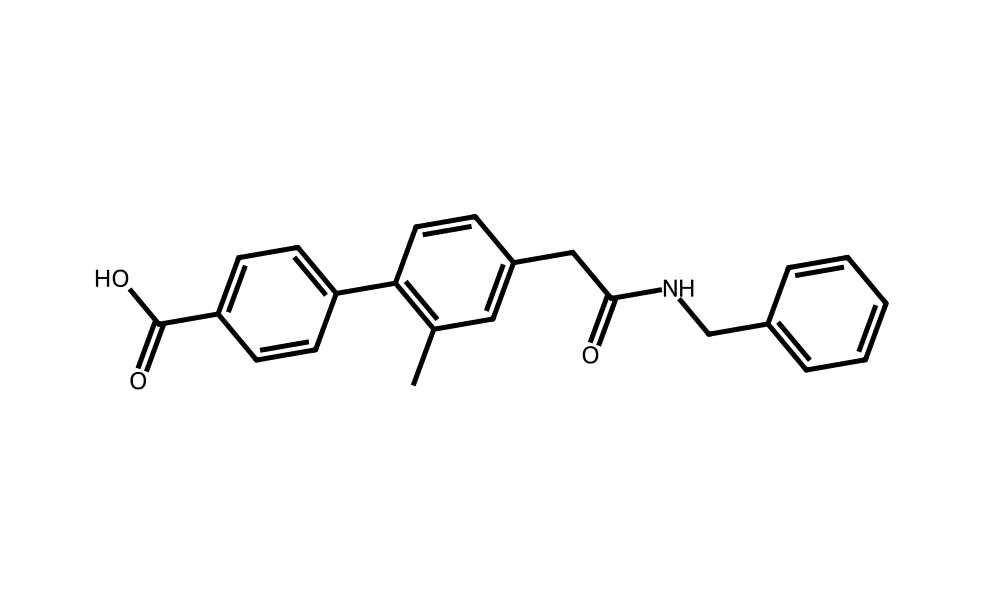} & \molfig{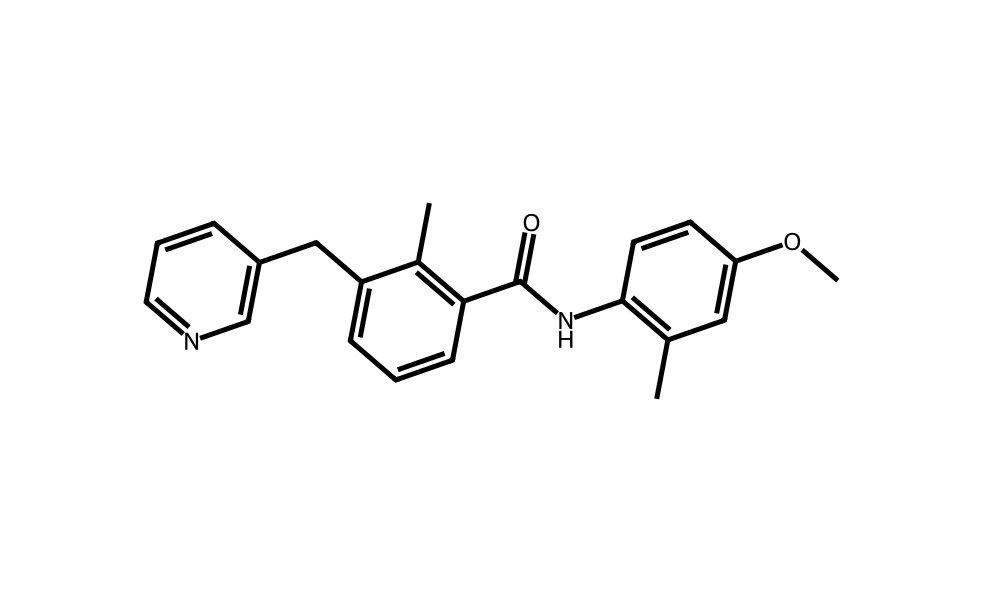} & \molfig{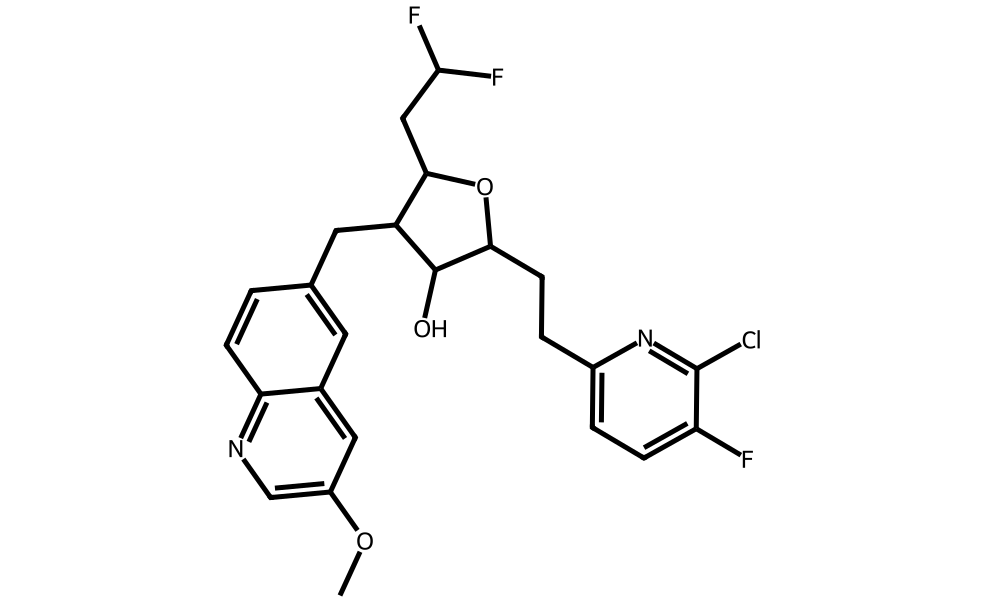} &                 &                
    \end{tabular}

    \caption{\CSFoneR ligand structures. Each structure is labeled by its ID.}
    \label{fig:csf1r_structure}
\end{figure}
\begin{table}[ht]
\centering
\small
\begin{threeparttable}
\caption{Kinase activity, molecular properties, and promiscuity profiling of \methodopt-generated \CSFoneR molecules and their analogs.}
\label{tab:csf1r_promiscuity}
\begin{tabular}{
    @{\hspace{2pt}}l@{\hspace{6pt}}
    @{\hspace{2pt}}l@{\hspace{6pt}}
    rrrrrr
    >{\raggedleft\arraybackslash}p{0.12\linewidth}
}
\toprule
ID & Source & IC$_{50}$(nM)$\downarrow$ & LogD & QED$\uparrow$ & LipE$\uparrow$ 
& \makecell{\#Targets\\Assayed} & \makecell{\#Target Hits\\(IC$_{50} \le 10~\mu$M)} & \makecell{Promiscuity\\Hit Rate (\%)$\downarrow$} \\
\midrule
{CL-0}  & PDB:8W1L              & 20                  & 1.92 & 0.36 & 5.78          & 36  & 4 & 11.1 \\
{CL-1}  & \methodopt analog (VS) & 200                 & 5.06 & 0.43 & 1.64 & 91  & 1 & 1.1  \\
{CL-2}  & \methodopt analog (VS) & 309                 & 5.75 & 0.32 & 0.77 & 126 & 3 & 2.4  \\
{CL-3}  & \methodopt analog (VS) & 2,500               & 5.05 & 0.46 & 0.55          & 127 & 0 & 0    \\
{CL-4}  & \methodopt analog (VS) & 3,363               & 2.69 & 0.48 & 2.79          & 175 & 3 & 1.7  \\
{CL-5}  & \methodopt analog (VS) & 3,736               & 2.14 & 0.69 & 3.29         & 89  & 0 & 0    \\
{CL-6}  & \methodopt analog (VS) & 5,939               & 1.87 & 0.42 & 3.35          & 112 & 0 & 0    \\
{CL-7}  & \methodopt analog (VS) & 11,700              & 2.35 & 0.63 & 2.58          & 95  & 0 & 0    \\
{CL-8}  & \methodopt analog (VS) & 21,257              & 0.93 & 0.43 & 3.74          & 27  & 0 & 0    \\
{CL-9}  & \methodopt analog (VS) & $\mathord{>}30,000$ & 1.24 & 0.48 & ND          & 94  & 0 & 0    \\
{CL-10} & \methodopt             & Inactive            & 2.82 & 0.69 & ND            & --  & -- & --   \\
{CL-11} & \methodopt             & Inactive            & 4.82 & 0.75 & ND            & --  & -- & --   \\
{CL-12} & \methodopt             & ND                  & 4.66 & 0.48 & ND            & --  & -- & --   \\
\bottomrule
\end{tabular}
\begin{tablenotes}[flushleft]
\item \!\!\footnotesize{Columns represent:
``Source'': the method used to obtain the molecule structure, where ``VS'' indicates 
the analog was identified via virtual screening with \methodopt-generated molecules as shape query;
``IC$_{50}$ (nM)'': kinase activity against \CSFoneR measured in nM;
``LogD'': distribution coefficient, with a range between 1--3 generally considered optimal in drug discovery;
``QED'': quantitative estimate of drug-likeness;
``LipE'': lipophilic efficiency;
``\#Targets Assayed'': the number of distinct targets assayed in the selectivity or promiscuity panel;
``\#Target Hits'': the number of targets with IC$_{50} \leq 10~\mu$M;
``Promiscuity Hit Rate (\%)'': the percentage of assayed targets with IC$_{50} \leq 10~\mu$M;
``Inactive'' indicates no significant inhibition was observed;
``ND'' indicates not determined;
$\uparrow$ / $\downarrow$ indicates that higher / lower values are better.}
\end{tablenotes}
\end{threeparttable}
\end{table}

Four compounds generated by \methodopt (PL-1, PL-2, CL-10, CL-11), nine analogs derived from 
ligand-based virtual screening of the Sanofi Corporate compound collection (CL-1, CL-2, CL-3, CL-4, CL-5, CL-6, CL-7, CL-8, CL-9), 
and two manually modified analogs (PL-3, PL-4) have been synthesized (Appendix~\ref{supp:synthesis} presents 
the synthesis process) and 
tested in biological assays (Appendix~\ref{supp:assay}  presents the biological testing protocols). 
Lipophilic ligand efficiency (LipE) is calculated by subtracting calculated LogD 
at pH 7.4 from pIC$_{50}$ and pK$_{\text{D}}$ in the case of \CSFoneR 
and \PDLone ligands, respectively.
LogD and QED were calculated using Pipeline Pilot~\cite{pilot2020version}.
The biological activity data for \PDLone and \CSFoneR are summarized in Tables~\ref{tab:pdl1_ligands}  and~\ref{tab:csf1r_promiscuity},
respectively; 
compound structures are presented in Figures~\ref{fig:pdl1_structure} and~\ref{fig:csf1r_structure}, respectively.
Appendix~\ref{supp:ic50} presents the dose-response curves for \CSFoneR.

For \PDLone, as shown in Table~\ref{tab:pdl1_ligands}, 
the \methodopt-generated ligands (PL-1, PL-2) and one of their modified analogs (PL-3) exhibit low-micromolar 
binding affinity,
with values ranging from 3.49 to 4.00~$\mu$M
and LipE values ranging from 2.42 to 2.70.
The other modified analog, PL-4, exhibits binding at 12.30~$\mu$M with a LipE of 4.58.
It is encouraging to note that published mean LipE value for oral drugs is 4.43~\cite{hopkins2014role}, 
and LipE values increase from hit finding 
to drug candidate discovery~\cite{kombo2025logic}-- Some \methodopt-generated 
ligands already achieve ligand efficiency profiles comparable to established orally available drugs.
All four novel ligands exhibit QED values
greater than the reference ligand ({PL-0}).

As shown in Table~\ref{tab:csf1r_promiscuity}, results indicate that using
\methodopt-generated ligands
as a template for building a 3D search query to find \CSFoneR ligands
yielded two compounds, {CL-1} and {CL-2}, with nanomolar kinase activity
against \CSFoneR, with IC$_{50}$ values of 200~nM and 309~nM, respectively,
and four other compounds, {CL-3}, {CL-4}, {CL-5}, and {CL-6}, 
with low micromolar kinase activity, with IC$_{50}$ values ranging from
2.500 to 5.939~$\mu$M.
Moreover, these compounds show low kinase promiscuity across the tested panels.
Specifically, under the
10~$\mu$M threshold,
{CL-1} and {CL-2} have promiscuity hit rates of only 1.1\% and 2.4\%, respectively, which are substantially lower than the 11.1\% observed for the reference ligand ({CL-0}).
Among the other four compounds, all except {CL-4} show no detectable off-target kinase hits under the same threshold, 
while {CL-4} shows only a low promiscuity hit rate of 1.7\%.
In addition, some of the virtual screening hits are known Factor Xa inhibitors,
as shown by an expired patent~\cite{ewing2001substituted},
which can inspire leveraging the observed polypharmacology to initiate compound
repositioning studies.
Taken together, the results obtained in both target cases suggest 
that \methodopt-generated ligands and their subsequent analogs 
have developability characteristics like those found in hit finding and hit expansion campaigns 
as carried out in drug discovery projects in pharmaceutical settings. 
Furthermore, we have predicted that \methodopt-generated ligands 
and their corresponding reference ligands have similar binding modes,
further emphasizing the ability of \methodopt 
when it comes to efficiently deciphering the binding site characteristics.


\section*{Discussions and Conclusions}

In this section, we discuss the limitations of \method and \methodopt and sketch promising directions for future work. 
\subsection*{Specializing Model for High-selectivity Ligands}
An important consideration in SBDD is ligand selectivity. 
While current diffusion-based models focus primarily on optimizing binding affinity to a given target, they often ignore off-target interactions, which can lead to undesirable side effects. 
A ligand that binds to both the intended target and unintended proteins may exhibit poor selectivity, reducing its therapeutic potential and increasing the risk of adverse effects. 
To address this, future work could explore strategies for specializing generative models to generate highly selective ligands. 
One approach is negative conditioning, training the model to avoid off-targets while retaining on-target affinity. 
Alternatively, optimization techniques over on- and off-target scores can balance affinity and selectivity during generation. 
These computational strategies could potentially improve the clinical relevance of structure-based generated ligands and advance the applicability of such AI-generative models in drug discovery.

\subsection*{Generalizing Model to Multi-objective Optimization}
While our current generation-time optimization targets a single ADMET property at a time, real discovery tasks generally require simultaneous consideration of multiple ADMET endpoints, since strong affinity alone often fails to translate when even one critical ADMET criterion is not met.
A natural extension is a multi-objective formulation that integrates multiple property signals simultaneously into the optimization module.
Such a method may construct composite update directions that coordinate improvements across all target properties
or dynamically adapt the relative emphasis of each property according to which constraints are currently violated. 
This would move the current single-objective setting toward Pareto-style multi-objective optimization and enable fine-grained control over affinity and {ADMET} trade-offs while preserving the training-free and plug-and-play nature of the optimization module. 
For instance, AI-MedCraft demonstrates how an adaptive Pareto-guided strategy can coordinate multiple competing objectives~\cite{barakat2026ai}.
In practice, such flexibility could support different stages of discovery.
Tighter ADMET constraints could reduce downstream risk when optimizing near-deployable candidates, whereas looser constraints could encourage broader exploration in early ideation, followed by gradual annealing toward stricter feasibility as optimization progresses.
Another promising direction is to make the optimization uncertainty-aware. Property predictors are often noisy, susceptible to distribution shift, and potentially vulnerable to exploitation under direct optimization. To address this, the objective could incorporate risk-sensitive criteria, such as lower-confidence-bound optimization or explicit penalties on predictive uncertainty, thereby steering updates toward candidates that are not only high-scoring but also more reliable.
Together, these extensions would better align generative optimization with pharmacological and drug-discovery practice, potentially accelerating ligand optimization toward clinic-viable candidates.

\subsection*{Incorporating Chemical Synthesis Pathway into Diffusion Models}
A major practical concern in drug discovery is the synthetic accessibility of proposed ligands. 
While conDitar and conDitar-dev generate chemically valid molecules and optimize the ADMET properties, they do not explicitly account for whether the binding ligand can be realized through feasible synthetic pathways. 
As a result, some generated ligands, despite exhibiting good \emph{in silico} profile, may be difficult to synthesize. 
To address this limitation, future work could explore incorporating synthesis pathway information into diffusion models. 
One possible direction is to introduce synthesis-aware priors or guidance signals, such as retrosynthetic accessibility scores or reaction constraints, to bias generation toward synthetically accessible regions of chemical space~\cite{shen2025compositional,cremer2024pilot}.
Such a space could include commercially available chemical building blocks commonly used in chemical reactions carried out by medicinal chemists. 
Alternatively, synthetic pathways could be modeled explicitly during diffusion training, enabling the diffusion process to simultaneously simulate molecular generation and the synthesis trajectory. 
Such synthesis-aware generative frameworks could enhance the alignment between generative model outputs and practical requirements in real-world drug discovery workflows.

\subsection*{Informing Diffusion Models with Physical and Chemical Knowledge}

Diffusion-based methods may occasionally generate molecules with atypical ring systems,
achieving high binding affinity scores while raising concerns regarding chemical realism.
To address this, one
possible direction is to incorporate chemical domain knowledge and physical constraints into the generative process,
thereby constructing a chemistry- and physics-informed diffusion framework.
For example, a framework like \methodopt
could be coupled with force field-based energy minimization
or other physics-based refinement procedures to improve the structural realism of the generated ligands. 
One such force field-based method
is OPLS-AA~\cite{jorgensen1988opls, jorgensen1996development, sambasivarao2009development, doherty2017revisiting},
which is widely used in molecular modeling and simulations of interactions between organic molecules and proteins. 
Furthermore, in future work, the generated ligands and pocket residue side chains should
be properly protonated at physiological pH (7.4)
prior to energy minimization and docking to ensure physically meaningful interaction modeling.
In addition to these improvements, we envision future work incorporating a robust protocol aimed at concomitantly filtering undesirable chemical moieties
as ligands are being generated to steer the design process towards a more druggable chemical space. 

\subsection*{Conclusions}

We present \methodopt, a conditional diffusion-based framework for SBDD
capable of generating ligands with strong binding affinities and
ADMET properties for drug discovery and development.
On \dataset, \method achieves better performance than the state-of-the-art SBDD baselines,
 achieving an average Vina D score of -8.85
 kcal/mol.
On five ADMET properties,
\methodopt achieves improvements of up to 73\%
over \method while retaining similar predicted binding affinity.
Furthermore, the molecules generated by \method and \methodopt exhibit strong stability scores, realistic chemical structures, and high drug discovery filter pass rates.
Together, these results establish \methodopt as a strong framework for generating developable drug molecules.
Moreover, \methodopt-designed molecules for \CSFoneR and \PDLone, together with their derivatives and analogs, have been experimentally synthesized and biologically tested.
Several compounds show promising binding affinities in the nanomolar and low-micromolar range,
validating \methodopt beyond the typical standard for \emph{in silico} SBDD frameworks, which typically lack experimental validation~\cite{hu2025target}.
This case study demonstrates how \methodopt can be used immediately by computational and medicinal chemists,
highlighting its potential to transform drug discovery.
We envision \methodopt as a foundation for more advanced SBDD frameworks
that incorporate additional developability objectives,
including expanded ADMET properties, synthetic accessibility,
structural constraints, and ligand selectivity.

\section*{Methods}

We introduce \methodopt, a new framework for generating 3D ligands tailored to specific protein pockets, while also optimizing additional ADMET properties of the generated ligands.
As illustrated in Figure~\ref{fig:model_architecture}, \methodopt comprises three modules: 
\textbf{(1)} \PRL, a pretrained pocket embedding module, 
\textbf{(2)} \method, a diffusion-based, pocket-conditioned ligand generation module, and \textbf{(3)} \opt, an optional optimization module applied at generation time of \method.
\PRL learns to represent 3D structures and physicochemical properties of protein binding pockets 
through an equivariant encoder. 
For a given pocket, referred to as the condition pocket, 
its binding pocket representation produced by \PRL will be used in diffusion to generate new ligands. 
\method generates new, realistic ligands in 3D with high binding affinities to the condition pocket by learning from 
the pocket-ligand interactions from existing complexes via a diffusion process~\cite{ho2020ddpm}.   
To further improve other desired pharmacological properties of the generated ligands (e.g., ADMET properties~\cite{hodgson2001admet}), 
\opt is built upon \method by performing gradient-based, zero\emph{th}-order optimization 
during its generation time without retraining \method, 
guiding the diffusion trajectories toward the desired properties. 
Table~\ref{tbl:notations} presents the key notations used in the methods. 
Particularly, subscripts $_i$ and $_j$ index atoms or residues. 
Superscripts indicate the data domain: 
$^g$ for ligand atoms, 
$^p$ for pocket atoms,  
$^r$ for pocket residues,
$^c$ for interactions between ligand atoms and pocket atoms, and $^a$ for interactions between ligand atoms and pocket residues.

\begin{table}[!h]
  \caption{{Notations}
  }
  \vspace{-10pt}
  \label{tbl:notations}
  \centering
  \begin{threeparttable}
     \begin{footnotesize}
      \begin{tabular}{
	@{\hspace{3pt}}l@{\hspace{3pt}}
    @{\hspace{3pt}}l@{\hspace{3pt}}         
	}
        \toprule
        notations & meanings \\
        \midrule
        \mol & a ligand \\
        \pocketset & a pocket\\
        $\molpocketatom_i$ & the $i$-th atom in \mol/\pocketset \\
        $\residue_i$ & the $i$-th residue in \pocketset\\
        $\molpocketatompos_i$ & the position of $\molpocketatom_i$/$\residue_i$ in 3D space\\
        $\molpocketatomtype_i$ & the atom/residue type of $\molpocketatom_i$/$\residue_i$\\

        \midrule

        $\pamolscalar_i$ & scalar embedding for atom $\molpocketatom_i$ \\
        $\pamolvec_i$ & vector embedding for atom $\molpocketatom_i$ \\
        \midrule
        $\distnotation{i}{j}$ & the distance between $\molpocketatom_i$ and $\molpocketatom_j$ \\
        $\bondtype_{ij}$ & the type of bond between $\molatom_i$ and $\molatom_j$ \\

        \midrule
        $^g$ & ligand atoms \\
        $^p$ & pocket atoms \\ 
        $^r$ & pocket residues \\
        $^c$ & interactions between ligand atoms and pocket atoms \\
        $^a$ & interactions between ligand atoms and pocket residues \\

        \bottomrule
      \end{tabular}
      \end{footnotesize}
  \end{threeparttable}
\end{table}

\subsection*{Preliminaries} 
\label{sec:method:problem_definition}


Given the condition protein pocket \pocketset, \method's objective is to generate ligands 
exhibiting strong binding affinities to \pocketset with realistic 3D structures. 
\methodopt extends this objective by optimizing additional physicochemical properties. 
In this manuscript, 
we represent a ligand \mol as a set of its atoms, defined as  
\begin{equation*}
\mol=\{\molatom_1, \molatom_2,\dots, \molatom_{N} \mid \molatom_i=(\molatompos_i, \molatomtype_i)\},
\end{equation*}
where $N$ denotes the number of atoms in \mol. 
Each atom $\molatom_i$ is characterized by its 3D coordinates $\molatompos_i \in \mathbb{R}^{1 \times 3}$ and a one-hot feature vector $\molatomtype_i \in \mathbb{R}^{1 \times {d_a}}$.
This feature vector encodes both the atom type and its aromaticity. 
Here, $d_a=13$ denotes the total number of distinct atom types considered (e.g., carbon, oxygen, nitrogen; hydrogen is excluded), 
including aromatic and non-aromatic variants.
Similarly, we represent a protein pocket \pocketset using its set of protein atoms
within a predefined radius of its reference ligand (as introduced in the {section \DatasetsSection)}
denoted as \pocketatomset, 
and the residues involved in the 
protein pocket, denoted as \pocketresidueset, that is, 
\begin{equation*}
\pocketset = (\pocketatomset, \pocketresidueset), 
\end{equation*}
where 
\begin{equation}
\label{eqn:pocket_rep}
\begin{aligned}
& \pocketatomset = \{\pocketatom_1, \pocketatom_2, \cdots, \pocketatom_{N_a}|\pocketatom_i=(\pocketatompos_i, \pocketatomtype_i)\}, \\
& \pocketresidueset = \{\residue_1, \residue_2, \cdots, \residue_{N_r}|\residue_i=(\residuepos_i, \residuetype_i)\}.
\end{aligned} 
\end{equation}
{Each} pocket atom $\pocketatom_i \in A_p$ is characterized by its 3D coordinates $\pocketatompos_i \in \mathbb{R}^{1 \times 3}$ 
and a one-hot feature vector $\pocketatomtype_i \in \mathbb{R}^{1 \times {d_a}}$, representing its atom type with $d_a=4$ (carbon, nitrogen, oxygen, or sulfur).
Similarly, each pocket residue $\residue_i \in R_p$ is represented using the 3D coordinates of 
its alpha carbon atom $\residuepos_i \in \mathbb{R}^{1 \times 3}$, 
and a one-hot feature vector $\residuetype_i \in \mathbb{R}^{1 \times {20}}$, representing the 20 amino acid types.
Note that each residue can be decomposed into its component 
atoms, represented as  
\begin{equation*}
\residue_i = \{\pocketatom_{i,1}, \dots, \pocketatom_{i, |\residue_i|}\}, 
\end{equation*}
where $\pocketatom_{i,j}$ is the $j$-th atom in $\residue_i$ ($\pocketatom_{i,j} \in A_p$), and $|\residue_i|$ is the number of atoms in $\residue_i$.
{For each $\pocketatom_j \in A_p$, the residue that it is in is denoted as $\residue(\pocketatom_j)$, that is, 
$\residue(\pocketatom_{i,j}) = \residue_i$. }

\subsection*{Pocket Representation Pretraining (\PRL)}
\label{sec:method:condition_pretrain}

We pretrain a pocket embedding module, denoted as \PRL, to encode the 3D structures of target protein pockets into pocket embeddings.  
\PRL employs an encoder, \pocketenc, which maps each pocket atom $\pocketatom_i$ into two latent representations: 
a scalar embedding $\pascalar_i \in \mathbb{R}^d$ and a vector embedding $\pavec_i \in \mathbb{R}^{d \times 3}$,
where $d$ is a hyperparameter of \PRL.
Together, these latent representations capture the essential structural information of the pocket.
Simultaneously, \PRL employs a decoder,
\pocketdec, to decode pocket embeddings to atom positions and types.
Following \mbox{Uni-Mol's}~\cite{zhou2023uni} approach,
{\PRL} is optimized by reconstructing 
noise-free pockets from corrupted ones using pocket embeddings, 
where the pockets are corrupted by
adding uniform noise into the positions of randomly selected 
atoms with their atom types masked off.
Existing diffusion-based methods~\cite{guan2023targetdiff,guan2023decompdiff,huang2024protein} typically jointly learn pocket representations and ligand 
representations together from pocket-ligand interactions. 
This approach can limit the quality of pocket representation learning 
since the comparatively larger size of pockets makes them more prone to being underrepresented 
or traded off relative to ligands.
In \method, we isolate the pocket representation learning within the pretraining of the \PRL module to ensure 
expressive pocket embeddings.

\subsubsection*{Pocket Encoder (\pocketenc)}
\label{sec:method:condition_pretrain:encode}

\pocketenc learns a {scalar} embedding $\pascalar_i$ and a vector embedding $\pavec_i$ 
for each atom $\pocketatom_i$ in the pocket \pocketset to encode pocket structure and physicochemical properties.  
These two embeddings integrate the information of {each atom $\pocketatom_i \in A_p$ and the residue $\residue(\pocketatom_i) \in \pocketresidueset$ to which $\pocketatom_i$ belongs to capture multi-granular information from the pockets. 

\paragraph{Pocket Atom Embeddings}
 
\pocketenc uses a multi-layer graph attention neural network~\cite{velickovic2018graph} 
augmented with a geometric vector perceptron (GVP)~\cite{jing2021learning} to learn 
embeddings of pocket atoms. 
At each layer $l$ of the graph neural network, 
\pocketenc learns the invariant features/embeddings $\pascalar_{i,l}$ and the equivariant features/embeddings $\pavec_{i,l}$
($\pascalar_{i,0} = \pocketatomtype_i$, $\pavec_{i,0} = \pocketatompos_i$).
Invariant features capture scalar properties such as atom types and interatomic distances, which remain unchanged under rotations and translations; 
equivariant features encode vector information such as atom positions and spatial directions that are difference vectors between atom positions, which transform consistently under these transformations.
After $L$ layers, \pocketenc outputs $\pascalar_{i, L}$ and $\pavec_{i, L}$ as the final pocket atom embeddings, that is, 
\begin{equation}
\pascalar_{i} = \pascalar_{i, L}, \pavec_{i} = \pavec_{i, L}.  
\end{equation}

Specifically, at the $l$-th layer, \pocketenc updates the embeddings $\pascalar_{i,l}$ and $\pavec_{i,l}$ for atom $\pocketatom_i$ by aggregating information from its neighboring atoms as follows:

\begin{equation}
\label{eqn:pred:gvp}
\begin{aligned}
(\pascalar_{i,l}, \pavec_{i,l}) & =  \text{GVP}(\mathbf{h}^p_{i,l}, \mathcal{Y}^p_{i,l}), \text{ where }\\
\mathbf{h}^p_{i,l} & =  \biggr[\pocketatomtype_i, \pascalar_{i,l-1}, \sum_{\pocketatom_j\in \scriptsize{\pocketneighbour(\pocketatom_i|\pocketset)}}e_{ji,l}\messscalar_{ji,l}\biggr],\\
\mathcal{Y}^p_{i,l} & =  \biggl[\pocketatompos_i, \pavec_{i,l-1}, \sum_{\pocketatom_j\in \scriptsize{\pocketneighbour(\pocketatom_i|\pocketset)}}e_{ji,l}\messvec_{ji,l}\biggr],
\end{aligned}
\end{equation}
where $\text{GVP}(\cdot)$ models interactions between $\mathbf{h}^p_{i,l}$ and $\mathcal{Y}^p_{i,l}$. 
Intuitively, $\mathbf{h}^p_{i,l}$ aggregates the chemical characteristics of $\pocketatom_i$, and
$\mathcal{Y}^p_{i,l}$ aggregates its geometric features.
$\pocketneighbour(\pocketatom_i|\pocketset)$ denotes the set of the $k$-nearest pocket atoms to $\pocketatom_i$, selected based on proximity among atoms in \pocketset; 
$\messscalar_{ji,l}$ and $\messvec_{ji,l}$ correspond to the scalar and vector message embeddings, respectively, which propagate information from atom $\pocketatom_j \in \pocketneighbour(\pocketatom_i|\pocketset)$ to $\pocketatom_i$;
and
$e_{ji,l}$ denotes the attention weight between neighbor atoms $\pocketatom_j$ and $\pocketatom_i$. The message embeddings $\messscalar_{ji,l}$ and $\messvec_{ji,l}$ are calculated as follows:
\begin{equation}
\begin{aligned}
(\messscalar_{ji,l}, \messvec_{ji,l}) & = \text{GVP}(\hat{\mathbf{m}}^p_{ji,l}, \hat{M}^p_{ji,l}), \text{ where }\\
\label{eqn:message}
\hat{\mathbf{m}}^p_{ji,l} & = [\pascalar_{j,l-1},{\distp{j}{i}}], \\
\hat{M}^p_{ji,l} & = [\pavec_{j,l-1}, {\pocketatompos_j - \pocketatompos_i]},
\end{aligned}
\end{equation}
where {$\distp{j}{i}$ represents the Euclidean distance between the positions of pocket atoms
$\pocketatom_j$ and $\pocketatom_i$.} 
The attention weights $e_{ji,l}$ are calculated as follows:
\begin{equation}
\label{eqn:attention}
\begin{aligned}
e_{ji,l} & = \frac{\exp(Q_{i,l}K_{ji,l})}{\sum_{\pocketatom_k\in \scriptsize{\pocketneighbour(\pocketatom_i|\pocketset)}}\exp(Q_{i,l}K_{ki,l})}, \text{ where } \\
Q_{i,l} &= \text{MLP}({[\pascalar_{i,l-1}, \|\pavec_{i,l-1}\|_2]}),\\
K_{ji,l} &= \text{MLP}([\messscalar_{ji,l}, \|\messvec_{ji,l}\|_2]),
\end{aligned}
\end{equation}
where $\|\cdot\|_2$ denotes the
Euclidean ($\ell_2$) norm of a 2D vector. 
Intuitively, the attention weight $e_{ji,l}$ measures how much neighbor atom $\pocketatom_j$ should 
influence the update of $\pocketatom_i$ based on both chemical  and spatial information encoded 
in $Q_{i,j}$ and $K_{ji,l}$, allowing the model to pay more attention to more informative neighbors.
Here, $Q_{i,l}$ represents the query of atom $\pocketatom_i$, summarizing 
the features of atom $\pocketatom_i$ and geometry surrounding it, 
while $K_{ji,l}$ represents the key of neighbor $\pocketatom_j$, capturing its message-level 
features that combine chemical and spatial information.

%
\paragraph{Pocket Residue Embeddings}

\pocketenc also learns residue embeddings, denoted as $\prscalar_i$ and $\prvec_i$, respectively, 
using a graph neural network (GNN) similar to the one described above (Equations~\ref{eqn:pred:gvp} to \ref{eqn:attention}) over residues. 
Incorporating this residue information allows \pocketenc to capture important
physicochemical properties at the residue level,
such as polarity, charge, and hydrophobicity,
that are critical for determining protein-ligand interactions~\cite{gilson2007calculation}. 
%

\paragraph{Pocket Embeddings}

\pocketenc integrates pocket atom embeddings and pocket residue embeddings into 
a single, enriched pocket embedding, comprehensively capturing the multi-granular spatial and physicochemical features of the protein pockets.
Particularly, \pocketenc combines each pocket atom embedding and the embedding 
of its belonging residue through a gating mechanism, 
which balances the two embeddings by adaptively weighting their contributions. 
Thus, the atom scalar embedding for $\pocketatom_i$ 
is updated as follows:
\begin{equation}
\begin{aligned}
\pascalar_i &= g^s_i  \pascalar_i + (1 - g^s_i)  \prscalar_{j}, ~~\residue_j = \residue(\pocketatom_i), \text{ where }\\
g^s_i &= \sigma(\text{MLP}([\pascalar_i, \prscalar_j]),  
\end{aligned}
\label{eqn:prlenc:residue_scalar}
\end{equation}
where $g^s_i$ denotes the gating weight, learned from the scalar embeddings 
of the pocket atom $\molatom_i$ and its corresponding residue $\residue(\pocketatom_i)$. 
Similarly, the atom vector embedding for $\pocketatom_i$ is updated as follows: 
\begin{equation}
\begin{aligned}
\pavec_i &= g^h_i  \pavec_i + (1 - g^h_i)  \prvec_{j}, ~~\residue_j = \residue(\pocketatom_i), \text{ where }\\
g^h_i &= { \sigma(\text{MLP}([\|\pavec_i\|_2, \|\prvec_j\|_2]))}.
\end{aligned}
\label{eqn:prlenc:residue_vec}
\end{equation}
This gating mechanism enables the model to dynamically integrate atom-level and residue-level information, leading to expressive pocket representations over pocket atoms.

\subsubsection{Pocket Decoder (\pocketdec)}
\label{sec:method:condition_pretrain:decode}

To train \pocketenc, we introduce a pocket decoder, denoted as \pocketdec,
to facilitate self-supervised training on a contrived noise prediction task, inspired by \mbox{Uni-Mol}~\cite{zhou2023uni}.
To create supervision signals, 
we corrupt each pocket by applying two types of perturbations: 
\textbf{(1)} atom position perturbation -- uniform noises are
added to the positions of a randomly selected subset of atoms, and \textbf{(2)} atom type masking -- the same subset of pocket atoms as in the position perturbing step has their atom types masked off.
The corrupted pocket is encoded to embeddings by \pocketenc, as described in section \PESection.

\pocketdec learns to recover the original atom positions and types from the embeddings of noisy pockets provided by \pocketenc.
Specifically,
\pocketdec predicts the normal noise $\hat{\mathbf{n}}^p_i$ added to the position of a pocket atom $\pocketatom_i$ using the following formulation:
\begin{equation}
\begin{aligned}
\hat{\mathbf{n}}^p_i & = \sum_{\pocketatom_j\in \scriptsize{\pocketneighbour(\pocketatom_i|\pocketset)}} \frac{c_{ji}({\pocketatompos_j - \pocketatompos_i})}{N_i}, \text{ where }\\
\quad c_{ji} &= \text{MLP}([\pascalar_i, \pascalar_j, {\distp{j}{i}}, \|\pavec_j - \pavec_i\|_2]),
\end{aligned}
\label{eqn:prldec:pos}
\end{equation}
where $N_i := |\pocketneighbour(\pocketatom_i|\pocketset)|$, which denotes the number of neighboring pocket atoms of 
$\pocketatom_i$.
Thus, \pocketdec estimates the positional noise of each atom~$\pocketatom_i$ by aggregating the relative position vectors from its neighboring atoms. 
The weighting coefficient~$c_{ji}$ from  
each neighboring atom~$\pocketatom_j$ is computed from the embeddings $\pascalar$ and $\pavec$
produced by \pocketenc.
This formulation exploits the local geometric structure of the pocket, forcing $\pascalar$ and $\pavec$ to encode spatial relationships among neighboring atoms and reflect local geometric information.
We predict the added positional noise rather than the clean positions here since the noise follows a simpler and more tractable distribution,
which stabilizes training and is easier to learn.

For masked atom types, \pocketdec predicts a probability vector over all possible atom types for each masked atom~$\pocketatom_i$, denoted as $\hat{\mathbf{v}}^p_i$, by a multi-layer perceptron (MLP):
\begin{equation}
\hat{\mathbf{v}}^p_i = \text{softmax}(\text{MLP}(\pascalar_i)).
\label{eqn:prldec:type}
\end{equation}
This formulation enables \pocketdec to infer atom types based on the learned pocket embedding $\pascalar$.
By designing \pocketdec to reconstruct atom types and positions, we guide \pocketenc to learn pocket embeddings that capture both atomic characteristics and local geometry for accurate recovery and refinement.
%

\subsubsection{\PRL Pretraining}
\label{sec:method:condition_pretrain:pretrain}

To pretrain \PRL, following \mbox{Uni-Mol}’s denoising pretraining~\cite{zhou2023uni}, we randomly corrupt 20\% of the atoms in pocket 
\pocketset.
The original positions and atom types of the noisy pocket \pocketset are reconstructed using {\pocketdec} by minimizing the loss function:
\begin{equation}
\label{eqn:pretrainloss}
\mathcal{L}_p
:= \frac{1}{|\mathcal{M}|}\sum_{\pocketatom_i\in\scriptsize{\pocketset}}
\mathbb{I}(\pocketatom_i\in\mathcal{M})\,
\Big(\,\|\hat{\mathbf{n}}^{p}_i - \mathbf{n}^p_i\|^2
\;+\; \mathsf{H}(\hat{\mathbf{v}}^p_i, \mathbf{v}^{p}_i)\Big),
\end{equation}
where $\mathcal{M}$ is the set of perturbed atoms in \pocketset,
$\mathbb{I}(\cdot)$ denotes the indicator function,
$\mathsf{H}(\cdot, \cdot)$ represents the cross-entropy loss, and
$\hat{\mathbf{n}}^p_i$ and $\mathbf{n}^p_i$ denote the predicted noise from the decoder and the ground-truth noise from the corruption step, respectively. 
The overall pretraining process is summarized in {Algorithm~\ref{alg:see_shaperep}.}

\begin{algorithm}[!h]
    \caption{\PRL}
    \label{alg:see_shaperep}
    \textbf{Required Input: all $\pocketset$ in Training data of $\CD$}
    \begin{algorithmic}[1]
        \While{$\text{not converged}$}
            \State Sample a batch $\mathcal{B}$ of \pocketset
            \State $\text{Loss} \gets 0$
            \For{$\pocketset \in \mathcal{B}$}
                \State $\pocketatompos, \pocketatomtype, \residuepos, \residuetype$ = $\text{addNoise}(\pocketset)$ \Comment{perturb pocket structure}
                \State $\pascalar, \pavec = \pocketenc(\pocketatompos, \pocketatomtype, \residuepos, \residuetype)$ \Comment{encode the positions and features into latent embeddings}
                \State $\hat{\pocketset} = \pocketdec(\pascalar, \pavec, \pocketatompos)$ \Comment{reconstruct pocket by decoding the embeddings}
                \State $\text{Loss} \gets \text{Loss} + \mathcal{L}_p(\hat{\pocketset}, \pocketset)$ \Comment{accumulate reconstruction loss (Equation~\ref{eqn:pretrainloss})}
            \EndFor
            \State $\text{Loss} \gets \frac{1}{|\mathcal{B}|}\text{Loss}$
            \State $\text{Loss}.\text{backward(\pocketenc,\;\pocketdec)}$             \Comment{Equation~\ref{eqn:pretrainloss}}
            \State Update \pocketenc,\;\pocketdec
        \EndWhile
        \State \Return $\pocketenc$
    \end{algorithmic}
\end{algorithm}

\subsection*{Pocket-conditioned Ligand Generation via Conditional Diffusion (\method)}
\label{sec:method:condition_generation}

We present a new conditional diffusion model, \method, for the generation of three-dimensional (3D) ligands
that bind to specific, given protein pockets, referred to as condition pockets.  
The proposed model generates realistic 3D ligands with high binding affinities by: 
\textbf{(1)} incorporating learned pocket structures and interactions within the pocket from {\pocketenc}, 
\textbf{(2)} modeling atomic interactions within the ligand, and 
\textbf{(3)} learning pocket-ligand interactions.
The subsequent sections provide a detailed description
of the diffusion process (Section~\DiffSubsection),  
the training methodology of \method (Section~\TrainSubsection), 
and the pocket-conditioned ligand generator (Section~\PredictorSubsection).
%


Following the framework of denoising diffusion probabilistic models~\cite{ho2020ddpm}, 
\method comprises a forward diffusion process that progressively adds noise to the atomic positions 
$\{\molatompos_i\}$ and features $\{\molatomtype_i\}$ of ligands, 
and a reverse generative process that is trained to denoise   
and generate new ligands during inference. 
During training, the model learns to invert the forward corruption process by iteratively denoising noisy ligands.
During inference, \method first samples noisy ligand atom positions and features at step $T$ from 
predefined simple distributions and then reconstructs realistic 3D ligand structures by iteratively removing noise until $t=0$.
At each reverse step $t$, given a noisy structure $\{(\molatompos_{i,t}, \molatomtype_{i,t})\}$,  {\method} employs a GNN-based ligand generator
that uses fixed pocket embeddings from {\pocketenc}
together with pocket atom/residue positions and types to predict the noise-free atom positions 
$\{\molatompos_{i,0}\}$ and features $\{\molatomtype_{i,0}\}$ of the ligand atoms. 
Unlike other diffusion-based SBDD methods~\cite{guan2023targetdiff, guan2023decompdiff, huang2024protein, gu2024aligning}, 
{\method} \textbf{(1)} leverages pocket positions and types exclusively to model 
pocket-ligand interactions during denoising, 
while the pocket structure
remains captured in fixed embeddings produced by {\pocketenc}.
The fixed pocket representation prevents embedding drifts and re-encoding artifacts across \method's denoising process, 
so the ligand generator
can learn pocket-ligand interactions
efficiently and with greater coherence across the generation process.
Like prior SBDD works~\cite{guan2023targetdiff, guan2023decompdiff, huang2024protein, gu2024aligning},
\method \textbf{(2)} models the atomic interactions between the ligand and pocket.
In addition {to} atom-level modeling, \method further integrates residue-level interactions.
This multi-scale featured design provides the ligand with a hierarchical view of the pocket, 
enabling a comprehensive understanding of pocket-ligand interactions. 

\subsubsection*{Forward Diffusion Process (\methodforward)}

Following the previous work~\cite{guan2023targetdiff}, 
each ligand atom position is progressively noised according to a Gaussian transition in the forward process~\cite{ho2020ddpm}:  
\begin{equation}
q(\molatompos_{i,t} \mid \molatompos_{i,t-1}) 
= \mathcal{N}\!\left(
\molatompos_{i,t};
\sqrt{1-\beta^{\mathbf{x}}_t}\,\molatompos_{i,t-1}, 
\beta^{\mathbf{x}}_t \mathbf{I}
\right),
\end{equation}
where $\beta^{\mathbf{x}}_t$ is a predefined variance schedule that controls the noise magnitude added at step $t$; $\mathbf{I}$ denotes the identity matrix.
Over $T$ steps, this produces a Markov chain that gradually transforms the clean coordinates into isotropic Gaussian noise.  
Similarly, atom types are corrupted through a categorical diffusion process. 
At each step $t$, the current atom type is retained with probability $1 - \beta^{\mathbf{v}}_t$, while the remaining probability mass is uniformly distributed among the other possible types:
\begin{equation}
q(\molatomtype_{i,t}\mid \molatomtype_{i,t-1})
= \mathcal{C}\!\Big(\molatomtype_{i,t};\,
(1-\beta^{\mathbf{v}}_t)\, {\molatomtype_{i,t-1}}
\;+\; \frac{\beta^{\mathbf{v}}_t}{d_a}\mathbf 1\Big),
\end{equation}
where $\beta^{\mathbf{v}}_t$ controls the level of corruption; $\mathbf 1$ denotes the all-ones vector indicating uniform probability over all $d_a$ atom types.
%

\subsubsection*{Reverse Generative Process (\methodbackward)}

{In the reverse process, } 
\method learns to reverse the forward process and generates 
ligands by denoising from $\{(\molatompos_{i,t}, \molatomtype_{i,t})\}$ to $\{(\molatompos_{i,t-1}, \molatomtype_{i,t-1})\}$.
Following Ho~\etal~\cite{ho2020ddpm}, 
\method approximates the probability of ($\molatompos_{i,t-1}$, $\molatomtype_{i,t-1}$) denoised from ($\molatompos_{i,t}$, $\molatomtype_{i,t}$) using the posterior $p(\molatompos_{i,t-1}|\molatompos_{i,t}, \molatompos_{i,0})$ and $p(\molatomtype_{i,t-1}|\molatomtype_{i,t}, \molatomtype_{i,0})$.
Given that $\molatompos_{i,0}$ and $\molatomtype_{i,0}$ are unknown in the backward process, 
at each step $t$, \method uses a new pocket-conditioned ligand generator, \predictor
(\PredictorSubsection section),
to estimate them as follows: 
\begin{equation}
    \label{eqn:predictor}
    \{(\tilde{\mathbf{x}}^g_{i, 0,t}, \tilde{\mathbf{v}}^g_{i, 0,t})\} = \predictor(\{\molatompos_{i,t}\}, \{\molatomtype_{i,t}\}, \{\pascalar_j\}, \{\pavec_j\}),
\end{equation}
where $\tilde{\mathbf{x}}^g_{i,0,t}$ and $\tilde{\mathbf{v}}^g_{i,0,t}$ 
are the estimates of $\molatompos_{i,0}$ and $\molatomtype_{i,0}$ at timestep $t$, respectively.
$\{\pascalar_j\}$ and $\{\pavec_j\}$ represent the pretrained scalar and vector pocket embeddings).
Using the estimates, $\molatompos_{i,t-1}$ can be sampled as:
\begin{equation}
\label{eqn:aprox_pos_posterior}
\begin{aligned}
p(\molatompos_{i,t-1}\mid\molatompos_{i,t}) & \approx q(\molatompos_{i,t-1}\mid\molatompos_{i,t}, \tilde{\mathbf{x}}^g_{i,0,t}) 
 = \mathcal{N}(\molatompos_{i,t-1}\mid\mu(\molatompos_{i,t}, \tilde{\mathbf{x}}^g_{i,0,t}),\tilde{\beta}_{t}^{\mathbf{x}}\mathbf{I}),
\end{aligned}
\end{equation}
where $\mu(\molatompos_{i,t},\tilde{\mathbf{x}}^g_{i,0,t})$ is the estimated mean of Gaussian transition distribution (detailed in Appendix~\ref{appendix:backward_diffusion}). 
Similarly, $\molatomtype_{i,t-1}$ can be sampled as 
\begin{equation}
    p(\molatomtype_{i,t-1}\mid\molatomtype_{i,t})\approx q(\molatomtype_{i,t-1}\mid\molatomtype_{i,t}, \tilde{\mathbf{v}}^g_{i,0,t}) 
=\mathcal{C}(\molatomtype_{i,t-1}\mid\mathbf{c}(\molatomtype_{i,t}, \tilde{\mathbf{v}}^g_{i,0,t})), 
\label{eqn:aprox_type_posterior}
\end{equation}
where $\mathbf{c}\big(\molatomtype_{i,t}, \tilde{\mathbf{v}}^{g}_{i,0,t}\big)$ denotes the estimated probability vector that parameterizes the categorical transition distribution (detailed in Appendix~\ref{appendix:backward_diffusion}).
In practice, the denoising update for $\molatompos_{i,t-1}$ can be calculated as
\begin{equation}
\label{eqn:pos_update}
    \molatompos_{i,t-1}=\mu(\molatompos_{i,t}, \tilde{\mathbf{x}}_{i,0,t}^g) + \sqrt{\tilde{\beta}^{\mathbf{x}}_{t}}\mathbf{z}^{\mathbf{x}}_{i,t-1},
\end{equation}
where $\mathbf{z}^{\mathbf{x}}_{i,t-1}\sim \mathcal{N}(\mathbf{0}, \mathbf{I})$ is a standard Gaussian noise 
sampled at time $t$, 
and the denoising update for $\molatomtype_{i,t-1}$ can be calculated as
\begin{equation}
\label{eqn:type_update}
\molatomtype_{i,t-1}
= \operatorname{onehot}\!\left(
    \arg\max_\ell \big[\,\log \mathbf{c}(\molatomtype_{i,t}, \tilde{\mathbf{v}}^{g}_{i,0,t})_{\ell}
                   + \mathbf{z}^{\mathtt v}_{i,t-1,\ell}\,\big]
\right),
\end{equation}
where $\mathbf{z}_{i,t-1,\ell}^{\mathbf{v}}\sim\text{Gumbel}(0,1)$ for $\ell=1,\dots,d_a$ ($d_a$ is the total number of atom types)
are independent Gumbel noises. 
Here, we use the Gumbel-max trick to sample from the discrete categorical distribution. More details of the backward process are available in {Appendix~{\ref{appendix:backward_diffusion}}}.

%

\paragraph{Determining Ligand Sizes}
%
During the generative process, the {ligand size} (i.e., the number of heavy atoms) 
is unknown and must be determined before atom positions and types can be generated.
The ligand size is sampled based on the volume of the condition pocket, referred to as the pocket size. 
The pocket size is determined by the region in proximity to the reference ligand. 
Pocket atoms that are among the three nearest neighbors of the ligand atoms are selected, 
and the maximum pairwise distance among them defines the pocket size.
Pocket sizes are grouped into ordered intervals, 
each associated with an empirical distribution of ligand atom counts from the training data.
The ligand size is then sampled from the empirical distribution corresponding to the pocket’s interval.

\subsubsection*{\method Model Training}
\label{sec:method:condition_generation:training}

\method is trained to recover original ligand atom positions $\molatompos_{i,0}$ and 
features $\molatomtype_{i,0}$, conditioned on the protein pocket.
Specifically, \method uses the combination of the following three losses as its loss function. 

\paragraph*{Atom Position Loss} 

This loss measures the Euclidean errors between the predicted positions $\tilde{\mathbf{x}}^g_{i, 0, t}$ at step $t$
and the ground-truth positions $\molatompos_{i, 0}$ as follows: 
\begin{equation}
\label{eqn:loss:atom_pos}
\begin{aligned}
\mathcal{L}_t^\mathbf{x}(\mol) & = w_t^{\mathbf{x}}\sum_{\forall \molatom_i\in\scriptsize{\mol}} \text{KL}(q(\molatompos_{i,t-1}|\molatompos_{i,t}, \molatompos_{i,0})||q(\molatompos_{i,t-1}|\molatompos_{i,t}, \tilde{\mathbf{x}}^g_{i, 0, t})) \\
& = w_t^{\mathbf{x}}\sum_{\forall \molatom_i\in\scriptsize{\mol}} \|\tilde{\mathbf{x}}^g_{i, 0, t} - \molatompos_{i, 0}\|, 
\end{aligned}
\end{equation}
where $w_t^\mathbf{x}$ is a time-dependent weight at step $t$, calculated by the signal-to-noise ratio~\cite{kingma2021variational} (detailed in Appendix~\ref{appendix:forward_diffusion}), and KL is the Kullback-Leibler divergence~\cite{kullback1951information}.
As $t$ increases, $\bar{\alpha}_t^\mathbf{x}$ decreases monotonically, resulting in an increase in noise level 
and a corresponding decrease in $w_t^\mathbf{x}$ until it reaches the threshold $\delta$.
As a result, it encourages the model to emphasize more accurate reconstruction of
 molecular structures when the data have low noise levels and contain sufficient signals.

\paragraph*{Atom Type Loss} 

This loss measures the KL divergence{~\cite{kullback1951information}
between the ground-truth posterior 
$q(\molatomtype_{i, t-1}|\molatomtype_{i,t}, \molatomtype_{i,0})$ and 
its estimate $q(\molatomtype_{i,t-1}|\molatomtype_t, \tilde{\mathbf{v}}^g_{i,0,t})$ as follows:
\begin{equation}
\label{eqn:loss:atom_type}
\begin{aligned}
\mathcal{L}_t^\mathbf{v}(\mol) & =\sum_{\forall \molatom_i\in\scriptsize{\mol}} \text{KL}(q(\molatomtype_{i,t-1}|\molatomtype_{i,t}, \molatomtype_{i,0})||q(\molatomtype_{i,t-1}|\molatomtype_{i,t}, \tilde{\mathbf{v}}^g_{i,0,t})) \\
& = \sum_{\forall \molatom_i\in\scriptsize{\mol}}\text{KL}(\mathbf{c}(\molatomtype_{i,t}, \molatomtype_{i,0})||\mathbf{c}(\molatomtype_{i,t}, \tilde{\mathbf{v}}^g_{i,0,t})), 
\end{aligned}
\end{equation}
where $\mathbf{c}(\molatomtype_{i,t}, \molatomtype_{i,0})$ is the categorical distribution of $\molatomtype_{i,t}$ and $\mathbf{c}_{\Theta}(\molatomtype_{i,t}, \tilde{\mathbf{v}}^g_{i,0,t})$ is an estimate of $\mathbf{c}(\molatomtype_{i,t},\molatomtype_{i,0})$.
%

\paragraph*{Bond Type Loss} 

Following the literature~\cite{chen2025generating}, 
this loss is to help \method better understand the relations among atoms.
By explicitly learning bonds, the model better enforces chemical validity and yields more realistic molecular structures.
To achieve this, at each time step $t$, for each $l$-th layer of ligand generator \predictor (\PredictorSubsection section),
\method is optimized to accurately predict the bond types among pairs of ligand atoms. 
Specifically, the bond type loss is defined as follows:
\begin{equation}
\label{eqn:loss_bond}
\mathcal{L}^{\mathtt{b}}_{t,l}({\mol}) = \sum_{\forall \molatom_i\in \scriptsize{\mol}}~\sum_{\forall \molatom_j \in \scriptsize{\molneighbour(\molatom_{i}| \mol)}} \mathsf{H}(\bondpred_{ij,t,l}, \bondtype_{ij}),
\end{equation}
where $\mathsf{H}(\cdot)$ denotes \mbox{cross-entropy} loss,
$\bondpred_{ij,t,l}$ 
represents the bond type predictions
(Equation~\ref{eqn:bond_type})
between atoms $\molatom_i$ and $\molatom_j$ at the $l$-th layer at time step $t$; {$\molneighbour(\molatom_{i}| \mol)$} 
denotes the $k$-nearest 
ligand atoms of atom $\molatom_i$ in position $\molatompos_{i,t}$;  
and $\bondtype_{ij}$ denotes one-hot vector indicating the ground-truth bond type
between $\molatom_i$ and $\molatom_j$. 
The total bond type prediction loss is computed by aggregating losses across different layers as follows:
\begin{equation}
\mathcal{L}^{\mathbf{b}}_{t}({\mol}) = \frac{w_t^\mathbf{x}}{L-1}\sum_{l=1}^{L-1} \mathcal{L}^{\mathbf{b}}_{t,l}({\mol}) + w_t^\mathbf{x}\mathcal{L}^{\mathbf{b}}_{t, L}({\mol}),
\label{eqn:diff:obj:bond}
\end{equation}
where $w_t^\mathbf{x}$ is the same timestep weight used in the atom position loss; 
$L$ represents the total number of layers in the \mbox{pocket-conditioned} ligand generator \predictor.
Similar to $\mathcal{L}^{\mathbf{x}}_t(\mol)$ (Equation~\ref{eqn:loss:atom_pos}), the weight $w_t^\mathbf{x}$ encourages the model to focus more on accurately predicting bond types 
when the data provides sufficient signals, rather than being confused by major noises in the data. 
Note that, similar to Jumper \etal~\cite{Jumper2021}, in Equation~\ref{eqn:diff:obj:bond}, \method 
assigns different weights to the final layer (i.e., $l$=$L$) compared to the other layers, 
as we empirically find this design benefits the generation performance. 

%
\paragraph{Overall \method loss}

The overall loss function for \method is defined as follows:
\begin{equation}
\mathcal{L}_{\text{\scriptsize diff}}
= \mathbb{E}_{\scriptsize{\mol \sim \text{\footnotesize$\mathtt{D}$}}}
\mathbb{E}_{t \sim \mathcal{U}(1, T)}
\big(\mathcal{L}^{\mathbf{x}}_t(\mol)
+ \xi\,\mathcal{L}^{\mathbf{v}}_t(\mol)
+ \zeta\,\mathcal{L}^{\mathbf{b}}_t(\mol)\big),
\end{equation}
where $\mathtt{D}$ is the training binding complexes; ${T}$ is the total number of diffusion timesteps; $\mathcal{U}(1, T)$ represents uniform distribution over these timesteps; $\xi > 0$ and $\zeta > 0$ are two \mbox{hyper-parameters} that balance $\mathcal{L}^{\mathbf{x}}_t$(\mol), $\mathcal{L}^{\mathbf{v}}_t$(\mol) and $\mathcal{L}^{\mathbf{b}}_{t}({\mol})$.

\subsection*{Pocket-conditioned Ligand Generator (\predictor)
}
\label{sec:method:condition_generation:predictor}


\method employs a pocket-conditioned ligand generator, \predictor, to denoise 
noisy ligand atom positions $\tilde{\mathbf{x}}^g_{i,t}$ and features $\tilde{\mathbf{v}}^g_{i,t}$ at each diffusion step. 
At a high level,  denoising involves predicting the added noises and recovering the original ligand structures by leveraging
chemical information and spatial information,
including pocket geometry encoded by pretrained \pocketenc, ligand geometry, and the ligand's relative spatial arrangement within the pocket.
To guide this process, \predictor learns expressive ligand atom representations by modeling two types of interactions: 
{\textbf{(1)} interactions within the ligand,
which model relationships among neighboring atoms to capture each atom’s local environment,}
and {\textbf{(2)} 
pocket-ligand
interactions, which capture pocket circumstances to provide pocket-informed ligand representations.}
Through modeling these interactions, \predictor captures both the ligand's structure and its external interaction patterns with the pocket.
Furthermore, by pretraining pocket representation, \predictor avoids modeling pocket structure and instead focuses on learning intra-ligand and 
pocket-ligand
interactions.
When no ambiguity arises, we will eliminate subscript $t$ in the notations and use $(\molatompos_i, \molatomtype_i)$ for brevity. 

\subsubsection{Ligand Representation Learning}

This section describes how \predictor learns ligand representations (embeddings with superscript $^g$) that capture both internal molecular structure and interactions with the pocket.
\predictor employs a multi-layer GNN to iteratively update ligand atom embeddings based on intra-ligand and pocket-ligand interactions.
Specifically, at the $l$-th layer of the GNN, for each ligand atom $\molatom_{i}$, \predictor 
learns an invariant scalar embedding $\atomscalar_{i,l}\in\mathbb{R}^{d \times 1}$ 
to capture atom features and an equivariant vector embedding 
$\atomvec_{i,l}\in\mathbb{R}^{d\times 3}$ for 3D structural features.
For $\molatom_{i}$, \predictor updates $\atomscalar_{i,l}$ and $\atomvec_{i,l}$ 
by aggregating information from neighboring atoms of $\molatom_{i}$ in the current noisy ligand 
and incorporating interactions with neighboring pocket atoms.
The embeddings $\atomscalar_{i, l}$ and $\atomvec_{i, l}$ are updated as follows: 
\begin{equation}
(\atomscalar_{i, l}, \atomvec_{i, l}) = \text{GVP}(\mathbf{h}_{i,l}, \mathcal{Y}_{i,l}), \text{ where }
\label{eqn:plp:gvp}
\end{equation}
\begin{equation}
\begin{aligned}
\mathbf{h}_{i,l} & = 
\biggr[\underbrace{\molatomtype_i, \atomscalar_{i,l-1}}_{\mathclap{\text{from the ligand}}}, 
~~~\underbrace{\pcscalar_{i,l-1}, \rcscalar_{i,l-1},}_{\mathclap{\text{from the pocket}}} 
~\sum_{\molatom_j\in \scriptsize{\molneighbour(\molatom_{i}| \mol)}}z_{ji,l}\messatomscalar_{ji,l}\biggr],\\
\mathcal{Y}_{i,l} & = 
\biggr[\underbrace{\molatompos_i, \atomvec_{i,l-1}}_{\mathclap{\text{from the ligand}}}, 
 ~\underbrace{\pcvec_{i,l-1}, \rcvec_{i,l-1},}_{\mathclap{\text{from the pocket}}} 
 \sum_{\molatom_j\in \scriptsize{\molneighbour(\molatom_{i}|\mol)}}z_{ji,l}\messatomvec_{ji,l}\biggr],
 \end{aligned}
 \label{eqn:plp:concat}
\end{equation}
where $^c$ represents embeddings that model pocket-ligand atomic interactions,
and {$^a$ represents residue-level interaction embeddings.}
$\pcscalar_{i,l-1}$ and $\pcvec_{i,l-1}$ denote the scalar and vector embeddings that encode 
how ligand atom $\molatom_i$ interacts with neighboring pocket atoms from the $(l-1)$-th layer of the GNN
(detailed in Equation~\ref{eqn:pc_embed}); 
$\rcscalar_{i,l-1}$ and $\rcvec_{i,l-1}$ denote the scalar and vector embeddings that capture structures from surrounding pocket residues for ligand atom $\molatom_i$
(detailed in Equation~\ref{eqn:rc_embed}); 
$\messatomscalar_{ji,l}$ and $\messatomvec_{ji,l}$ represent the scalar and vector message embeddings
in the $(l-1)$-th layer of the GNN
 to propagate information from $\molatom_i$'s neighboring atom $\molatom_j$ in the ligand \mol (i.e., $\molneighbour(\molatom_{i}| \mol)$) to $\molatom_i$; 
$z_{ji,l}$ is the attention weights modeled in a similar way as in Equation~\ref{eqn:attention}.
The convolution in Equation~\ref{eqn:plp:gvp} combines information from ligand atom neighbors and  pocket atom/residue neighbors, allowing each ligand atom to refine its representation with structural and chemical information
from both the ligand and the pocket.
$\messatomscalar_{ji,l}$ and $\messatomvec_{ji,l}$ are updated as follows:
\begin{equation}
(\messatomscalar_{ji,l}, \messatomvec_{ji,l}) = \text{GVP}({{\messatomscalarhat_{ji,l}}}, {{\messatomvechat_{ji,l}}}),\text{ where }
\label{eqn:mess:diff_gvp}
\end{equation}
\begin{equation}
\label{eqn:mess:mess_pass}
\begin{aligned}
{\messatomscalarhat_{ji,l}} &= 
[{\mathbf{m}}^g_{ji,l-1}, {\distg{j}{i}}, \bondpred_{ji,l-1}], \\
\quad {\messatomvechat_{ji,l}} &= 
[\messatomvec_{ji,l-1}, {\molatompos_j - \molatompos_i}], 
\end{aligned}
\end{equation}
where $\bondpred_{ji,l-1}$ is the embedding of the bond type between $\molatom_i$ and $\molatom_j$ (detailed in Equation~\ref{eqn:bond_type}), and {$\distg{j}{i}$ is the distance between $\molatom_i$ 
and $\molatom_j$}, and $(\molatompos_j - \molatompos_i)$ represents the displacement vector from $\molatom_i$ to $\molatom_j$.
\predictor utilizes bond type embeddings $\bondpred_{ji,l}$ to facilitate its understanding of relations among atoms.
Particularly, \predictor generates these bond type embeddings as follows:
\begin{equation}
\bondpred_{ji,l} = 
	\begin{cases}
	\text{MLP}([\atomscalar_{i,l} + \atomscalar_{j,l}, \text{abs}(\atomscalar_{i,l} - \atomscalar_{j,l}), \distg{j}{i}]), & \text{if} \; l = 0, \\
	\text{MLP}([\atomscalar_{i,l} + \atomscalar_{j,l}, \text{abs}(\atomscalar_{i,l} - \atomscalar_{j,l}), \|\atomvec_{i,l}\|_2 + \|\atomvec_{j,l}\|_2, \text{abs}(\|\atomvec_{i,l}\|_2 - \|\atomvec_{j,l}\|_2)]), & \text{ otherwise}.
	\end{cases}
	\label{eqn:bond_type}
\end{equation}
where abs($\cdot$) represents the absolute difference.
Intuitively, the bond type embeddings are constructed to be agnostic to the order of 
atoms $\molatom_i$ and $\molatom_j$ by using two invariant operations: the sum and the absolute difference operation. 
The sum combines features from neighboring atoms $\molatom_i$ and $\molatom_j$, and the absolute difference estimates the
distance between $\molatom_i$ and $\molatom_j$ on latent space,
offering a comprehensive representation of the bond.
After $L$ layers, \predictor estimates the atom position and type for each ligand atom $\molatom_i$ using the learned atom embeddings as follows:
\begin{equation}
{\tilde{\mathbf{x}}^g_{i} = \molatompos_i + 
\atomvec_{i,L}}
, \quad \tilde{\mathbf{v}}^g_{i} = \text{softmax}(\text{MLP}(\atomscalar_{i,L})), 
\label{eqn:pred:pred}
\end{equation}
where {$\tilde{\mathbf{x}}^g_{i}$ and $\molatompos_i$ are the predicted noise-free atom positions} and 
{input noisy atom positions} of atom $\molatom_i$, 
respectively;  $\atomvec_{i,L}\in\mathbb{R}^{1\times3}$ is the predicted noise for $\molatom_i$, and $\atomscalar_{i,L}$ is used for atom type prediction of $\molatom_i$.
$\tilde{\mathbf{v}}^g_i$ represents the predicted clean categorical distribution of atom features.
%

\subsubsection{Pocket-ligand Interaction Learning}  
%
This section describes how to learn pocket-ligand interaction embeddings (embeddings with superscript $^c$ or $^r$ in Equation~\ref{eqn:plp:concat}) by leveraging pocket embeddings (embeddings with superscript $^p$) encoded by \pocketenc.
For each ligand atom $\molatom_i$, at each $l$-th layer of the GNN,
\predictor learns 
two invariant embeddings $\pcscalar_{i,l}$ and $\rcscalar_{i,l}$, and two equivariant embedding $\pcvec_{i,l}$ and  $\rcvec_{i,l}$,
encoding how $\molatom_i$ interacts with neighboring pocket atoms and residues.
These embeddings are learned in a recurrent manner as follows:
\begin{equation}
\label{eqn:pc_embed}
\pcscalar_{i,l}  = \text{MLP}([\tilde{\mathbf{s}}^c_{i,l}, \pcscalar_{i,l-1}]), ~~\pcvec_{i,l}  = \text{VN-MLP}([\tilde{\mathcal{H}}^c_{i,l}, \pcvec_{i,l-1}]), 
\end{equation}
\begin{equation}
\label{eqn:rc_embed}
\rcscalar_{i,l}  = \text{MLP}([\tilde{\mathbf{s}}^a_{i,l}, \rcscalar_{i,l-1}]), ~~\rcvec_{i,l}  = \text{VN-MLP}([\tilde{\mathcal{H}}^a_{i,l}, \rcvec_{i,l-1}]). 
\end{equation}
That is, \predictor updates the atom-level $(\pcscalar_{i}, \pcvec_{i})$ and residue-level $(\rcscalar_{i}, \rcvec_{i})$ pocket-ligand interaction embeddings recurrently across layers.
We model
$\tilde{\mathbf{s}}^c_{i,l}$ and $\tilde{\mathcal{H}}^c_{i,l}$ as follows:
\begin{equation}
(\tilde{\mathbf{s}}^c_{i,l}, \tilde{\mathcal{H}}^c_{i,l})  = \text{GVP}(\mathbf{b}_{i,l}^c, \mathcal{I}_{i,l}^c), \text{ where }
\label{eqn:pil:gvp}
\end{equation}
\begin{equation}
\begin{aligned}
\mathbf{b}_{i,l}^c  & = \sum_{j\in 
\scriptsize{\interactionneighbour(\molatom_i|\pocketset)}}^n w_{ij, l}^c \text{MLP}([\dist{i}{j}, \atomscalar_{i,l}, \pascalar_j]), \\
\mathcal{I}_{i,l}^c  & = \sum_{\pocketatom_j\in \scriptsize{\interactionneighbour(\molatom_i|\pocketset)}}^n w_{ij,l}^c \text{VN-MLP}([\pocketatompos_j - \molatompos_i, \atomvec_{i,l}, \pavec_{j}]),
\end{aligned}
\end{equation}
where $\pocketset$ is the target pocket of ligand $\mol$, and ${\interactionneighbour(\molatom_i|\pocketset)}$ denotes the $n$-nearest pocket atoms 
surrounding the ligand atom $\molatom_i$;
$\pascalar_j$ and $\pavec_j$ are pretrained pocket atom embeddings (Equation~\ref{eqn:pred:gvp});
$\atomscalar_{i,l-1}$ and $\atomvec_{i,l-1}$ are ligand atom embeddings as presented in 
Equation \ref{eqn:plp:gvp}; and $\dist{i}{j}$ is the Euclidean distance between $\molatom_i$ and $\pocketatom_j$.
The attention weights are determined based on the ligand embeddings (Equation~\ref{eqn:plp:gvp}) and pocket embeddings provided by $\pocketenc$ as follows:
\begin{equation}
\begin{aligned}
w_{ij,l}^c & = \text{softmax}(\text{MLP}([\dist{i}{j}, \mathbf{v}^c_{ij,l}, \atomscalar_{i,l}, \pascalar_j])), \\
\mathbf{v}_{ij,l}^c &= \|\text{VN-MLP}([\pocketatompos_j - \molatompos_i, \atomvec_{i,l}, \pavec_{j}])\|_2,
\end{aligned}
\label{eqn:pil:attn}
\end{equation}
where $\mathbf{v}_{ij,l}$ encodes the spatial interactions between $\molatom_i$ and $\pocketatom_j$.
While $(\tilde{\mathbf{s}}^c_{i,l}, \tilde{\mathcal{H}}^c_{i,l})$ accounts for atomic pocket-ligand interactions, we also model interactions between ligand atoms and pocket residues. $(\tilde{\mathbf{s}}^a_{i,l}, \tilde{\mathcal{H}}^a_{i,l})$ are obtained in a similar manner by aggregating interactions between $\molatom_i$ and $m$-nearest pocket residue $\residue_j$. We summarize the ligand generation procedure of \method incorporating \pocketenc and \predictor in Algorithm~\ref{alg:p2diff}.

\begin{algorithm}[!h]
    \caption{\method for ligand generation}
    \label{alg:p2diff}
    	\textbf{Required Input: Ligand $\mol$, pocket $\pocketset$} 
    \begin{algorithmic}[1]
        \State $n = \text{{sampleNumAtoms}}(\mol, \pocketset)$ \Comment{sample the number of ligand atoms from pocket size}
        \State $\{\molatompos_T\}^n \sim \mathcal{N}(0, \mathbf{I})$
        \Comment{initialize positions of n ligand atoms}
        \State $\{\molatomtype_T\}^n \sim \mathcal{C}(K, \frac{1}{K})$ \Comment{initialize types of n ligand atoms}
        \State $\pascalar, \pavec = \pocketenc(\pocketatompos, \pocketatomtype, \residuepos, \residuetype)$ \Comment{encode pocket into embeddings using \pocketenc}
        \For{$t = T$ to $1$}
            \State $(\xtilg_{0,t}, \vtilg_{0,t}) = \predictor(\molatompos_t, \molatomtype_t, \pascalar, \pavec)$ 
            \Comment{predict noise-free ligand using \predictor}
            \State $\molatompos_{t-1} = q(\molatompos_{t-1}|\molatompos_t, \xtilg_{0,t})$ \Comment{sample $\molatompos_{t-1}$ using Gaussian posterior  (Equation \ref{eqn:pos_update})}
            \State $\molatomtype_{t-1} = q(\molatomtype_{t-1}|\molatomtype_t, \vtilg_{0,t})$ \Comment{sample $\molatomtype_{t-1}$ using categorical posterior (Equation \ref{eqn:type_update})}
        \EndFor
        \State $\mol_{gen} = (\molatompos_0, \molatomtype_0)$ 
        \State \Return $\mol_{gen}$ \Comment{return the generated ligand}
    \end{algorithmic}
\end{algorithm}

\subsection*{\method with Generation-time Property-aware Optimization (\methodopt)}

\method is well tailored for pocket-specific ligand generation. In this section, we develop a new method that further improves the ligands generated by \method to exhibit additional desired properties (e.g., ADMET).  
To this end, we extend \method by introducing an iterative, generation-time property-aware optimization procedure that refines the generated ligands with respect to these desired properties. 
In essence, this optimization process ``aligns'' the model with desired properties beyond binding affinity or drug-likeness. We formulate this ``alignment'' process as a generation-time optimization problem, in which the objective is to steer the generated ligands toward exhibiting
more favorable properties, such as higher BBBP or lower carcinogenicity.
The proposed optimization scheme, named \opt,  operates entirely during the model's inference process.
That is, all model parameters of \method are kept fixed at this stage. 
Without requiring any additional model training, the approach is both computationally efficient and readily adaptable to different optimization objectives while preserving \method's original generative performance on general molecular metrics. We refer to the resulting framework that combines \method with \opt as \methodopt.
%

\subsubsection{Alignment Problem Formulation}
\label{sec:method:optimization:formulation}
%
In \method, the stochasticity of ligand generation arises from noise trajectories of atom positions and atom types ($\mathbf{z}^{\mathbf{x}}_{t}$ in Equation \ref{eqn:pos_update} and $\mathbf{z}^{\mathbf{v}}_{t}$ in Equation \ref{eqn:type_update}). Noise trajectories govern the denoising dynamics throughout the diffusion process, determining how ligand structures evolve over time and ultimately shaping the final generated ligand.

Therefore, with the diffusion model parameters fixed, the entire generation process can be viewed as a deterministic mapping that takes the noise trajectories as input and outputs the generated ligand. 
These noise trajectories constitute the optimization variables in \methodopt. 
After generation, the ligand is evaluated by an evaluator function $R(\cdot)$, which defines the objective function for the optimization problem. 
We then employ gradient-based optimization to iteratively update the noise trajectories, using the gradients of the objective function to steer the generation process towards ligands with more favorable properties.

\paragraph{Noise as Optimization Variables} 
%
The generation process of \method, described in Algorithm~\ref{alg:p2diff}, consists of $T$ denoising steps. It begins with atom positions initialized as pure Gaussian noise,
$\molatompos_{i,T}\sim \mathcal{N}(\mathbf{0}, \mathbf{I})$ (here, we define $\noisepos_{i,T} = \molatompos_{i,T} $), 
and atom types sampled uniformly at random,
$\molatomtype_{i,T} \sim C(K, \frac{1}{K}).$ 
The uniform categorical sampling is implemented via the Gumbel-max trick. Specifically, for each atom $i$ and type $k$, in \methodopt, we draw $\noisetype_{i,T,k}\sim\mathrm{Gumbel}(0,1)$ for $k=1, \dots, K$, and set $\molatomtype_{i,T}=\arg\max_k\{\noisetype_{i,T,k}\}$. 
At each subsequent step $t$, the atom positions $\molatompos_{i,t}$ and types $\molatomtype_{i,t}$ are updated 
according to Equations~\ref{eqn:pos_update} and \ref{eqn:type_update}, where Gaussian noise $\noisepos_{i,t-1}\sim\mathcal{N}(\mathbf{0}, \mathbf{I})$ and Gumbel noises $\noisetype_{i,t-1,k}\sim\text{Gumbel(0,1)}$ for $k=1,\dots,K$ are injected into the denoising updates to produce the predicted $\molatompos_{i,t-1}$ and $\molatomtype_{i,t-1}$. 
For compactness, we write $[\noisetype_{i,t,1}, \dots, \noisetype_{i,t,k}]$ as $\noisetype_{i,t}$. 
We collect these noises into two noise trajectories: Gaussian noise trajectory $\{\noisepos_{i,T}, \noisepos_{i,T-1}, \dots, \noisepos_{i,0} \}$ and Gumbel noise trajectory $\{\noisetype_{i,T}, \noisetype_{i,T}, \dots, \noisetype_{i,0}\}$. 
By treating the noise trajectories as optimization variables, we aim to iteratively refine them to guide the generative process towards chemical space with improved target properties.
\paragraph{Objective Function} 
Let $D_0$ represent the generated ligand at the final timestep. To evaluate its quality with respect to a desired ADMET property, we employ a property-specific evaluator function. 
We denote the evaluator as $R(\cdot)$, which serves as the objective function in our formulation. The choice of evaluator is flexible and can be tailored to specific target properties. For example, if the goal is to improve the blood-brain barrier permeability of the generated ligand $D_0$, then $R(D_0)$ can be defined as the predicted probability (given by an external property classifier) that ligand $D_0$ is blood-brain barrier permeable. 

\paragraph{Noise Optimization} For notational simplicity, we denote the entire Gaussian noise trajectory $\{\noisepos_{i,T}, \noisepos_{i,T-1}, \dots, \noisepos_{i,0} \}$ as $\tau^{\mathbf{x}}$ and the entire Gumbel noise trajectory $\{\noisetype_{i,T}, \noisetype_{i,T-1}, \dots, \noisetype_{i,0}\}$ as $\tau^{\mathbf{v}}$. 
With \predictor fixed, $\tau^{\mathbf{x}}$ and $\tau^{\mathbf{v}}$ fully determine the generated ligand $\mol_0$. Therefore, we write $\mol_0=M(\tau^{\mathbf{x}}, \tau^{\mathbf{v}})$, where $M$ denotes the deterministic mapping from noise trajectories to the generated ligand.
Our goal is to adjust the noise trajectories $\tau^{\mathbf{x}}$ and $\tau^{\mathbf{v}}$ so that the generated ligand has a more favorable property score $R$. The generation-time noise optimization problem is thus formulated as:
\begin{equation}
\label{eqn:noise_opt}
\max_{\tau^{\mathbf{x}}, \tau^{\mathbf{v}}} R(M(\tau^{\mathbf{x}}, \tau^{\mathbf{v}}))
\end{equation}

\subsubsection{\methodopt Optimization}

We adopt a straightforward gradient-based method, as presented in Algorithm~\ref{alg:noiseopt}, to solve 
the noise optimization problem {(Equation}~\ref{eqn:noise_opt})}. 
We use the noise trajectories produced by \method as a starting point to be optimized.
The initial noise trajectories are iteratively refined through gradient-based updates.
Specifically, at each iteration, we evaluate the objective $R(M(\tau^{\mathbf{x}}, \tau^{\mathbf{v}}))$, calculate the gradient of $R$ with respect to $\tau^{\mathbf{x}}, \tau^{\mathbf{v}}$, and update the noise accordingly via gradient ascent. 
This approach is applicable when the gradient of the objective function $R$ is available.
However, in practice, such gradients are often inaccessible, since many ADMET evaluators operate as black-box predictors that provide only scalar property scores.
To overcome this limitation, we employ zero\emph{th}-order gradient estimation, enabling gradient-based updates without explicit analytic gradients. Algorithm~\ref{alg:p2diffopt} presents the \methodopt algorithm. 

\begin{algorithm}[!h]
    \caption{\noiseopt for noise trajectory optimization}
    \label{alg:noiseopt}
    	\textbf{Required Input}: Initial noise trajectories $\tau^{\mathbf{x}}_0, \tau^{\mathbf{v}}_0$, iteration number $N$, step size $\alpha$
    \begin{algorithmic}[1]
    \State $r_{\mathrm{best}}\gets R(M(\tau^{\mathbf{x}}_0, \tau^{\mathbf{v}}_0)), n_{\mathrm{best}} \gets 0$ \Comment{evaluate objective at initialization}
        \For{$n = 0$ to $N-1$}
            \State $r \gets R(M(\tau^{\mathbf{x}}_n, \tau^{\mathbf{v}}_n))$ \Comment{evaluate current objective}
            \If{$r>r_{\mathrm{best}}$}
                \State $r_{\mathrm{best}} \gets r, n_{\mathrm{best}} \gets n$ \Comment{update best step}
            \EndIf
            \State $\hat{\nabla}R(M(\tau^{\mathbf{x}}_n, \tau^{\mathbf{v}}_n))=\zograd(\tau^{\mathbf{x}}_n, \tau^{\mathbf{v}}_n)$ \Comment{estimate zeroth-order gradients via Algorithm~\ref{alg:zograd}}
            \State $\tau^{\mathbf{x}}_{n+1}=\tau^{\mathbf{x}}_n+\alpha\hat{\nabla}_{\tau^{\mathbf{x}}}R(M(\tau^{\mathbf{x}}_n, \tau^{\mathbf{v}}_n))$ \Comment{update Gaussian noise trajectory}
            \State $\tau^{\mathbf{v}}_{n+1}=\tau^{\mathbf{v}}_n+\alpha\hat{\nabla}_{\tau^{\mathbf{v}}}R(M(\tau^{\mathbf{x}}_n, \tau^{\mathbf{v}}_n))$ \Comment{update Gumbel noise trajectory}
        \EndFor
        \State $\mol_{\mathrm{best}}\gets M(\tau^{\mathbf{x}}_{n_{\mathrm{best}}}, \tau^{\mathbf{v}}_{n_{\mathrm{best}}})$
        \State \Return $\mol_{\mathrm{best}}$ \Comment{return the best ligand found among all iterations}
    \end{algorithmic}
\end{algorithm}

\paragraph{Zero\emph{th}-order Gradient Estimation} Algorithm~\ref{alg:zograd} outlines the zero\emph{th}-order gradient estimation procedure used in \methodopt. 
The key idea is to estimate the gradient of the objective function by observing how small random perturbations to the input variables change the output score. 
At each iteration, we sample random perturbation directions for both the Gaussian and Gumbel noise trajectories, $u^{\mathbf{x}}\sim\mathcal{N}(0, \mathbf{I}^{\mathbf{x}})$ and $u^{\mathbf{v}}\sim\mathcal{N}(0, \mathbf{I}^{\mathbf{v}})$, and generate perturbed versions of the trajectories: $(\tau^{\mathbf{x}}+\mu u^{\mathbf{x}}, \tau^{\mathbf{v}}+\mu u^{\mathbf{v}})$ and $(\tau^{\mathbf{x}}-\mu u^{\mathbf{x}}, \tau^{\mathbf{v}}-\mu u^{\mathbf{v}})$, where $\mu$ is the perturbation magnitude.
The objective function is then evaluated at these perturbed points, and the gradient is estimated from the scaled difference between the corresponding objective values, providing a stochastic estimation of the true gradient direction.
By averaging the estimated directional derivatives over multiple perturbations, we obtain unbiased gradient estimates with respect to both $\tau^{\mathbf{x}}$ and $\tau^{\mathbf{v}}$. This enables effective gradient-based optimization even when the evaluator is a black-box predictor that provides only function values without analytic gradients.
\begin{algorithm}[!h]
\caption{\zograd for zeroth-order gradient estimation}
\label{alg:zograd}
\textbf{Required Input}: Current trajectories $(\tau^{\mathbf{x}}, \tau^{\mathbf{v}})$, smoothing $\mu>0$, number of random perturbations $H$
\begin{algorithmic}[1]
  \State Initialize accumulators: $g^{\mathbf{x}}\!\gets \mathbf{0}$, $g^{\mathbf{v}}\!\gets \mathbf{0}$
  \For{$h=1$ to $H$}
    \State $u^{\mathbf{x}}_h\sim\mathcal{N}(0, \mathbf{I^{\mathbf{x}}})$,
    $u^{\mathbf{v}}_h\sim\mathcal{N}(0, \mathbf{I^{\mathbf{v}}})$ \Comment{sample random perturbations}
    \State $r^{+}_h \gets R\!\left(M\!\big(\tau^{\mathbf{x}}+\mu u^{\mathbf{x}}_h,\; \tau^{\mathbf{v}}+\mu u^{\mathbf{v}}_h\big)\right)$ \Comment{positive perturbations}

    \State $r^{-}_h \gets R\!\left(M\!\big(\tau^{\mathbf{x}}-\mu u^{\mathbf{x}}_h,\; \tau^{\mathbf{v}}-\mu u^{\mathbf{v}}_h\big)\right)$ \Comment{negative perturbations}

    \State $\Delta r_h \gets \dfrac{r^{+}_h - r^{-}_h}{2\mu}$ \Comment{calculate directional difference}
    \State
      $g^{\mathbf{x}} \gets g^{\mathbf{x}} + \Delta r_h \, u^{\mathbf{x}}_h, 
      \qquad
      g^{\mathbf{v}} \gets g^{\mathbf{v}} + \Delta r_h \, u^{\mathbf{v}}_h$ \Comment{accumulate estimators}
  \EndFor
  \State $\widehat{\nabla}_{\tau^{\mathbf{x}}} R(M(\tau^{\mathbf{x}},\tau^{\mathbf{v}})) \gets \dfrac{1}{H} g^{\mathbf{x}}$,
  \quad
  $\widehat{\nabla}_{\tau^{\mathbf{v}}} R(M(\tau^{\mathbf{x}},\tau^{\mathbf{v}})) \gets \dfrac{1}{H} g^{\mathbf{v}}$
  \State \textbf{Return} $\widehat{\nabla}_{\tau^{\mathbf{x}}} R(M(\tau^{\mathbf{x}},\tau^{\mathbf{v}}))$, $\widehat{\nabla}_{\tau^{\mathbf{v}}} R(M(\tau^{\mathbf{x}},\tau^{\mathbf{v}}))$
  
\end{algorithmic}
\end{algorithm}

\begin{algorithm}[h]
    \caption{\methodopt for ligand generation and optimization}
    \label{alg:p2diffopt}
    	\textbf{Required Input}: $\mol$, $\pocketset$, iteration number $N$ 
    \begin{algorithmic}[1]
        \State $n = \text{{sampleNumAtoms}}(\mol, \pocketset)$ \Comment{sample the number of ligand atoms from pocket size}
        \State $\{\noisepos_{i,T}\}_{i=1}^n \sim \mathcal{N}(0, \mathbf{I}), \quad \{\molatompos_T\}_{i=1}^n=\{\noisepos_{T}\}_{i=1}^n$ \Comment{initialize positions of $n$ ligand atoms}
        \State $\{\noisetype_{i,T}\}_{i=1}^{n} \sim \mathrm{Gumbel}(0,1)^{K},\quad
      \{\molatomtype_{T}\}_{i=1}^n = \left\{\mathrm{onehot}\!\left(\arg\max_{k\in[K]} \noisetype_{i,T,k}\right)\right\}_{i=1}^n$
      \Comment{initialize types of $n$ ligand atoms}
        \State $\tau^{\mathbf{x}} \gets \emptyset,\; \tau^{\mathbf{v}} \gets \emptyset$ 
        \Comment{initialize noise trajectories}
        \State $\pascalar, \pavec = \pocketenc(\pocketatompos, \pocketatomtype, \residuepos, \residuetype)$ \Comment{encode pocket into embeddings using \pocketenc}
        \For{$t = T$ to $1$}
            \State $(\xtilg_{0,t}, \vtilg_{0,t}) = \predictor(\molatompos_t, \molatomtype_t, \pascalar, \pavec)$ \Comment{predict noise-free ligand using the \predictor}
            \State $\noisepos_{t-1}\sim\mathcal{N}(0,\mathbf{I}),\quad \molatompos_{t-1}= q(\molatompos_{t-1}|\molatompos_t, \xtilg_{0,t})$ 
            \Comment{sample $\molatompos_{t-1}$ using Gaussian posterior (Equation~\ref{eqn:pos_update})}
            \State $\noisetype_{t-1}\sim\text{Gumbel}(0,1), \quad \molatomtype_{t-1}= q(\molatomtype_{t-1}|\molatomtype_t, \vtilg_{0,t})$ 
            \Comment{sample $\molatomtype_{t-1}$ using categorical posterior (Equation~\ref{eqn:type_update})}
            \State $\tau^{\mathbf{x}} \gets \tau^{\mathbf{x}} \cup \{\noisepos_{t-1}\}, \;\; \tau^{\mathbf{v}} \gets \tau^{\mathbf{v}} \cup \{\noisetype_{t-1}\}$
            \Comment{append noise to trajectories}
        \EndFor
        \State $\mol_{\text{best}} = \mathsf{NoiseOpt}(\tau^{\mathbf{x}}, \tau^{\mathbf{x}}, N)$
        \Comment{Algorithm \ref{alg:noiseopt}, noise optimization}
        \State \Return $\mol_{\text{best}}$
    \end{algorithmic}
\end{algorithm}

The overall \methodopt framework is summarized in Algorithm~\ref{alg:p2diffopt}. At a high level, \methodopt records the noise trajectories produced by \method during sampling and then optimizes these trajectories to refine the resulting ligand. Specifically, it first samples the ligand atom count based on the pocket size and initializes atom positions with Gaussian noise and atom types with Gumbel noise. Conditioned on pocket embeddings produced by \pocketenc, \methodopt runs the generation process from $t=T$ to $1$: at each step, \predictor estimates the noise-free ligand $(\xtilg_{0,t}, \vtilg_{0,t})$, after which atom positions and types are updated by sampling from the corresponding Gaussian and categorical posteriors. The injected position and type noises throughout sampling are recorded to the trajectories $(\tau^{\mathbf{x}}, \tau^{\mathbf{v}})$. Finally, the noise optimization module refines the sampled ligand by optimizing the trajectories for $N$ iterations and returns the optimized molecule $\mol$.

\section*{Data Availability}
\label{section:data_availability}

The \dataset and \CD datasets are available via GitHub at \url{https://github.com/ninglab/conDitar-dev}, and the CrossDocked2020 dataset is available
via GitHub at \url{https://github.com/gnina/models/tree/master/data/CrossDocked2020}. Additional data, including our generated molecules
and trained models, are publicly available via GitHub at \url{https://github.com/ninglab/conDitar-dev}.

\section*{Code Availability}
\label{section:code_availability}

The code is publicly available via GitHub at \url{https://github.com/ninglab/conDitar-dev}.

\section*{Acknowledgements}
\label{section:acknowledgements}

 This project was made possible, in part, by support from the National Science Foundation grant nos. 2435819 (X.N.) and 2450988 (M.H.),  
 the National Library of Medicine grant no. 1R01LM014385 (X.N.), 
the National Center for Advancing Translational Sciences grant no. UM1TR004548 (X.N.), and Sanofi iDEA-TECH Awards North America (X.N.).
 Any opinions, findings, and conclusions or recommendations expressed in this manuscript are those of the authors, and do not necessarily reflect the views of the funding agencies.
We thank Benjamin Burns, Reza Averly, Maggie Samaan, and Trieu Nguyen for their help with paper writing.
We are grateful for their careful review, constructive feedback, and assistance in improving the clarity 
and organization of the manuscript.
We thank Avery Meyer for her contributions to the design and implementation of the graphical user interface, which helps improve the usability and accessibility of the method.

\section*{Author Contributions}
\label{section:author_contribution}

X.N. conceived the research and conducted the project administration. 
M.H. and X.N. investigated the research, 
obtained funding and resources for the research, and supervised the student authors (X.N. supervised 
R.G, Z.C., and F.B.; M.H. supervised J.P.).
R.G., Z.C. and X.N. designed the computational methodologies of \method. 
J.P. and M.H. designed the computational methodologies of \opt.
R.G. conducted data curation, formal analysis, computational methodology (\method) implementation, result analysis, and visualization. 
J.P. conducted formal analysis, computational methodology (\opt) implementation, result analysis, and visualization. 
Z.C. contributed to the methodology design and implementation (\method). 
F.B. contributed to formal analysis, result analysis, and visualization. 
D.K. designed the computational chemistry evaluation and 
experimental validation of the research and contributed to the computational chemistry results analysis and visualization. 
J.K. and A.S. conducted the molecular synthesis experiments and the results analysis.
H-P.B., Y.L. and M.L. contributed to the computational chemistry evaluation and experimental validation analysis. 
L.I. conducted biological assays for \PDLone.
R.G., J.P., M.H., and X.N. drafted the original manuscript. 
R.G., J.P., F.B., D.K., M.H., and X.N. conducted the manuscript editing and revision. 
All authors reviewed the final paper.


\clearpage

\bibliographystyle{naturemag}
\bibliography{paper}

\newpage
\setcounter{page}{1}

\appendix

\renewcommand{\thetable}{A\arabic{table}}
\setcounter{table}{0}
\renewcommand{\thefigure}{A\arabic{figure}}
\setcounter{figure}{0}

\section{Equivariance and Invariance}
\label{appendix:concept}

\subsection{{Equivariance}}
\label{appendix:concept:equivariance}

By definition, a function $f(\mathsf{x})$ is equivariant if all translation and rotation transformations from the special Euclidean group SE(3)~\cite{Atz2021} applied to the input $\mathsf{x}\in\mathbb{R}^3$ are mirrored accordingly in the output as follows:
\begin{equation}
f(\mathbf{R}\mathsf{x}+\mathbf{t}) = \mathbf{R}f(\mathsf{x}) + \mathbf{t},
\end{equation}
where, $\mathbf{t}\in\mathbb{R}^3$ is a translation transformation
and $\mathbf{R}\in\mathbb{R}^{3\times3}$ ($\mathbf{R}^{\mathsf{T}}\mathbf{R}=\mathbf{I}$) is a rotation transformation.
\method leverages GVP and VN-MLP to ensure \PRL and \diff are equivariant, effectively capturing the geometric features of objects regardless of any translation or rotation transformations. 

\subsection{{Invariance}}
\label{appendix:concept:invariance}

A function $f(\mathsf{x})$ is invariant if its output remains constant under all translation and rotation transformations of the input $\mathsf{x}$:
\begin{equation}
f(\mathbf{R}\mathsf{x}+\mathbf{t}) = f(\mathsf{x}),
\end{equation}
where $\mathbf{t}$ and $\mathbf{R}$ represents any translation and rotation transformation, respectively.
\method learns invariant scalar embeddings for pocket atoms and ligand atoms, capturing inherent features (e.g., atom features) that remain invariant to any translation or rotation transformation.

\section{Forward Diffusion}
\label{appendix:forward_diffusion}

{In the forward process, \diff adds noises step by step to the atom position ($\molatompos_{i,t}$) and atom feature ($\molatomtype_{i,t}$) in the training ligands.}
For brevity, in this section, we eliminate the subscript $i$ in the notations when no ambiguity arises.
{
The probability of atom positions $\molatompos_t$ sampled given $\molatompos_{t-1}$, denoted as $q(\molatompos_t|\molatompos_{t-1})$, is defined as follows:
\begin{equation}
q(\molatompos_t|\molatompos_{t-1}) = \mathcal{N}(\molatompos_t|\sqrt{1-\beta^{\mathbf{x}}_t}\molatompos_{t-1}, \beta^{\mathbf{x}}_t\mathbf{I}), 
\label{eqn:noiseposinter}
\end{equation}
where $\mathcal{N}(\cdot)$ is a Gaussian distribution of $\molatompos_t$ with mean $\sqrt{1-\beta_t^{\mathbf{x}}}\molatompos_{t-1}$ and covariance $\beta_t^{\mathbf{x}}\mathbf{I}$.
The probability of atom features at time step $t$, $\molatomtype_t$, given that at time step $t-1$,  $\molatomtype_{t-1}$ is defined as follows:
\begin{equation}
q(\molatomtype_t|\molatomtype_{t-1}) = \mathcal{C}(\molatomtype_t|(1-\beta^{\mathbf{v}}_t) \molatomtype_{t-1}+\beta^{\mathbf{v}}_t\mathbf{1}/d_a),
\label{eqn:noisetypeinter}
\end{equation}
where $\mathcal{C}$ is a categorical distribution.

Given the above definitions, the probability of atom positions ($\molatompos_t$) and atom features ($\molatomtype_t$) at any time step $t$ can be derived from those at the initial time step ($\molatompos_0$ and $\molatomtype_0$) as follows: 
\begin{eqnarray}
& q(\molatompos_t|\molatompos_{0}) & = \mathcal{N}(\molatompos_t|\sqrt{\cumalpha^{\mathbf{x}}_t}\molatompos_0, (1-\cumalpha^{\mathbf{x}}_t)\mathbf{I}), \label{eqn:noisepos}\\
& q(\molatomtype_t|\molatomtype_0)  & = \mathcal{C}(\molatomtype_t|\cumalpha^{\mathbf{v}}_t\molatomtype_0 + (1-\cumalpha^{\mathbf{v}}_t)\mathbf{1}/K), \label{eqn:noisetype}\\
& \text{where }\cumalpha^{\mathtt{u}}_t & = \displaystyle\prod_{\tau=1}^{t}\alpha^{\mathtt{u}}_\tau, \ \alpha^{\mathtt{u}}_\tau=1 - \beta^{\mathtt{u}}_\tau, \ {\mathtt{u}}={\mathbf{x}} \text{ or } {\mathbf{v}},\;\;\;\label{eqn:noiseschedule}
\label{eqn:pos_prior}
\end{eqnarray}
where $\bar{\alpha}^{\mathtt{u}}_t$ is a weight decreasing monotonically from 1 to 0 over $t=[1,T]$.
Specifically, $\cumalpha^{\mathtt{u}}_t$ ($\mathtt{u}={\mathbf{x}}\text{ or }{\mathbf{v}}$) approaches 1 as $t \rightarrow 1$, allowing $\molatompos_t$ or $\molatomtype_t$ to approximate $\molatompos_0$ or $\molatomtype_0$.
Conversely, 
$\cumalpha^{\mathtt{u}}_t$ ($\mathtt{u}={\mathbf{x}}\text{ or }{\mathbf{v}}$) approaches 0 as $t \rightarrow T$, 
{which makes} $q(x_T^g \mid x_0^g)$ resemble $\mathcal{N}(0, I)$ and $q(v_T^g \mid v_0^g)$ resemble $\mathcal{C}(1/d_a)$.
}

As shown by Ho~\etal~\cite{ho2020ddpm}, the ground-truth Normal posterior of atom positions, $p(\molatompos_{t-1}|\molatompos_t, \molatompos_0)$, could be calculated in a closed form as below:
\begin{eqnarray}
& q(\molatompos_{t-1}|\molatompos_t, \molatompos_0) = \mathcal{N}(\molatompos_{t-1}|\mu(\molatompos_t, \molatompos_0), \tilde{\beta}^\mathbf{x}_t\mathbf{I}), \label{eqn:gt_pos_posterior_1}\\
&\mu(\molatompos_t, \molatompos_0)=\frac{\sqrt{\bar{\alpha}^{\mathbf{x}}_{t-1}}\beta^{\mathbf{x}}_t}{1-\bar{\alpha}^{\mathbf{x}}_t}\molatompos_0\!+\!\frac{\sqrt{\alpha^{\mathbf{x}}_t}(1-\bar{\alpha}^{\mathbf{x}}_{t-1})}{1-\bar{\alpha}^{\mathbf{x}}_t}\molatompos_t, \\
& \tilde{\beta}^\mathbf{x}_t=\frac{1-\bar{\alpha}^{\mathbf{x}}_{t-1}}{1-\bar{\alpha}^{\mathbf{x}}_{t}}\beta^{\mathbf{x}}_t.\;\;\;
\end{eqnarray}
Similarly, as shown in Hoogeboom~\etal~\cite{hoogeboom22diff}, the ground-truth categorical posterior of atom features $p(\molatomtype_{t-1}|\molatomtype_{t}, \molatomtype_0)$ can be calculated as below:
\begin{eqnarray}
& q(\molatomtype_{t-1}|\molatomtype_{t}, \molatomtype_0) = \mathcal{C}(\molatomtype_{t-1}|\mathbf{c}(\molatomtype_t, \molatomtype_0)), \label{eqn:gt_atomfeat_posterior_1}\\
& \mathbf{c}(\molatomtype_t, \molatomtype_0) = \tilde{\mathbf{c}}/{\sum_{k=1}^K \tilde{c}_k}, \label{eqn:gt_atomfeat_posterior_2} \\
& \tilde{\mathbf{c}} = [\alpha^{\mathbf{v}}_t\molatomtype_t + \frac{1 - \alpha^{\mathbf{v}}_t}{d_a}]\odot[\bar{\alpha}^{\mathbf{v}}_{t-1}\molatomtype_{0}+\frac{1-\bar{\alpha}^{\mathbf{v}}_{t-1}}{d_a}], 
\label{eqn:gt_atomfeat_posterior_3}
\end{eqnarray}
where $\mathbf{c}(\molatomtype_t, \molatomtype_0)$ denotes the probability over the $d_a$ classes,  $\tilde{c}_k$ denotes the likelihood of the $k$-th class, and $\odot$ is the element-wise product operation.

\section{Backward Generative Process}
\label{appendix:backward_diffusion}

In the backward process, \diff generates realistic binding ligands from random noise. 
Particularly, conditioned on $\molatompos_{i,t}$ and $\tilde{\mathsf{x}}^g_{i,0,t}$, the probability $p(\molatompos_{i,t-1}|\molatompos_{i,t})$ could be estimated using the approximated posterior $p_{\boldsymbol{\Theta}}(\molatompos_{i,t-1}|\molatompos_{i,t}, \tilde{\mathsf{x}}^g_{i,0,t})$, as shown in Ho \etal~\cite{ho2020ddpm}.
Same as Appendix~\ref{appendix:forward_diffusion}, we eliminate the subscript $i$ in the notations when no ambiguity arises.
The approximation of $p(\molatompos_{t-1}|\molatompos_t)$ is calculated as follows:
{
\begin{equation}
\begin{aligned}
p(\molatompos_{t-1}|\molatompos_t) & \approx q(\molatompos_{t-1}|\molatompos_t, \tilde{\mathbf{x}}^g_{0,t}) \\
& = \mathcal{N}(\molatompos_{t-1}|\mu(\molatompos_t, \tilde{\mathbf{x}}^g_{0,t}),\tilde{\beta}_t^{\mathbf{x}}\mathbf{I}),
\end{aligned}
\label{eqn:aprox_pos_posterior}
\end{equation}
where $\mu(\molatompos_t, \tilde{\mathsf{x}}^g_{0,t})$ is an estimate
of $\mu(\molatompos_t, \molatompos_{0})$ by replacing $\molatompos_0$ with its approximation $\tilde{\mathsf{x}}^g_{0,t}$ 
in Equation~{\ref{eqn:gt_pos_posterior_1}}.
Similarly, as shown in Hoogeboom~\cite{hoogeboom22diff}, given $\molatomtype_t$ and $\tilde{\mathsf{v}}^g_{0,t}$, the probability of $\molatomtype_{t-1}$ conditioned on $\molatomtype_{t}$, $p(\molatomtype_{t-1}|\molatomtype_t)$, 
can be estimated 
by the approximated posterior $q(\molatomtype_{t-1}|\molatomtype_t, \tilde{\mathsf{v}}^g_{0,t})$ as below:
\begin{equation}
\begin{aligned}
p(\molatomtype_{t-1}|\molatomtype_t)\approx q(\molatomtype_{t-1}|\molatomtype_{t}, \tilde{\mathsf{v}}^g_{0,t}) 
=\mathcal{C}(\molatomtype_{t-1}|\mathbf{c}(\molatomtype_t, \tilde{\mathsf{v}}^g_{0,t})),\!\!\!\!
\end{aligned}
\label{eqn:aprox_atomfeat_posterior}
\end{equation}
where $\mathbf{c}(\molatomtype_t, \tilde{\mathsf{v}}^g_{0,t})$ is an estimate of $\mathbf{c}(\molatomtype_t, \molatomtype_0)$
by replacing $\molatomtype_0$  
with its estimate $\tilde{\mathsf{v}}^g_{0,t}$ in Equation~\ref{eqn:gt_atomfeat_posterior_1}.
}

\section{Parameters for Reproducibility}
\label{appendix:param}

{In \method, we trained two models \PRL and \diff for pocket representation learning and pocket-conditioned ligand generation, respectively.}
We implemented both models using Python 3.9.18 and PyTorch 2.1.0, together with the corresponding PyTorch Geometric dependencies, including torch-scatter 2.1.2, torch-cluster 1.6.3, and torch-geometric 2.6.1. The environment is built with CUDA 12.1 (pytorch-cuda 12.1).
We trained both models on an NVIDIA A100 GPU with 40GB memory and a CPU with 80GB memory.

\subsection{Parameters for \PRL}
\label{appendix:param:prl}

{In \PRL, we set the dimension of all the hidden layers, including GVP layers (Equation~\ref{eqn:pred:gvp} and \ref{eqn:message}) and MLP layers (Equation~\ref{eqn:attention} to \ref{eqn:prldec:type}) as 128, and the dimension of both residue scalar and vector embeddings ($\prscalar$ and $\prvec$) and pocket atom scalar and vector embeddings ($\pascalar$ and $\pavec$) as 128.
To represent the pocket structure, we constructed two graphs using the $k$-nearest neighbors based on Euclidean distance: one for pocket atoms with $k_a=16$ and one for residues with $k_r=8$. 
We set the layer number of graph neural networks for both atom and residue as 3.
We optimized the \PRL model with Adam with its parameters (0.950, 0.999), learning rate 0.001, and batch size 64.
We trained \PRL for maximum 100 epochs and the training took $\sim$36 hours in total.
}

\subsection{Parameters for \predictor}
\label{appendix:param:diff}

{In \predictor, we set the dimension of all the scalar hidden layers, including GVP layers (Equation~\ref{eqn:plp:gvp}, \ref{eqn:mess:diff_gvp} and \ref{eqn:pil:gvp}) and VN-MLP and MLP layers (Equation~\ref{eqn:bond_type} to \ref{eqn:pil:attn}) as 128.
We set the dimensions of all the vector hidden layers in GVPs as 32. 
We set the number of layers $L$ in \predictor as 10. 
We built the distance-based atomic graphs for ligands using the $m$-nearest neighbors based on Euclidean distance with $m=8$. 
For each ligand atom, we consider its nearest $n_a=32$ pocket atoms for protein-ligand interaction learning.
In addition, we consider residue-level interaction by connecting each ligand atom to its $n_r=4$ nearest pocket residues.
}

{In the forward process of {the} diffusion model, following Guan \etal~\cite{guan2023targetdiff}, we used a sigmoid $\beta$ schedule for the variance schedule $\beta_t^{\mathbf{x}}$ of atom positions to add noises into atom positions as below:
\begin{equation}
\beta_t^{\mathbf{x}} = \text{sigmoid}(w_1(2 t / T - 1)) (w_2 - w_3) + w_3
\end{equation}
in which $w_i$($i$=1,2, or 3) with $w_1=6$, $w_2=1.e-7$ and $w_3=0.01$ are hyperparameters; $T=1,000$ is the maximum step.
For atom types, we used a cosine $\beta$ schedule~\cite{nichol2021} for $\beta_t^{\mathbf{v}}$ as below:
\begin{equation}
\begin{aligned}
& \bar{\alpha}_t^{\mathbf{v}} = \frac{f(t)}{f(0)}, f(t) = \cos(\frac{t/T+s}{1+s} \cdot \frac{\pi}{2})^2\\
& \beta_t^{\mathbf{v}} = 1 - \alpha_t^{\mathbf{v}} = 1 - \frac{\bar{\alpha}_t^{\mathbf{v}} }{\bar{\alpha}_{t-1}^{\mathbf{v}} }
\end{aligned}
\end{equation}
in which $s$ is a hyperparameter and set as 0.01.
Same as \PRL, we optimized \diff using Adam with its parameters (0.950, 0.999), the learning rate 0.001, and batch size 16.
The training takes $\sim$100 hours in total.
}

\subsection{Parameters for \methodopt}
In \methodopt, we set the number of optimization iterations to $N=10$ and the step size to $\alpha=0.1$ for noise optimization. For zero\emph{th}-order gradient estimation, we use $H=4$ random perturbations with a smoothing parameter of $\mu=0.03$.

\section{Details of \dataset dataset}
\label{appendix:dataset_info}

\newcolumntype{C}[1]{>{\centering\arraybackslash}p{#1}}
\newcolumntype{U}[1]{>{\raggedright\collectcell\MakeUppercase}p{#1}<{\endcollectcell}}

\begin{table}[H]
	\centering
		\caption{Summary of \dataset}
	\label{tbl:target_property}
\begin{threeparttable}
	\begin{scriptsize}
	\begin{tabular}{
	U{0.05\linewidth}
	U{0.05\linewidth}
	U{0.05\linewidth}
	p{0.4\linewidth}
	@{\hspace{2pt}}c@{\hspace{2pt}}
	@{\hspace{2pt}}c@{\hspace{2pt}}
	@{\hspace{2pt}}c@{\hspace{2pt}}
	@{\hspace{2pt}}c@{\hspace{2pt}}
	@{\hspace{2pt}}c@{\hspace{2pt}}
	@{\hspace{2pt}}c@{\hspace{2pt}}
	}
		\toprule
		\textbf{P\MakeLowercase{rotein} PDB ID} & \textbf{G\MakeLowercase{ene} N\MakeLowercase{ame}} & \textbf{L\MakeLowercase{igand} ID} & \multirow{3}{*}{\textbf{Disease Category}} & \multirow{3}{*}{\textbf{HIA}} & \multirow{3}{*}{\textbf{BBBP}} & \multirow{3}{*}{\textbf{hERG}} & \multirow{3}{*}{\textbf{Ames}} & \multirow{3}{*}{\textbf{Carci}} & \multirow{3}{*}{\textbf{Source}} \\
        \midrule
        14gs & GSTP1 & cbd   & Breast, lung and prostate cancer & & & & \checkmark & \checkmark & Crossdocked \\
		1d7j & FKB1A  & fk5   & Neurodegenerative/Alzheimer’s Disease & & \checkmark & \checkmark & & & Crossdocked \\ 
        1dxo & NQO1   & e09  & Cancer; Neurodegenerative diseases & & \checkmark & & & \checkmark & Crossdocked \\ 
        1jv3 & UAP1   & ud2   & Metabolic sugar-processing disorders; Cancer& \checkmark & & & & \checkmark & Ortholog\\ 
        1r1h & NEP    & bir   & Alzheimer’s Disease; Acute lymphoblastic leukemia & & \checkmark & \checkmark & & & Additional\\ 
        2d82 & CBP    & ttr   & Genetic developmental disorders; Cancer, Neurodegenerative diseases& & \checkmark & & & \checkmark & Additional \\ 
        2ohf & OLA1   & acp  & Cancer; Neurodegenerative diseases & & \checkmark & & & \checkmark & Ortholog\\ 
        2pqw & LMBL1  & mlz   & Cancer & & & & \checkmark & \checkmark & Crossdocked\\ 
        2rfn & MET    & am7  & Cancer; Neurodevelopmental disorders & & \checkmark & & & \checkmark & Additional\\ 
        2rma & PPIA   & ea4   & Cancer; Autoimmune diseases & & & & \checkmark & \checkmark & Crossdocked\\ 
        2z7q & KS6A1  & acp   & Breast, prostate, and colon cancer; Head and neck squamous cell carcinoma; Glioma & & & & \checkmark & \checkmark & Ortholog\\ 
        2zen & QPCT   & 1bn   & Neurodegenerative diseases  & & \checkmark & & & \checkmark & Crossdocked\\ 
        3b6h & PTGIS  & mxd   & Cardiovascular diseases & \checkmark & & \checkmark & & & Crossdocked\\ 
        3c3u & AK1C1  & c2u  & Developmental disorders; Metabolic diseases; Cancer progression & & & & & \checkmark & Ortholog\\ 
        3dzh & CD38   & cvr   & Cancer (including chronic lymphocytic leukemia (CLL) and multiple myeloma) & & & & \checkmark & \checkmark & Crossdocked\\ 
        3ej8 & NOS2   & itu   & Cancer development; Inflammatory diseases & & & \checkmark & & \checkmark & Crossdocked\\ 
        3gvu & ABL2   & sti  & Brain cancer & & \checkmark & & & \checkmark &  Ortholog\\ 
        3hy9 & ATS5   & 099   & Cartilage degradation; Connective tissue inflammation & & & \checkmark & & \checkmark & Crossdocked\\ 
        3jyh & DPP2   & opy  & Immune-related diseases; Inflammatory conditions& & & \checkmark & & \checkmark & Crossdocked\\ 
        3kc1 & F16P1  & 2t6  & Metabolic Disorders; Cancer & \checkmark & & & & \checkmark & Crossdocked\\ 
        3l3n & ACE    & nxa   & Cardiovascular diseases; Diabetic nephropathy & \checkmark & & \checkmark & & & Crossdocked\\ 
        3nos & NOS3   & h4b  & Cardiovascular diseases & \checkmark & & \checkmark & & & Ortholog\\ 
        3o96 & AKT1   &  iqo  & Cancer (especially breast cancer, lung cancer, glioblastoma) & & & & \checkmark & \checkmark & Crossdocked\\ 
        4aua & CDK6   & 4au   & Acute lymphoblastic leukemia (ALL); Breast cancer; Glioblastoma & & & & \checkmark & \checkmark & Additional\\ 
        4azf & DYRK2  & 7aa  & Cancer (especially gliomas and breast cancer); Neurodegenerative/Alzheimer’s disease; Neurodevelopmental disorders& & & & \checkmark &  \checkmark & Crossdocked\\ 
        4bel & BACE2  & dbo   & Alzheimer's disease & & \checkmark & \checkmark & & & Crossdocked\\ 
        4g3d & M3K14  & 13v  & Immune system dysfunction & & & \checkmark & & \checkmark & Crossdocked\\ 
        4gvd & TIAM1  & ans   & Neurodevelopmental disorders; Cancer & & \checkmark & & & \checkmark & Crossdocked\\ 
        4iwq & TBK1   & su6   & Neurodegenerative diseases & & \checkmark & & & \checkmark & Crossdocked\\ 
        4ja8 & IDHP   & 1k9   & Cancers (especially blood cancers such as acute myeloid leukemia (AML) and gliomas); Neurological disorders  & & & & \checkmark & \checkmark & Crossdocked\\ 
        4nbl & CASP6  & 2j6   & Neurodegenerative diseases & & \checkmark & & & \checkmark & Additional\\ 
        4ntj & Q9H244 & AZJ & Bleeding disorders and inflammation & & & \checkmark & & \checkmark & Additional\\ 
        4oiv & NR1H4  & xx9   & Liver Diseases & \checkmark & & & & \checkmark & Additional\\ 
        4pxz & Q9H244 & 6AD & Bleeding disorders and inflammation & & & \checkmark & & \checkmark & Crossdocked\\ 
        4rn0 & HDAC8  & l8g  & Developmental disorders; Cancers & & \checkmark & & & \checkmark & Crossdocked\\ 
        4rt7 & FLT3   & p30   & Acute myeloid leukemia (AML) & & & & \checkmark & \checkmark & Additional\\ 
        4tos & TNKS1  & 355  & Cancer; Genetic bone disorders & & & & & \checkmark & Crossdocked\\ 
        4u0i & KIT    & 0li   & Tyrosine kinase inhibitors (TKIs); Acute myeloid leukemia (AML) & \checkmark & & \checkmark & & & Additional\\ 
        4yhj & GRK4   & an2  & Hypertension & \checkmark & & \checkmark & & & Crossdocked\\ 
        5aeh & TNKS2  & 8ir   & Cancer; Neurodegenerative diseases & & \checkmark & & & \checkmark & Crossdocked\\ 
        5d7n & SIR3   & 1ns  & Neurodegenerative diseases & & \checkmark & & & \checkmark & Crossdocked\\ 
        5i0b & PAK4   & m77   & Cancer & & & & \checkmark & \checkmark & Crossdocked\\ 
        5liu & AK1BA  & qap   & Metabolic disorders & \checkmark & & & & \checkmark & Crossdocked\\ 
        5mgl & BAZ2A  & 7mu   & Cancer; Inflammatory disorders; Developmental defects & & & \checkmark & & \checkmark & Crossdocked\\ 
        5ngz & UBE2T  & 2bg  & Cancer; Genetic disorders & & & & \checkmark & \checkmark & Crossdocked\\ 
        5q0k & NR1H4  & 9ld   & Liver diseases & \checkmark & & & & \checkmark & Crossdocked\\ 
        5tjn & BGAT   & nlc   & Blood group–related transfusion reactions; Liver diseases & & & \checkmark & \checkmark & \checkmark & Crossdocked\\ 
        5w2g & IPMK   & adp  & Hereditary cancers; Immune-related diseases & & & \checkmark & & \checkmark & Crossdocked\\ 
        6cm4 & DRD2   & 8nu   & Psychiatric disorders; Addictive behaviors; Movement disorders  & & \checkmark & \checkmark & & & Additional\\ 
        6i4u & DLDH   & fad   & Genetic metabolic disorders & & & & & \checkmark & Additional\\ 
        6nco & APOE   & kqp  & Alzheimer’s disease; Cardiovascular disease & & \checkmark & \checkmark & & & Additional\\ 
        6ng1 & NOS1   & kly & Neurodegenerative diseases & & \checkmark & & & \checkmark & Ortholog\\ 
        6orv & GNAS2  & n2v   & Genetic/Developmental Disorders & \checkmark & & \checkmark & & \checkmark & Additional\\ 
        6suk & NEP    & ft8  & Alzheimer’s disease; Acute lymphoblastic leukemia & & \checkmark & \checkmark & & & Additional\\ 
        7aqf & P05121  & RV2 & Abnormal bleeding & & & \checkmark & & \checkmark & Additional\\ 
        7c2e & GNAS2  & ffr  & Bone development; Hormone regulation & & & & & \checkmark & Additional\\ 
        7cm0 & QPCT   & nt6  & Alzheimer’s disease & & \checkmark & \checkmark & & & Additional\\ 
        7m3z & HAVR2  & yqd & Cancer; Autoimmune diseases; Viral infections & & & \checkmark & & \checkmark & Additional\\ 
        7sh0 & ERAP2  & giy  & Autoimmune diseases & & & \checkmark & & \checkmark & Additional\\ 
        7t1d & SIR2   & e7k  & Neurodegenerative diseases & & \checkmark & & & \checkmark & Additional\\ 
        7umo & U119A  & nt6  &  Immunodeficiency disorderss & & & \checkmark & & \checkmark & Additional\\ 
        7wc8 & 5HT2A  & 92s  & Mental health disorders & & \checkmark & \checkmark & & & Additional\\ 
        8ayh & CO5    & h1h  & Immune deficiency; Autoimmune disorders; Inflammatory diseases& & & \checkmark & & \checkmark & Additional\\ 
        8d0e & SARM1  & q0c  & Neurodegenerative diseases & & \checkmark & & & \checkmark & Additional\\ 
        8enk & DX39B  & adp & Autoimmune diseases; Cancer; Neurodegenerative diseases & & & \checkmark & & \checkmark & Ortholog\\ 
        8hfl & SC6A2  & 1xr  & Psychiatric disorders & & \checkmark & \checkmark & & & Additional\\ 
        8jfl & KPB1   & adp  & Glycogen storage diseases (GSD)& & & & & \checkmark & Ortholog\\ 
        8jzx & S15A4  & q09 & Autoimmune diseases; Immune system regulation & & & \checkmark & & \checkmark & Additional\\ 
        8qu8 & VHL    & wyl & Von Hippel–Lindau disease & & & \checkmark & & \checkmark & Additional\\ 
        8s4d & APJ    & a1d5n & Cardiovascular; Metabolic diseases & \checkmark & & \checkmark & & & Additional \\ 
        9fkr & KAT6A  & a1idr & KAT6A syndrome & & \checkmark & & & \checkmark & Additional\\ 
        9fmm & ACE2   & a1idx  & COVID-19; Hypertension; Cardiovascular diseases; Acute respiratory distress syndrome (ARDS) & \checkmark & & \checkmark & & & Additional\\ 
		\bottomrule
	\end{tabular}
	\end{scriptsize}
    \begin{tablenotes}[flushleft]
	\footnotesize
        \item[]
    \!\!{{Columns represent: }``Protein PDB ID''{:}
    the PDB identifier of the protein;
    ``Gene Name''{:} 
    the name of the target gene;
    ``Ligand ID''{:} 
    the identifier of the 
    {cognate ligand bound} to the target;
    {``}Disease Category{'':} the type of disease associated with the target;
    {``HIA'', ``BBBP'', ``hERG'', ``Ames'', and ``Carci'': ADMET properties selected for optimization on the target, a} 
    $\checkmark$ indicates that the corresponding property is particularly important 
    and
    therefore selected for optimization; 
    {``Source'': the origin of the targets,} {``Crossdocked''} indicates that the target protein can also be found in the \CD test set. {``Ortholog''} indicates that the target protein is a human ortholog of a non-human protein in the \CD test set. ``Additional'' indicates that the target protein is not contained in the \CD test set, and was instead sourced from clinically-relevant human proteins in PDB.} \par
	\end{tablenotes}
\end{threeparttable}
\end{table}

\section{Overall Comparison on \CD}
\label{supp:cd}

\begin{table}[H]
	\centering
		\caption{Overall Comparison on \CD}
	\label{tbl:overall_results_crossdock}
\begin{threeparttable}
	\begin{tabular}{
		@{\hspace{2pt}}l@{\hspace{2pt}}
		@{\hspace{2pt}}r@{\hspace{2pt}}
		@{\hspace{2pt}}r@{\hspace{2pt}}
		@{\hspace{4pt}}r@{\hspace{4pt}}
		@{\hspace{2pt}}r@{\hspace{2pt}}
		@{\hspace{2pt}}r@{\hspace{2pt}}
		@{\hspace{5pt}}r@{\hspace{5pt}}
		@{\hspace{2pt}}r@{\hspace{2pt}}
		@{\hspace{2pt}}r@{\hspace{2pt}}
		@{\hspace{5pt}}r@{\hspace{5pt}}
		@{\hspace{2pt}}r@{\hspace{2pt}}
	         @{\hspace{2pt}}r@{\hspace{2pt}}
		@{\hspace{5pt}}r@{\hspace{5pt}}
		@{\hspace{2pt}}r@{\hspace{2pt}}
		@{\hspace{2pt}}r@{\hspace{2pt}}
		@{\hspace{5pt}}r@{\hspace{5pt}}
		@{\hspace{2pt}}r@{\hspace{2pt}}
		@{\hspace{2pt}}r@{\hspace{2pt}}
		@{\hspace{5pt}}r@{\hspace{5pt}}
		@{\hspace{2pt}}r@{\hspace{2pt}}
		@{\hspace{2pt}}r@{\hspace{2pt}}
		@{\hspace{5pt}}r@{\hspace{5pt}}
		@{\hspace{2pt}}r@{\hspace{2pt}}
		@{\hspace{2pt}}r@{\hspace{2pt}}
		@{\hspace{2pt}}r@{\hspace{2pt}}
		%
		}
		\toprule
		\multirow{2}{*}{Method} & \multicolumn{2}{c}{Vina S$\downarrow$} & & \multicolumn{2}{c}{Vina M$\downarrow$} & & \multicolumn{2}{c}{Vina D$\downarrow$} & & \multicolumn{2}{c}{{{HA}\%$\uparrow$}}  & & \multicolumn{2}{c}{QED$\uparrow$} & & \multicolumn{2}{c}{SA$\uparrow$} & & \multicolumn{2}{c}{Div$\uparrow$} & & \multirow{2}{*}{SR\%$\uparrow$} & 
        \\
	    \cmidrule{2-3}\cmidrule{5-6} \cmidrule{8-9} \cmidrule{11-12} \cmidrule{14-15} \cmidrule{17-18} \cmidrule{20-21} 
		& Avg. & Med. &  & Avg. & Med. &  & Avg. & Med. & & Avg. & Med.  & & Avg. & Med.  & & Avg. & Med.  & & Avg. & Med.  &  \\
		\midrule
		Reference                          & -6.36 & -6.46 & &  -6.71 & -6.49 & &  -7.45 & -7.26 & &  - & - & &  0.48 & 0.47 & &  0.73 & 0.74 & &  - & - & & 25.0 &  \\
		\midrule
		\AR & {-5.75} & -5.64 & &  -6.18 & -5.88 & &  -6.75 & -6.62 & &  37.7 & 31.0 & &  0.51 & 0.50 & &  0.64 & 0.63 & &  0.70 & 0.70  & &  7.0 &  \\
		\pockettwomol   &  -5.14 & -4.70 & &  -6.42 & -5.82 & &  -7.15 & -6.79 & &  48.0 & 51.0 & & \textbf{0.57} & \textbf{0.58} & & \textbf{0.76} & \textbf{0.76} & &  0.69 & 0.71  & &  \underline{24.9} &  \\
		\diffsbdd  & 5.76 & -4.07 & &  -0.80 & -5.68 & &  -3.50 & -7.62 & &  54.2 & 58.0 & &  0.47 & 0.47 & &  0.55 & 0.56 & &  \textbf{0.76} & \textbf{0.75}  & &  6.2 &  \\
		\targetdiff     & -5.47 & \underline{-6.30} & &  {-6.64} & {-6.83} & &  {-7.80} & {-7.91} & &  {58.1} & {59.1} & &  0.48 & 0.48 & &  0.58 & 0.58 & & \underline{0.72} & \underline{0.71}  & &  10.9 &  \\
        \decompdiff & {-5.67} & -6.04 & & \underline{-7.04} &  \underline{-6.91} & &  \underline{-8.39} & \underline{-8.43}  && \underline{64.4} & \underline{71.0} && 0.45 & 0.43 && 0.61 & 0.60 && 0.68 & 0.68 && {24.5} \\
        \decompopt & \underline{-5.87} & \textbf{-6.81} && \textbf{-7.35} & \textbf{-7.72} && \textbf{-8.98} & \textbf{-9.01} && \textbf{73.5} & \textbf{93.3} && 0.48 & 0.45 && \underline{0.65} & \underline{0.65} && 0.60 & 0.61 && \textbf{52.5} \\
        \grey{\ipdiff} & \grey{-6.42} & \grey{-7.01} & & \grey{-7.45} & \grey{-7.48} & & \grey{-8.60} & \grey{-8.57} & & \grey{69.5} & \grey{75.5} & & \grey{0.51} & \grey{0.51} & & \grey{0.56} & \grey{0.56} & & \grey{0.72} & \grey{0.73} & & \grey{9.6} & \\
        \grey{\alidiff} & \grey{-7.07} & \grey{-7.95} & & \grey{-8.09} & \grey{-8.17} & & \grey{-8.76} & \grey{-8.65} & & \grey{73.4} & \grey{81.4} & & \grey{0.50} & \grey{0.51} & & \grey{0.57} & \grey{0.56} & & \grey{0.74} & \grey{0.73} & & \grey{10.7} & \\
		\midrule
		\method       &  \textbf{-5.96} & {-6.22} & &  {-6.76} & {-6.81} & &  -7.74 & -7.80 & &  {60.4} & {63.9} & &  \underline{0.53} & \underline{0.54} & &  \underline{0.65} & {0.64} & &  {0.61} & {0.61}  & &  21.2 & \\
		\bottomrule
	\end{tabular}%
	\begin{tablenotes}[normal,flushleft]
		\begin{footnotesize}
	\item 
\!\!Columns represent: ``Vina S'': the binding affinities between the initially generated poses of ligands and the protein pockets; 
		``Vina M'': the binding affinities between the poses after local structure minimization and the protein pockets;
		``Vina D'': the binding affinities between the poses determined by AutoDock Vina and the protein pockets;
		``HA'': the percentage of generated ligands with Vina D lower than those of reference ligands;
		``QED'': the quantitative estimate of drug-likeness;
		``SA'': the synthetic accessibility score;
		``Div'': the diversity among generated ligands;
            ``SR'': the percentage of generated molecules with binding affinities, QED, and SA above a certain threshold. 
            The best values are in \textbf{bold}, and the second-best values are
            \underline{underlined}. Rows in \grey{gray} are excluded from the comparison. 
$\uparrow$ / $\downarrow$ indicate that higher / lower values are better.
		\par
		\end{footnotesize}
	\end{tablenotes}
\end{threeparttable}    
\end{table}

Table~\ref{tbl:overall_results_crossdock} presents a comprehensive comparison between \method and 
all the baselines in terms of generating drug-like and diverse ligands that effectively bind to protein pockets 
on the benchmark \CD dataset. The baseline results in this table are those reported in their respective papers. \ipdiff and \alidiff are reported for completeness but excluded from best/second-best marking. Following the analysis performed on the generated ligands in \dataset (Details are in the  \DistributionalAnalysisSection section), we also analyzed atom and bond distributions on \CD. The composition results indicate that both \ipdiff and \alidiff tend to generate simplified structures with excessive carbons and single bonds, consistent with our observations on \dataset.

\method is best on Vina S (Avg) and competitive on the median, showing strong initial pose quality. 
However, after energy minimization and re-docking, its performance falls behind \decompdiff and \decompopt. 
We hypothesize that the relatively lower performance is due to the stochastic search of the optimal pose in the process of energy minimization and re-docking.
In our evaluation, we conduct a single run without pose selection, 
whereas {\decompdiff and \decompopt}
may perform multiple runs and select the best pose.
The selection strategy will improve Vina M and Vina D.
Despite lower Vina M and Vina D than \decompdiff and \decompopt, \method still outperforms most other baselines in terms of original pose quality, which could be preserved after docking through pose selection.

{\method ranks third on HA\%, just underperforming \decompdiff and \decompopt. The higher HA\% achieved by \decompdiff and \decompopt originates from their stronger Vina D performance. Improving Vina D via selection is expected to further increase HA\% for \method.}

\method is the second-best in QED and SA 
(Avg/Med), while \pockettwomol attains the highest scores. 
However, the high scores of \pockettwomol are attributed to its smaller generated molecules.
Please refer to discussions on Table~\ref{tbl:overall_results_curated} for details.
SR\% of \method is on par with the reference data, 
suggesting that its generated ligands exhibit similar quality as the reference ligands in terms of binding and chemical properties. 
We observed that \decompopt achieves significantly higher SR\% than all baselines and reference data, 
which is consistent with its low Vina D alongside good QED and SA.
However, the initial poses generated by \decompopt are of lower binding quality than those from \method.
The low Vina D can be achieved by running multiple stochastic docking attempts, which can also be applied to \method.

\section{Additional \methodopt Results}
\label{supp:opt_results}

Tables~\ref{tbl:overall_results_ames}--\ref{tbl:overall_results_hia}, together with Table~\ref{tbl:overall_results_carci} in the main text, present the comparisons of 
generated molecules with respect to different ADMET properties of interest. 
Overall, the results show that \methodopt generates molecules with substantially better ADMET properties compared to the baselines.
Specifically, Tables~\ref{tbl:overall_results_ames} and \ref{tbl:overall_results_herg} show that \methodopt produces ligands with the lowest predicted Ames mutagenicity and hERG risk, achieving approximately 60\% and 30\% better average scores, respectively, than the best baseline.
Similarly, Table~\ref{tbl:overall_results_bbbp} demonstrates that \methodopt improves the BBBP score by 17\% relative to the best baseline (\pockettwomol) across targets associated with central nervous system disorders.
Table~\ref{tbl:overall_results_hia}
shows that \methodopt can enhance absorption properties like HIA,
achieving an impressive mean score of 0.99 and outperforming all baselines.

\begin{table}[H]
	\centering
		\caption{Comparison on \dataset with optimized property Ames}
	\label{tbl:overall_results_ames}
\begin{threeparttable}
\resizebox{\textwidth}{!}{
\begin{minipage}{\textwidth}
	\begin{tabular}{
		@{\hspace{2pt}}l@{\hspace{2pt}}
		@{\hspace{2pt}}r@{\hspace{2pt}}
		@{\hspace{2pt}}r@{\hspace{2pt}}
		@{\hspace{4pt}}r@{\hspace{4pt}}
		@{\hspace{2pt}}r@{\hspace{2pt}}
		@{\hspace{2pt}}r@{\hspace{2pt}}
		@{\hspace{5pt}}r@{\hspace{5pt}}
		@{\hspace{2pt}}r@{\hspace{2pt}}
		@{\hspace{2pt}}r@{\hspace{2pt}}
		@{\hspace{5pt}}r@{\hspace{5pt}}
		@{\hspace{2pt}}r@{\hspace{2pt}}
	         @{\hspace{2pt}}r@{\hspace{2pt}}
		@{\hspace{5pt}}r@{\hspace{5pt}}
		@{\hspace{2pt}}r@{\hspace{2pt}}
		@{\hspace{2pt}}r@{\hspace{2pt}}
		@{\hspace{5pt}}r@{\hspace{5pt}}
		@{\hspace{2pt}}r@{\hspace{2pt}}
		@{\hspace{2pt}}r@{\hspace{2pt}}
		@{\hspace{5pt}}r@{\hspace{5pt}}
		@{\hspace{2pt}}r@{\hspace{2pt}}
		@{\hspace{2pt}}r@{\hspace{2pt}}
		@{\hspace{5pt}}r@{\hspace{5pt}}
		@{\hspace{2pt}}r@{\hspace{2pt}}
		@{\hspace{2pt}}r@{\hspace{2pt}}
		@{\hspace{2pt}}r@{\hspace{2pt}}
		@{\hspace{2pt}}r@{\hspace{2pt}}
		@{\hspace{2pt}}r@{\hspace{2pt}}
		}
		\toprule
		\multirow{2}{*}{Method} & \multicolumn{2}{c}{Vina S$\downarrow$} & & \multicolumn{2}{c}{Vina M$\downarrow$} & & \multicolumn{2}{c}{Vina D$\downarrow$} & & \multicolumn{2}{c}{{{HA}\%$\uparrow$}}  & & \multicolumn{2}{c}{QED$\uparrow$} & & \multicolumn{2}{c}{SA$\uparrow$} & & \multicolumn{2}{c}{Div$\uparrow$} & & \multirow{2}{*}{SR\%$\uparrow$} & 
		& \multicolumn{2}{c}{{Ames}$\downarrow$} 
        \\
	    \cmidrule{2-3}\cmidrule{5-6} \cmidrule{8-9} \cmidrule{11-12} \cmidrule{14-15} \cmidrule{17-18} \cmidrule{20-21}  \cmidrule{25-26}  
		& Avg. & Med. &  & Avg. & Med. &  & Avg. & Med. & & Avg. & Med.  & & Avg. & Med.  & & Avg. & Med.  & & Avg. & Med.  &  & & &  Avg. & Med. &\\
		\midrule
		Reference & -7.52 & -5.90 && -7.70 & -6.43 && -8.16 & -7.62 && - & - && 0.44 & 0.43 && 0.76 & 0.82 && - & - && 30.8 && 0.33 & 0.28\\
		\midrule
		\AR & -6.71 & -6.43 && -6.95 & -6.62 && -7.22 & -6.92 && 38.4 & 32.7 && 0.55 & 0.55 && \underline{0.67} & 0.66 && 0.68 & 0.70 && 14.2 && 0.48 & 0.47 \\
		\pockettwomol & -5.68 & -5.36 && -6.70 & -6.30 && -7.44 & -7.03 && 40.4 & 37.0 && \textbf{0.62} & \underline{0.62} && \textbf{0.76} & \textbf{0.77} && \textbf{0.82} & \textbf{0.84} && \textbf{30.2} && 0.42 & 0.40 \\
		\diffsbdd & -4.32 & -5.36 && -5.81 & -5.99 && -7.14 & -7.10 && 37.2 & 28.3 && 0.51 & 0.53 && 0.65 & 0.63 && \underline{0.78} & \underline{0.74} && 13.0 && 0.46 & 0.43 \\
		\targetdiff     & -6.42 & {-6.45} & &  {-7.20} & {-7.08} & & {-8.08} & {-7.97} & & 46.9 & 46.0 & &  0.50 & 0.52 & &  0.60 & 0.59 & & 0.73 & 0.68 & &  11.8  && 0.36 & 0.29 \\
        \decompdiff & -6.18 & -5.72 && -6.95 & -6.39 && -7.71 & -7.21  && 53.6 & 50.8 && 0.52 & 0.56 && 0.65 & 0.64 && 0.70 & 0.68 && 11.3 && \underline{0.34} & \underline{0.27} \\
        \decompopt & -6.26 & -5.82 && -6.90 & -6.38 && -7.71 & -7.30 && 52.9 & 51.0 && 0.51 & 0.54 && 0.66 & 0.64 && 0.71 & 0.68 && 11.7 && 0.35 & 0.29 \\
        \grey{\ipdiff}  & \grey{-8.32} & \grey{-8.25} & & \grey{-8.73} & \grey{-8.48} & & \grey{-9.39} & \grey{-8.97} & & \grey{70.7} & \grey{86.7} & & \grey{0.53} & \grey{0.55} & & \grey{0.58} & \grey{0.57} & & \grey{0.73} & \grey{0.71} & & \grey{17.1} && \grey{0.24} & \grey{0.16} \\
		\midrule
        \method    & \textbf{-7.04} & \textbf{-6.94} & & \textbf{-7.69} & \textbf{-7.36} & & \textbf{-8.48} & \textbf{-8.04} & & \textbf{63.3} & \textbf{70.7} & & {0.57} & {0.61} & & {0.65} & \underline{0.67} & & {0.68} & {0.72} & & \underline{22.9} && {0.37} & {0.38} \\
        \methodopt & \underline{-6.92} & \underline{-6.86} & & \underline{-7.56} & \underline{-7.32} & & \underline{-8.37} & \underline{-8.01} & & \underline{60.1} & \underline{66.7} & & \underline{0.59} & \textbf{0.64} & & {0.66} & \underline{0.67} & & {0.70} & {0.73} & & {22.7} && \textbf{0.10} & \textbf{0.09} \\
		\bottomrule
	\end{tabular}%
    \vspace{2pt}
	\begin{tablenotes}[flushleft]
	\footnotesize
	\item 
    {\!\!Columns represent: ``Vina S'': the predicted binding affinities between the initially generated poses of ligands and the protein pockets; 
		``Vina M'': the predicted binding affinities between the poses after local structure minimization and the protein pockets;
		``Vina D'': the predicted binding affinities between the poses determined by AutoDock Vina and the protein pockets;
		``HA'': the percentage of generated ligands with Vina D lower than those of reference ligands;
		``QED'': the quantitative estimate of drug-likeness; 
		``SA'': the synthetic accessibility score;
		``Div'': the diversity among generated ligands;
            ``SR'': the percentage of generated molecules with binding affinities, QED, and SA above a certain threshold. 
            ``Ames'': the predicted Ames mutagenicity score for generated molecules;
            The best values are in \textbf{bold}, and the second-best values are
            \underline{underlined}. Rows in \grey{gray} are excluded from the comparison. 
            $\uparrow$ / $\downarrow$ indicate that higher / lower values are better.}
	\end{tablenotes}
	\end{minipage}}
\end{threeparttable}
  \vspace{-10pt}    
\end{table}

\begin{table}[H]
	\centering
		\caption{Comparison on \dataset with optimized property hERG}
	\label{tbl:overall_results_herg}
\begin{threeparttable}
\resizebox{\textwidth}{!}{
\begin{minipage}{\textwidth}

	\begin{tabular}{
		@{\hspace{2pt}}l@{\hspace{2pt}}
		@{\hspace{2pt}}r@{\hspace{2pt}}
		@{\hspace{2pt}}r@{\hspace{2pt}}
		@{\hspace{4pt}}r@{\hspace{4pt}}
		@{\hspace{2pt}}r@{\hspace{2pt}}
		@{\hspace{2pt}}r@{\hspace{2pt}}
		@{\hspace{5pt}}r@{\hspace{5pt}}
		@{\hspace{2pt}}r@{\hspace{2pt}}
		@{\hspace{2pt}}r@{\hspace{2pt}}
		@{\hspace{5pt}}r@{\hspace{5pt}}
		@{\hspace{2pt}}r@{\hspace{2pt}}
	         @{\hspace{2pt}}r@{\hspace{2pt}}
		@{\hspace{5pt}}r@{\hspace{5pt}}
		@{\hspace{2pt}}r@{\hspace{2pt}}
		@{\hspace{2pt}}r@{\hspace{2pt}}
		@{\hspace{5pt}}r@{\hspace{5pt}}
		@{\hspace{2pt}}r@{\hspace{2pt}}
		@{\hspace{2pt}}r@{\hspace{2pt}}
		@{\hspace{5pt}}r@{\hspace{5pt}}
		@{\hspace{2pt}}r@{\hspace{2pt}}
		@{\hspace{2pt}}r@{\hspace{2pt}}
		@{\hspace{5pt}}r@{\hspace{5pt}}
		@{\hspace{2pt}}r@{\hspace{2pt}}
		@{\hspace{2pt}}r@{\hspace{2pt}}
		@{\hspace{2pt}}r@{\hspace{2pt}}
		@{\hspace{2pt}}r@{\hspace{2pt}}
		@{\hspace{2pt}}r@{\hspace{2pt}}
		}
		\toprule
		\multirow{2}{*}{Method} & \multicolumn{2}{c}{Vina S$\downarrow$} & & \multicolumn{2}{c}{Vina M$\downarrow$} & & \multicolumn{2}{c}{Vina D$\downarrow$} & & \multicolumn{2}{c}{{{HA}\%$\uparrow$}}  & & \multicolumn{2}{c}{QED$\uparrow$} & & \multicolumn{2}{c}{SA$\uparrow$} & & \multicolumn{2}{c}{Div$\uparrow$} & & \multirow{2}{*}{SR\%$\uparrow$} & 
		& \multicolumn{2}{c}{{hERG}$\downarrow$} 
        \\
	    \cmidrule{2-3}\cmidrule{5-6} \cmidrule{8-9} \cmidrule{11-12} \cmidrule{14-15} \cmidrule{17-18} \cmidrule{20-21} \cmidrule{25-26}  
		& Avg. & Med. &  & Avg. & Med. &  & Avg. & Med. & & Avg. & Med.  & & Avg. & Med.  & & Avg. & Med.  & & Avg. & Med. &  & & &  Avg. & Med. &  \\
		\midrule
        Reference 
        & -7.18 & -7.05 
        && -7.63 & -7.62 
        && -7.69 & -7.60 
        && - & - 
        && 0.45 & 0.47 
        && 0.72 & 0.75 
        && - & - 
        && 29.4 
        && 0.44 & 0.38 \\
        \midrule
        \AR 
        & -6.14 & -6.20 
        && -6.58 & -6.45 
        && -7.04 & -6.98 
        && 37.2 & 21.3 
        && \underline{0.52} & \underline{0.52} 
        && \underline{0.65} & 0.63 
        && 0.71 & 0.72 
        && 10.4 
        && 0.31 & 0.21 \\
        \pockettwomol 
        & -4.93 & -4.77 
        && -6.16 & -5.95 
        && -6.93 & -6.77 
        && 35.8 & 17.7 
        && \textbf{0.62} & \textbf{0.62} 
        && \textbf{0.79} & \textbf{0.80} 
        && \textbf{0.80} & \textbf{0.81} 
        && \textbf{26.3} 
        && \underline{0.22} & \textbf{0.11} \\
        \diffsbdd 
        & -3.28 & -4.77 
        && -5.20 & -5.69 
        && -6.85 & -7.20 
        && 37.4 & 23.2 
        && 0.47 & 0.48 
        && 0.61 & 0.60 
        && \underline{0.77} & \underline{0.75} 
        && 9.1 
        && 0.30 & 0.21 \\
        \targetdiff 
        & -6.11 & \textbf{-6.62} 
        && \underline{-7.02} & \textbf{-7.17} 
        && \textbf{-8.22} & \textbf{-8.15} 
        && 51.0 & 50.6 
        && 0.45 & 0.45 
        && 0.57 & 0.57 
        && 0.71 & 0.71 
        && 12.3 
        && 0.35 & 0.26 \\
        \decompdiff 
        & -6.00 & -6.03 
        && -6.76 & -6.76 
        && -7.72 & -7.84 
        && 49.9 & 50.3 
        && 0.47 & 0.46 
        && 0.64 & 0.62 
        && 0.67 & 0.67 
        && 20.1 
        && 0.38 & 0.28 \\
        \decompopt 
        & -6.09 & -6.15 
        && -6.78 & -6.83 
        && -7.74 & -7.74 
        && 50.2 & 52.9 
        && 0.47 & 0.46 
        && \underline{0.65} & \underline{0.64} 
        && 0.67 & 0.68 
        && \underline{21.7} 
        && 0.37 & 0.26 \\
       \grey{\ipdiff}  
        & \grey{-7.70} & \grey{-7.70} 
        && \grey{-8.13} & \grey{-7.93} 
        && \grey{-8.97} & \grey{-8.59} 
        && \grey{63.4} & \grey{71.1} 
        && \grey{0.45} & \grey{0.45} 
        && \grey{0.54} & \grey{0.54} 
        && \grey{0.71} & \grey{0.71} 
        && \grey{8.5} 
        && \grey{0.43} & \grey{0.46} \\
		\midrule
        \method    & \textbf{-6.34} & \underline{-6.34} & & \textbf{-7.12} & \underline{-7.06} & & \underline{-8.13} & \underline{-8.10} & & \textbf{66.3} & \underline{64.6} & & {0.48} & {0.47} & & {0.62} & {0.60} & & {0.66} & {0.65} & & {21.5} && {0.43} & {0.44}\\
        \methodopt & \underline{-6.23} & -6.27 & & \underline{-7.02} & -7.02 & & -8.09 & -8.04 & & \underline{59.1} & \textbf{72.6} & & {0.47} & {0.45} & & {0.59} & {0.57} & & {0.65} & {0.63} & & {16.4} && \textbf{0.15} & \underline{0.12} \\
		\bottomrule
	\end{tabular}%
    \vspace{2pt}
	\begin{tablenotes}[flushleft]
	\footnotesize
	\item 
    {\!\!Columns represent: ``Vina S'': the predicted binding affinities between the initially generated poses of ligands and the protein pockets; 
		``Vina M'': the predicted binding affinities between the poses after local structure minimization and the protein pockets;
		``Vina D'': the predicted binding affinities between the poses determined by AutoDock Vina and the protein pockets;
		``HA'': the percentage of generated ligands with Vina D lower than those of reference ligands;
		``QED'': the quantitative estimate of drug-likeness;
		``SA'': the synthetic accessibility score;
		``Div'': the diversity among generated ligands;
            ``SR'': the percentage of generated molecules with binding affinities, QED, and SA above a certain threshold. 
            ``hERG'': the predicted hERG inhibition score for generated molecules;
            The best values are in \textbf{bold}, and the second-best values are
            \underline{underlined}. Rows in \grey{gray} are excluded from the comparison. 
            $\uparrow$ / $\downarrow$ indicate that higher / lower values are better.}
	\end{tablenotes}
    \end{minipage}}
\end{threeparttable}
  \vspace{-10pt}    
\end{table}

\begin{table}[H]
	\centering
	\caption{Comparison on \dataset with optimized property BBBP}
	\label{tbl:overall_results_bbbp}
\begin{threeparttable}
\resizebox{\textwidth}{!}{
\begin{minipage}{\textwidth}
	\begin{tabular}{
		@{\hspace{2pt}}l@{\hspace{2pt}}
		@{\hspace{2pt}}r@{\hspace{2pt}}
		@{\hspace{2pt}}r@{\hspace{2pt}}
		@{\hspace{4pt}}r@{\hspace{4pt}}
		@{\hspace{2pt}}r@{\hspace{2pt}}
		@{\hspace{2pt}}r@{\hspace{2pt}}
		@{\hspace{5pt}}r@{\hspace{5pt}}
		@{\hspace{2pt}}r@{\hspace{2pt}}
		@{\hspace{2pt}}r@{\hspace{2pt}}
		@{\hspace{5pt}}r@{\hspace{5pt}}
		@{\hspace{2pt}}r@{\hspace{2pt}}
	    @{\hspace{2pt}}r@{\hspace{2pt}}
		@{\hspace{5pt}}r@{\hspace{5pt}}
		@{\hspace{2pt}}r@{\hspace{2pt}}
		@{\hspace{2pt}}r@{\hspace{2pt}}
		@{\hspace{5pt}}r@{\hspace{5pt}}
		@{\hspace{2pt}}r@{\hspace{2pt}}
		@{\hspace{2pt}}r@{\hspace{2pt}}
		@{\hspace{5pt}}r@{\hspace{5pt}}
		@{\hspace{2pt}}r@{\hspace{2pt}}
		@{\hspace{2pt}}r@{\hspace{2pt}}
		@{\hspace{5pt}}r@{\hspace{5pt}}
		@{\hspace{2pt}}r@{\hspace{2pt}}
		@{\hspace{2pt}}r@{\hspace{2pt}}
		@{\hspace{2pt}}r@{\hspace{2pt}}
		@{\hspace{2pt}}r@{\hspace{2pt}}
		@{\hspace{2pt}}r@{\hspace{2pt}}
	}
	\toprule
	\multirow{2}{*}{Method} 
	& \multicolumn{2}{c}{Vina S$\downarrow$} & 
	& \multicolumn{2}{c}{Vina M$\downarrow$} & 
	& \multicolumn{2}{c}{Vina D$\downarrow$} & 
	& \multicolumn{2}{c}{{HA}\%$\uparrow$}  & 
	& \multicolumn{2}{c}{QED$\uparrow$} & 
	& \multicolumn{2}{c}{SA$\uparrow$} & 
	& \multicolumn{2}{c}{Div$\uparrow$} & 
	& \multirow{2}{*}{SR\%$\uparrow$} & 
	& \multicolumn{2}{c}{BBBP$\uparrow$} 
	\\
	\cmidrule{2-3}\cmidrule{5-6}\cmidrule{8-9}\cmidrule{11-12}
	\cmidrule{14-15}\cmidrule{17-18}\cmidrule{20-21}\cmidrule{25-26}
	& Avg. & Med. &  & Avg. & Med. &  & Avg. & Med. &
	  & Avg. & Med. &  & Avg. & Med. &  & Avg. & Med. &
	  & Avg. & Med. &  & & & Avg. & Med. \\
	\midrule
    Reference 
    & -7.03 & -7.06 
    && -7.61 & -7.26 
    && -7.72 & -7.53 
    && - & - 
    && 0.54 & 0.56 
    && 0.74 & 0.78 
    && - & - 
    && 28.0 
    && 0.66 & 0.70 \\
    \midrule
    \AR 
    & \textbf{-6.99} & \textbf{-6.87} 
    && \underline{-7.18} & -7.00 
    && -7.65 & -7.47 
    && 37.6 & 26.5 
    && 0.52 & 0.53 
    && 0.60 & 0.60 
    && 0.67 & 0.67 
    && 13.6 
    && 0.70 & 0.77 \\
    \pockettwomol 
    & -5.32 & -4.95 
    && -6.49 & -6.13 
    && -7.23 & -7.02 
    && 27.7 & 6.9 
    && \textbf{0.61} & \textbf{0.61} 
    && \textbf{0.79} & \textbf{0.79} 
    && \textbf{0.80} & \textbf{0.79} 
    && \textbf{33.5} 
    && \underline{0.76} & \underline{0.84} \\
    \diffsbdd 
    & -1.37 & -4.84 
    && -4.36 & -5.73 
    && -6.90 & -7.12 
    && 30.5 & 14.1 
    && 0.50 & 0.51 
    && 0.63 & 0.62 
    && \underline{0.79} & \textbf{0.79} 
    && 12.7 
    && 0.71 & 0.77 \\
    \targetdiff 
    & -6.21 & \underline{-6.75} 
    && -7.11 & \underline{-7.36} 
    && -8.34 & -8.26 
    && 49.6 & 45.9 
    && 0.47 & 0.47 
    && 0.58 & 0.57 
    && 0.70 & \underline{0.70} 
    && 12.7 
    && 0.68 & 0.72 \\
    \decompdiff 
    & -6.15 & -5.99 
    && -6.91 & -6.68 
    && -7.85 & -7.83 
    && 49.6 & 42.9 
    && 0.51 & 0.53 
    && 0.64 & 0.63 
    && 0.68 & \underline{0.70} 
    && 19.6 
    && 0.70 & 0.75 \\
    \decompopt 
    & \underline{-6.40} & -6.18 
    && -7.06 & -6.86 
    && -8.03 & -7.82 
    && 51.7 & 54.7 
    && 0.52 & 0.54 
    && \underline{0.65} & \underline{0.65} 
    && 0.68 & 0.68 
    && 23.7 
    && 0.69 & 0.75 \\
    \grey{\ipdiff} 
    & \grey{-7.82} & \grey{-8.26} 
    && \grey{-8.33} & \grey{-8.42} 
    && \grey{-9.37} & \grey{-9.02} 
    && \grey{63.0} & \grey{65.1} 
    && \grey{0.45} & \grey{0.45} 
    && \grey{0.56} & \grey{0.55} 
    && \grey{0.73} & \grey{0.71} 
    && \grey{10.3} 
    && \grey{0.80} & \grey{0.86} \\
	\midrule
	\method    & -6.18 & -6.43 && -7.16 & -7.27 && \underline{-8.47} & \underline{-8.46} && \underline{55.1} & \textbf{59.0} && {0.51} & {0.53} && {0.60} & {0.60} && {0.64} & {0.62} && {24.2} && {0.64} & {0.62} \\
	\methodopt & -6.30 & -6.58 && \textbf{-7.31} & \textbf{-7.39} && \textbf{-8.62} & \textbf{-8.58} && \textbf{59.0} & \underline{55.0} && \underline{0.53} & \underline{0.56} && {0.60} & {0.60} && {0.64} & {0.64} && \underline{25.9} && \textbf{0.89} & \textbf{0.90}\\
	\bottomrule
	\end{tabular}

    \vspace{2pt}
	\begin{tablenotes}[flushleft]
	\footnotesize
	\item 
    {\!\!Columns represent: ``Vina S'': the predicted binding affinities between the initially generated poses of ligands and the protein pockets; 
		``Vina M'': the predicted binding affinities between the poses after local structure minimization and the protein pockets;
		``Vina D'': the predicted binding affinities between the poses determined by AutoDock Vina and the protein pockets;
		``HA'': the percentage of generated ligands with Vina D lower than those of reference ligands;
		``QED'': the quantitative estimate of drug-likeness;
		``SA'': the synthetic accessibility score;
		``Div'': the diversity among generated ligands;
            ``SR'': the percentage of generated molecules with binding affinities, QED, and SA above a certain threshold. 
            ``BBBP'': the predicted blood–brain barrier permeability score for generated molecules;
            The best values are in \textbf{bold}, and the second-best values are
            \underline{underlined}. Rows in \grey{gray} are excluded from the comparison. 
            $\uparrow$ / $\downarrow$ indicate that higher / lower values are better.}
	\end{tablenotes}
\end{minipage}
}
\end{threeparttable}
\vspace{-10pt}
\end{table}

We also compare the 
{overall performance}
of \methodopt's and \method's generated ligands to assess whether 
{optimization using \opt}
preserves other essential characteristics of the initial generated ligands.
In general, the results show that ligands produced by \methodopt achieve Vina S/M/D, QED, and SA scores that are on par with, and in some cases {better than}, those of \method.
This indicates that \opt does not disrupt ligand generation ---
rather, it refines the process to improve ADMET properties
while retaining strong predicted binding affinities, drug-likeness, and synthetic accessibility.
In particular, as shown in Tables~\ref{tbl:overall_results_ames}--\ref{tbl:overall_results_herg}, optimizing Ames and hERG leads to only a slight decrease (less than 2.0\%) in Vina S/M/D scores, while the predicted binding affinities of \methodopt ligands remain superior to those from the baselines (except \ipdiff).
Meanwhile, optimizing BBBP and HIA slightly improves the predicted binding affinities of the generated ligands, as reported in Tables~\ref{tbl:overall_results_bbbp} and \ref{tbl:overall_results_hia}.
For QED and SA scores, \methodopt generally improves QED while maintaining SA at a level comparable to that of \method. 
This suggests
the optimization for ADMET properties tends to improve drug-likeness {and has negligible impact on synthetic accessibility.
{Please} note,
the diversity of \methodopt ligands remains similar to that of \method ligands, which is also reflected in Table~\ref{tbl:overall_results_carci}.}

\begin{table}[H]
	\centering
	\caption{Comparison on \dataset with optimized property HIA}
	\label{tbl:overall_results_hia}
\begin{threeparttable}
\resizebox{\textwidth}{!}{
\begin{minipage}{\textwidth}
	\begin{tabular}{
		@{\hspace{2pt}}l@{\hspace{2pt}}
		@{\hspace{2pt}}r@{\hspace{2pt}}
		@{\hspace{2pt}}r@{\hspace{2pt}}
		@{\hspace{4pt}}r@{\hspace{4pt}}
		@{\hspace{2pt}}r@{\hspace{2pt}}
		@{\hspace{2pt}}r@{\hspace{2pt}}
		@{\hspace{5pt}}r@{\hspace{5pt}}
		@{\hspace{2pt}}r@{\hspace{2pt}}
		@{\hspace{2pt}}r@{\hspace{2pt}}
		@{\hspace{5pt}}r@{\hspace{5pt}}
		@{\hspace{2pt}}r@{\hspace{2pt}}
	         @{\hspace{2pt}}r@{\hspace{2pt}}
		@{\hspace{5pt}}r@{\hspace{5pt}}
		@{\hspace{2pt}}r@{\hspace{2pt}}
		@{\hspace{2pt}}r@{\hspace{2pt}}
		@{\hspace{5pt}}r@{\hspace{5pt}}
		@{\hspace{2pt}}r@{\hspace{2pt}}
		@{\hspace{2pt}}r@{\hspace{2pt}}
		@{\hspace{5pt}}r@{\hspace{5pt}}
		@{\hspace{2pt}}r@{\hspace{2pt}}
		@{\hspace{2pt}}r@{\hspace{2pt}}
		@{\hspace{5pt}}r@{\hspace{5pt}}
		@{\hspace{2pt}}r@{\hspace{2pt}}
		@{\hspace{2pt}}r@{\hspace{2pt}}
		@{\hspace{2pt}}r@{\hspace{2pt}}
		@{\hspace{2pt}}r@{\hspace{2pt}}
		@{\hspace{2pt}}r@{\hspace{2pt}}
		}
		\toprule
		\multirow{2}{*}{Method} & \multicolumn{2}{c}{Vina S$\downarrow$} & & \multicolumn{2}{c}{Vina M$\downarrow$} & & \multicolumn{2}{c}{Vina D$\downarrow$} & & \multicolumn{2}{c}{{{HA}\%$\uparrow$}}  & & \multicolumn{2}{c}{QED$\uparrow$} & & \multicolumn{2}{c}{SA$\uparrow$} & & \multicolumn{2}{c}{Div$\uparrow$} & & \multirow{2}{*}{SR\%$\uparrow$} & 
		& \multicolumn{2}{c}{{HIA}$\uparrow$} 
        \\
	    \cmidrule{2-3}\cmidrule{5-6} \cmidrule{8-9} \cmidrule{11-12} \cmidrule{14-15} \cmidrule{17-18} \cmidrule{20-21} \cmidrule{25-26}  
		& Avg. & Med. &  & Avg. & Med. &  & Avg. & Med. & & Avg. & Med.  & & Avg. & Med.  & & Avg. & Med.  & & Avg. & Med. &  & & &  Avg. & Med. &  \\
		\midrule
        Reference 
        & -7.49 & -7.76 
        && -7.85 & -7.72 
        && -8.59 & -8.89 
        && - & - 
        && 0.39 & 0.39 
        && 0.70 & 0.73 
        && - & - 
        && 38.5 
        && 0.73 & 0.98 \\
        \midrule
        \AR 
        & -6.88 & -6.41 
        && -7.15 & -6.65 
        && -7.65 & -7.29 
        && 35.6 & 28.7 
        && 0.49 & 0.48 
        && 0.60 & 0.61 
        && 0.69 & 0.71 
        && 12.5 
        && 0.88 & \underline{0.99} \\
        \pockettwomol 
        & -5.16 & -4.95 
        && -6.19 & -6.08 
        && -6.98 & -6.93 
        && 25.3 & 13.8 
        && \textbf{0.61} & \textbf{0.62} 
        && \textbf{0.79} & \textbf{0.80} 
        && \textbf{0.80} & \textbf{0.81} 
        && \underline{27.0} 
        && \underline{0.94} & \textbf{1.00} \\
        \diffsbdd 
        & -4.98 & -5.40 
        && -6.07 & -6.12 
        && -7.38 & -7.47 
        && 30.3 & 24.5 
        && 0.47 & 0.47 
        && 0.62 & 0.60 
        && \underline{0.77} & \underline{0.75} 
        && 8.6 
        && 0.91 & \textbf{1.00} \\ 
        \targetdiff 
        & -6.37 & -6.43 
        && -6.95 & -6.94 
        && -7.85 & -8.01 
        && 36.6 & 22.6 
        && 0.46 & 0.46 
        && 0.59 & 0.58 
        && 0.72 & 0.73 
        && 11.2 
        && 0.92 & \textbf{1.00} \\
        \decompdiff 
        & -6.01 & -6.11 
        && -6.80 & -7.06 
        && -7.79 & -8.03 
        && 40.9 & 39.1 
        && 0.45 & 0.43 
        && 0.65 & 0.63 
        && 0.67 & 0.66 
        && 22.3 
        && 0.82 & \underline{0.99} \\
        \decompopt 
        & -6.22 & -6.36 
        && -6.91 & -7.10 
        && -7.79 & -8.15 
        && 41.1 & 37.7 
        && 0.46 & 0.45 
        && \underline{0.66} & \underline{0.65} 
        && 0.67 & 0.65 
        && 24.7 
        && 0.82 & \textbf{1.00} \\
        \grey{\ipdiff} 
        & \grey{-8.12} & \grey{-7.92} 
        && \grey{-8.35} & \grey{-8.08} 
        && \grey{-8.91} & \grey{-8.72} 
        && \grey{53.1} & \grey{49.5} 
        && \grey{0.43} & \grey{0.44} 
        && \grey{0.55} & \grey{0.54} 
        && \grey{0.71} & \grey{0.69} 
        && \grey{5.9} 
        && \grey{0.97} & \grey{1.00} \\
		\midrule
        \method  & \underline{-6.93} & \underline{-7.00} & & \underline{-7.44} & \underline{-7.44} & & \underline{-8.30} & \underline{-8.31} & & \underline{49.7} & \underline{52.7} & & {0.49} & {0.49} & & {0.63} & {0.63} & & {0.68} & {0.68} & & {24.6} && {0.93} & {0.98} \\
        \methodopt & \textbf{-7.09} & \textbf{-7.16} & & \textbf{-7.61} & \textbf{-7.65} & & \textbf{-8.39} & \textbf{-8.43} & & \textbf{52.6} & \textbf{58.3} & & \underline{0.53} & \underline{0.54} & & {0.64} & \underline{0.65} & & {0.67} & {0.67} & & \textbf{29.2} && \textbf{0.99} & \textbf{1.00} \\
		\bottomrule
	\end{tabular}%
    \vspace{2pt}
	\begin{tablenotes}[flushleft]
	\footnotesize
	\item 
    {\!\!Columns represent: ``Vina S'': the predicted binding affinities between the initially generated poses of ligands and the protein pockets; 
		``Vina M'': the predicted binding affinities between the poses after local structure minimization and the protein pockets;
		``Vina D'': the predicted binding affinities between the poses determined by AutoDock Vina and the protein pockets;
		``HA'': the percentage of generated ligands with Vina D lower than those of reference ligands;
		``QED'': the quantitative estimate of drug-likeness;
		``SA'': the synthetic accessibility score;
		``Div'': the diversity among generated ligands;
            ``SR'': the percentage of generated molecules with binding affinities, QED, and SA above a certain threshold. 
            ``HIA'': the predicted human intestinal absorption score for generated molecules;
            The best values are in \textbf{bold}, and the second-best values are
            \underline{underlined}. Rows in \grey{gray} are excluded from the comparison. 
            $\uparrow$ / $\downarrow$ indicate that higher / lower values are better.}
	\end{tablenotes}
    \end{minipage}}
\end{threeparttable}
  \vspace{-10pt}    
\end{table}

\section{Synthesis of \methodopt-generated Molecules}
\label{supp:synthesis}

\subsection{Synthesis of \PDLone Ligands}

\subsubsection{Synthesis of \cmpSixteen: (S)-5-(2-(difluoromethyl)-4'-fluoro-3'-methoxy-[1,1'-biphenyl]-3-yl)-2-(pyrrolidin-3-yl)isoindolin-1-one}
\label{supp:synthesis:cmp16}

The synthesis of \CmpSixteen was carried out in five steps.

\paragraph{Step 1: Preparation of tert-butyl (S)-3-(5-bromo-1-oxoisoindolin-2-yl)pyrrolidine-1-carboxylate (Figure~\ref{fig:pdl1_syn_2_1})}

\begin{figure}[H]
\centering
\includegraphics[width=0.3\linewidth]{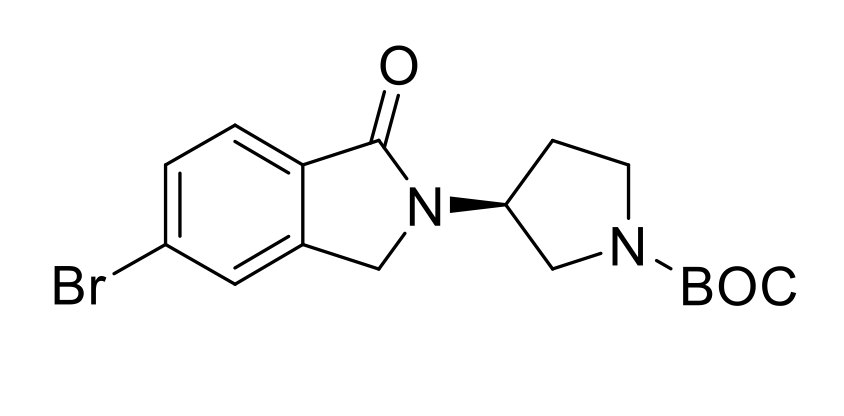}
\caption{Intermediate to be prepared in Step 1 for the synthesis of the \cmpSixteen}
\label{fig:pdl1_syn_2_1}
\end{figure}

Methyl 4-bromo-2-(bromomethyl)benzoate (1012 mg, 3.29 mmol) was added to a stirring solution of tert-butyl (3R)-3-aminopyrrolidine-1-carboxylate (510 mg, 2.74 mmol) and triethylamine (0.76 mL, 5.48 mmol) in acetonitrile (10 mL) at room temperature. Reaction mixture was stirred overnight at room temperature. LC/MS analysis indicated the formation of the desired product. Reaction was diluted with water and extracted with ethyl acetate. Combined organic extracts were washed with brine, dried over magnesium sulfate, filtered and concentrated. Purified on CombiFlash (24 g SiO2 gold column) using a 0-100\% ethyl acetate/heptanes gradient. Appropriate fractions were concentrated to provide tert-butyl (3S)-3-(5-bromo-1-oxo-isoindolin-2-yl)pyrrolidine-1-carboxylate as a white solid (747 mg, 72\% yield): $^1$H NMR (400 MHz, CDCl$_3$) $\delta$ 7.72 (d, J = 8.6 Hz, 1H), 7.62 (d, J = 6.9 Hz, 2H), 5.01 (p, J = 6.8 Hz, 1H), 4.38 (s, 2H), 3.87--3.20 (m, 4H), 2.27 (dq, J = 13.0, 6.8 Hz, 1H), 2.17--1.95 (m, 1H), 1.53--1.36 (m, 9H) ppm; ({M} + 1) = 381. 

\paragraph{Step 2: Preparation of tert-butyl (S)-3-(1-oxo-5-(4,4,5,5-tetramethyl-1,3,2-dioxaborolan-2-yl)isoindolin-2-\\yl)pyrrolidine-1-carboxylate (Figure~\ref{fig:pdl1_syn_2_2})} 

\begin{figure}[H]
\centering
\includegraphics[width=0.4\linewidth]{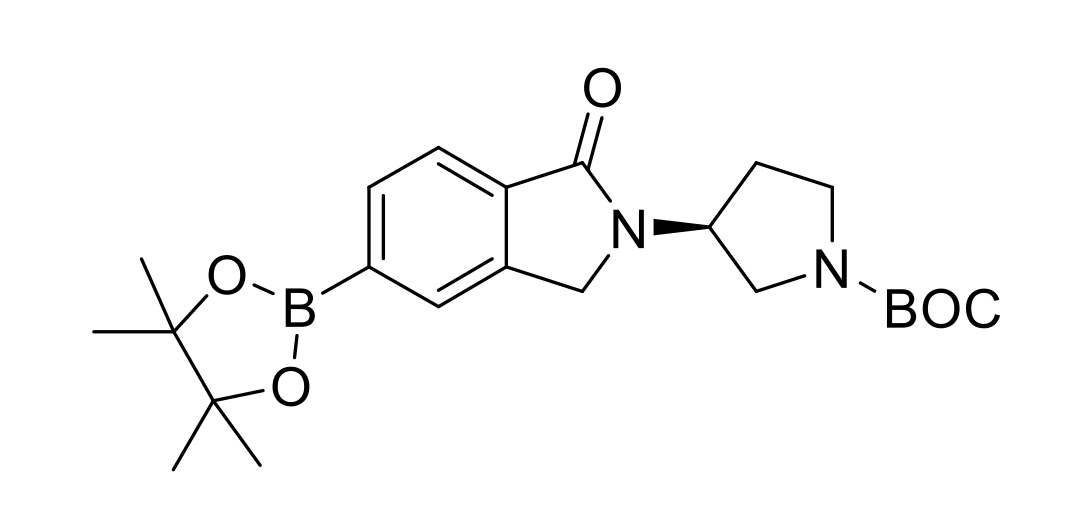}
\caption{Intermediate to be prepared in Step 2 for the synthesis of the \cmpSixteen}
\label{fig:pdl1_syn_2_2}
\end{figure}

Potassium acetate (577 mg, 5.88 mmol) was added to a stirring mixture of tert-butyl (3S)-3-(5-bromo-1-oxo-isoindolin-2-yl)pyrrolidine-1-carboxylate (747 mg, 1.96 mmol) and bis(pinacolato)diboron (1094 mg, 4.31 mmol) in 1,4-dioxane (10 ml). Nitrogen gas was bubbled through solution for 5 min, at which point \iupac{[1,1'-bis(diphenylphosphino)ferrocene]dichloropalladium(II)} (143 mg, 0.196 mmol) was added. Microwave vial was capped, bubbled nitrogen for 2 min then heated to $100^{\circ} \mathrm{C}$ for 2 h. LC/MS analysis indicated product formation. Cooled to room temperature, diluted with water and extracted with ethyl acetate. Combined organic extracts were washed with brine, dried over magnesium sulfate, filtered and concentrated. Purified by CombiFlash (12 g SiO2 gold column) using a 0-75\% ethyl acetate/heptanes gradient to provide tert-butyl (S)-3-(1-oxo-5-(4,4,5,5-tetramethyl-1,3,2-dioxaborolan-2-yl)isoindolin-2-yl)pyrrolidine-1-carboxylate (827 mg, 98\% yield): $^1$H NMR (400 MHz, CDCl$_3$) $\delta$ 8.23--7.58 (m, 3H), 5.03 (p, J = 6.9 Hz, 1H), 4.38 (s, 2H), 3.94--2.87 (m, 4H), 2.26 (dq, J = 12.6, 6.9 Hz, 1H), 2.10 (ddd, J = 19.0, 14.7, 6.9 Hz, 1H), 1.48 (s, 9H), 1.37 (s, 12H) ppm; ({M} + 1) = 429. 

\paragraph{Step 3: Preparation of \iupac{tert-butyl (S)-3-(5-(3-bromo-2-(difluoromethyl)phenyl)-1-oxoisoindolin-2-yl)pyrrolidine-1-carboxylate} (Figure~\ref{fig:pdl1_syn_2_3})}

\begin{figure}[H]
\centering
\includegraphics[width=0.4\linewidth]{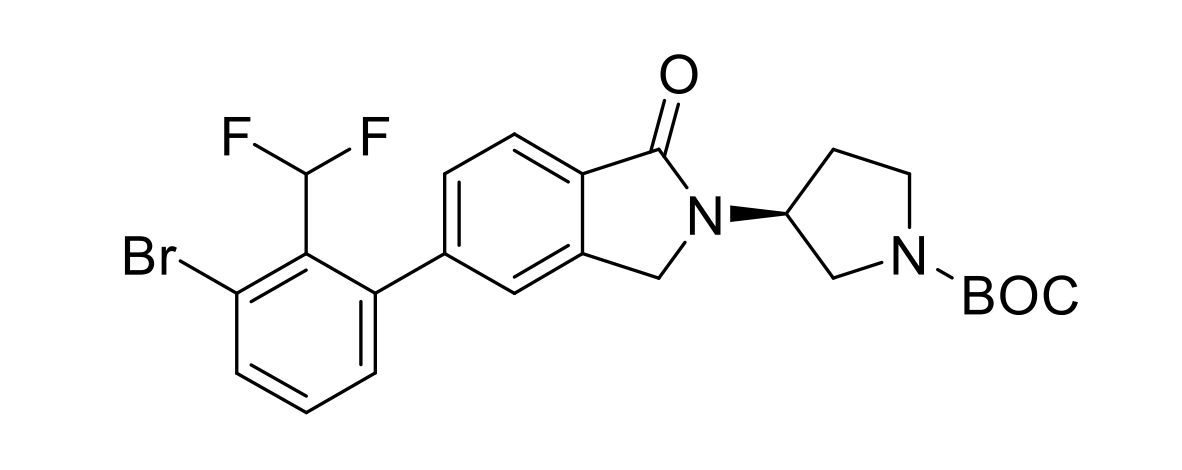}
\caption{Intermediate to be prepared in Step 3 for the synthesis of the \cmpSixteen}
\label{fig:pdl1_syn_2_3}
\end{figure}

To a microwave vial charged with tert-butyl 3-[(2S)-1-oxo-5-(4,4,5,5-tetramethyl-1,3,2-dioxaborolan-2-yl)isoindolin-2-yl]pyrrolidine-1-carboxylate (593 mg, 1.39 mmol), \iupac{1,3-dibromo-2-(difluoromethyl)benzene} (330 mg, 1.15 mmol), and cesium carbonate (3 equiv., 3.46 mmol) in 1,4-dioxane (8 mL) and water (2 mL) was added SPHOS PD G4 (92 mg, 0.12 mmol) and the vial was evacuated and purged with N2 (x3). The vial was sealed and the reaction was stirred at $90^{\circ} \mathrm{C}$ overnight. LC/MS analysis indicated the reaction was completed. Cooled to room temperature, diluted with water and extracted with dichloromethane. Combined organic extracts were washed with brine, dried over magnesium sulfate, filtered and concentrated. Purified on CombiFlash (24 g SiO2 gold column) using a 0-100\% ethyl acetate/heptanes gradient to provide tert-butyl (S)-3-(5-(3-bromo-2-(difluoromethyl)phenyl)-1-oxoisoindolin-2-yl)pyrrolidine-1-carboxylate (130 mg, 22\% yield): $^1$H NMR (400 MHz, CDCl$_3$) $\delta$ 7.90 (dd, J = 7.6, 1.0 Hz, 1H), 7.72 (dd, J = 8.0, 1.2 Hz, 1H), 7.43 (d, J = 8.5 Hz, 2H), 7.37 (dd, J = 8.4, 7.2 Hz, 1H), 7.25 (d, J = 7.4 Hz, 1H), 6.88 (t, J = 53.4 Hz, 1H), 5.07 (p, J = 6.9 Hz, 1H), 4.47 (s, 2H), 3.87--3.13 (m, 5H), 2.29 (dq, J = 13.2, 6.6 Hz, 1H), 2.23--2.06 (m, 1H), 1.48 (d, J = 5.9 Hz, 9H) ppm; ({M} + 1) = 509.

\paragraph{Step 4: Preparation of tert-butyl (S)-3-(5-(2-(difluoromethyl)-4'-fluoro-3'-methoxy-[1,1'-biphenyl]-3-yl)-1-\\oxoisoindolin-2-yl)pyrrolidine-1-carboxylate (Figure~\ref{fig:pdl1_syn_2_4})} 

\begin{figure}[H]
\centering
\includegraphics[width=0.45\linewidth]{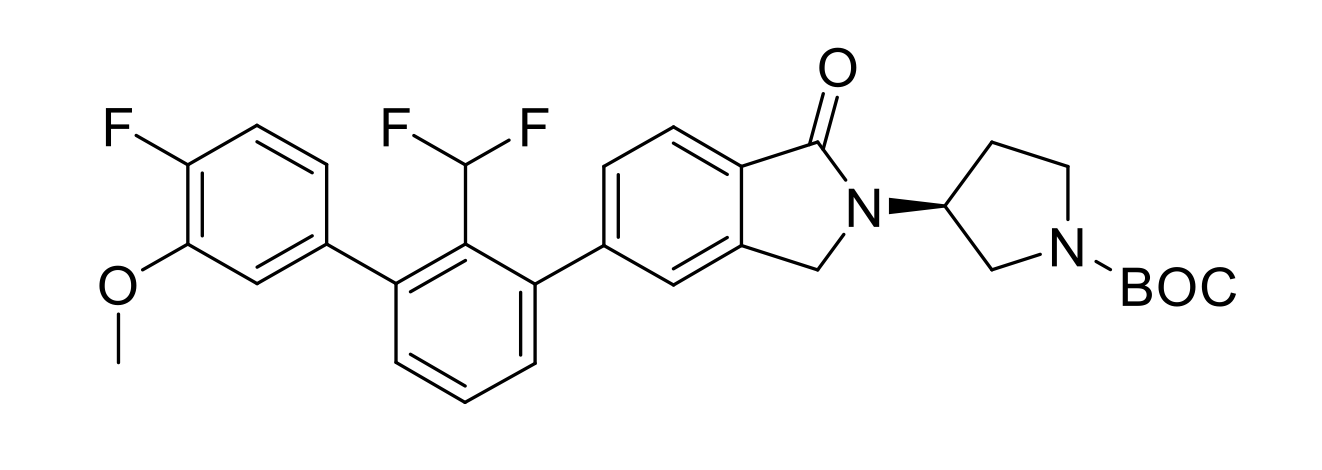}
\caption{Intermediate to be prepared in Step 4 for the synthesis of the \cmpSixteen}
\label{fig:pdl1_syn_2_4}
\end{figure}

To a vial charged with \iupac{tert-butyl (S)-3-(5-(3-bromo-2-(difluoromethyl)phenyl)-1-oxoisoindolin-2-yl)pyrrolidine-1-carboxylate} (65 mg, 0.128 mmol), \iupac{2-(4-fluoro-3-methoxy-phenyl)-4,4,5,5-tetramethyl-1,3,2-dioxaborolane} (32 mg, 0.128 mmol), and cesium carbonate (125 mg, 0.384 mmol) in 1,4-dioxane (2 mL) and water (0.5 mL) was added SPHOS PD G4 (10 mg, 0.0128 mmol) and the vial was evacuated and purged with N2 (x3). The vial was sealed and the reaction was stirred at $90^{\circ} \mathrm{C}$ overnight. LC/MS analysis indicated the reaction was completed. Cooled to room temperature, diluted with water and extracted with dichloromethane. Combined organic extracts were washed with brine, dried over magnesium sulfate, filtered and concentrated. Purified on CombiFlash (12 g SiO2 gold column) using a 0-100\% ethyl acetate/heptanes gradient to provide tert-butyl (S)-3-(5-(2-(difluoromethyl)-4'-fluoro-3'-methoxy-[1,1'-biphenyl]-3-yl)-1-oxoisoindolin-2-yl)pyrrolidine-1-carboxylate (65 mg, 91\% yield): $^1$H NMR (400 MHz, CDCl$_3$) $\delta$ 7.96--7.85 (m, 1H), 7.59--7.48 (m, 3H), 7.39 (d, J = 7.6 Hz, 1H), 7.33 (dd, J = 7.6, 1.2 Hz, 1H), 7.16 (dd, J = 11.1, 8.3 Hz, 1H), 7.04 (dd, J = 8.1, 2.1 Hz, 1H), 6.94 (ddd, J = 8.2, 4.2, 2.1 Hz, 1H), 6.46 (t, J = 53.8 Hz, 1H), 5.07 (p, J = 6.7 Hz, 1H), 4.47 (d, J = 5.8 Hz, 2H), 3.93 (s, 3H), 3.81--3.28 (m, 4H), 2.28 (dt, J = 14.7, 6.5 Hz, 1H), 2.23--2.06 (m, 1H), 1.48 (t, J = 5.5 Hz, 9H) ppm; ({M} + 1) = 553.

\paragraph{Step 5: Preparation of \iupac{(S)-5-(2-(difluoromethyl)-4'-fluoro-3'-methoxy-[1,1'-biphenyl]-3-yl)-2-(pyrrolidin-3-yl)isoindolin-1-one} (Figure~\ref{fig:pdl1_syn_2_5})} 
\begin{figure}[H]
\centering
\includegraphics[width=0.45\linewidth]{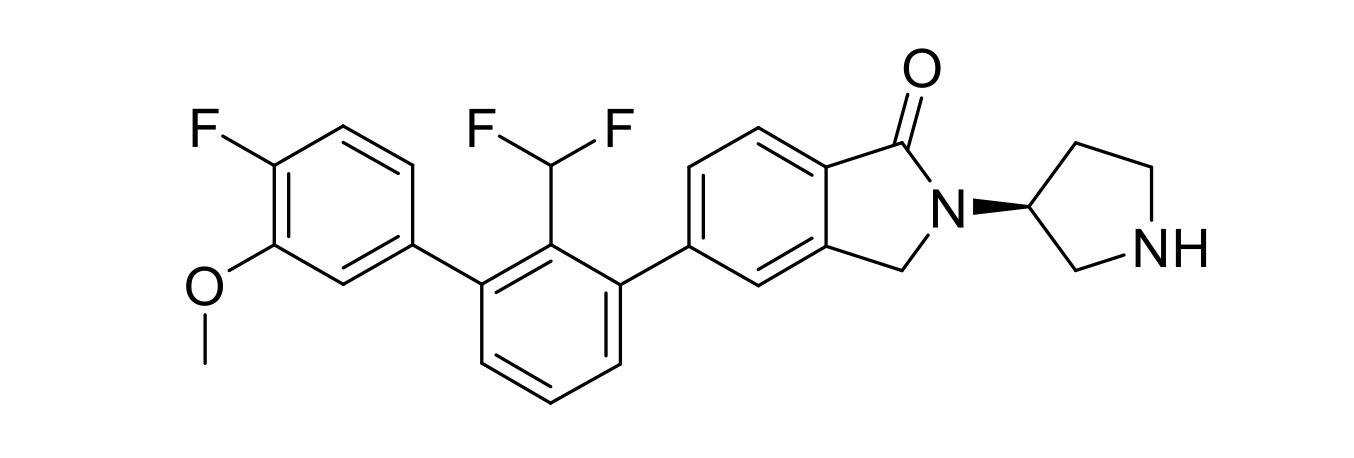}
\caption{Final compound (\cmpSixteen) to be synthesized {in Step 5}}
\label{fig:pdl1_syn_2_5}
\end{figure}

TFA (2 mL, 26.0 mmol) was added to a stirring mixture of tert-butyl (3S)-3-[5-[2-(difluoromethyl)-3-(4-fluoro-3-methoxy-phenyl)phenyl]-1-oxo-isoindolin-2-yl]pyrrolidine-1-carboxylate (65 mg, 0.118 mmol) in dichloromethane (4 mL) at $0^{\circ} \mathrm{C}$. Reaction was stirred for 2 h at room temperature. LC/MS analysis indicated the reaction was completed. Reaction mixture was concentrated. Residue was partitioned between NaHCO3 and dichloromethane. Combined organic extracts were dried over magnesium sulfate, filtered and concentrated. Purified on CombiFlash (12 g SiO2 gold column) using a 0-10\% methanol/ dichloromethane gradient to provide \iupac{5-[2-(difluoromethyl)-3-(4-fluoro-3-methoxy-phenyl)phenyl]-2-[(3S)-pyrrolidin-3-yl]isoindolin-1-one} (18 mg, 34\% yield) as a white solid: $^1$H NMR (400 MHz, DMSO) $\delta$ 7.82--7.54 (m, 1H), 7.45 (d, J = 7.9 Hz, 1H), 7.33 (d, J = 8.5 Hz, 1H), 6.59 (d, J = 52.9 Hz, 1H), 4.89 (s, 2H), 4.58 (m, 4H), 3.87 (s, 3H), 3.24 (m, 1H), 2.93 (m, 2H), 2.71 (m, 1H), 2.06 (m, 2H) ppm; ({M} + 1) = 453.

\subsubsection{Synthesis of \cmpFifteen: (R)-5-(2-(difluoromethyl)-4'-fluoro-3'-methoxy-[1,1'-biphenyl]-3-yl)-2-(pyrrolidin-3-yl)isoindolin-1-one}
\label{supp:synthesis:cmp15}

The synthesis of \CmpFifteen was carried out in five steps.

\paragraph{Step 1: Preparation of tert-butyl (R)-3-(5-bromo-1-oxoisoindolin-2-yl)pyrrolidine-1-carboxylate (Figure~\ref{fig:pdl1_syn_1_1})}

\begin{figure}[H]
\centering
\includegraphics[width=0.35\linewidth]{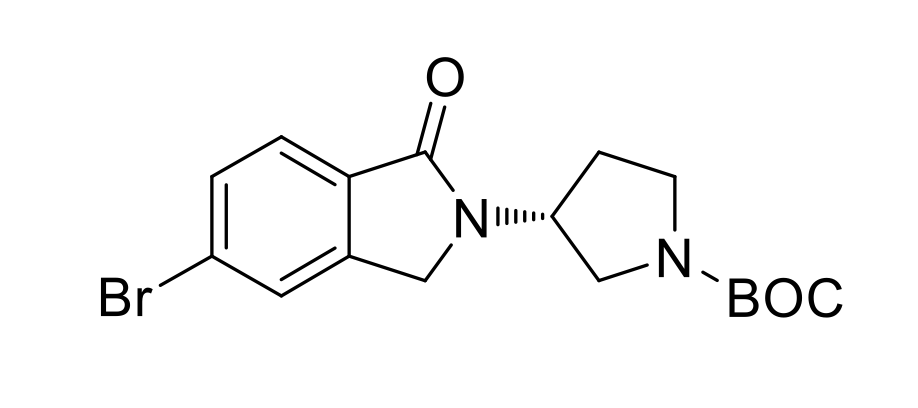}
\caption{Intermediate to be prepared in Step 1 for the synthesis of the \cmpFifteen}
\label{fig:pdl1_syn_1_1}
\end{figure}

Methyl 4-bromo-2-(bromomethyl)benzoate (873 mg, 2.83 mmol) was added to a stirring solution of (S)-(-)-1-BOC-3-aminopyrrolidine (440 mg, 2.36 mmol) and triethylamine (0.66 mL, 4.72 mmol) in acetonitrile (10 mL) at room temperature. Reaction mixture was stirred overnight at room temperature. LC/MS analysis indicated the formation of the desired product. Reaction was diluted with water and extracted with ethyl acetate. Combined organic extracts were washed with brine, dried over magnesium sulfate, filtered and concentrated. Purified on CombiFlash (24 g SiO2 gold column) using a 0-100\% ethyl acetate/heptanes gradient. Appropriate fractions were concentrated to provide tert-butyl (R)-3-(5-bromo-1-oxoisoindolin-2-yl)pyrrolidine-1-carboxylate (700 mg, 77\% yield) as a white solid: $^1$H NMR (400 MHz, CDCl$_3$) $\delta$ 7.72 (d, J = 8.6 Hz, 1H), 7.67--7.57 (m, 2H), 5.01 (p, J = 6.8 Hz, 2H), 4.38 (s, 2H), 3.91--3.10 (m, 4H), 2.27 (dq, J = 13.0, 6.8 Hz, 1H), 2.21--1.96 (m, 1H), 1.55 (s, 9H) ppm; ({M} + 1) = 381.

\paragraph{Step 2: Preparation of {tert-butyl (R)-3-(1-oxo-5-(4,4,5,5-tetramethyl-1,3,2-dioxaborolan-2-yl)isoindolin-2\\-yl)pyrrolidine-1-carboxylate} (Figure~\ref{fig:pdl1_syn_1_2})}

\begin{figure}[H]
\centering
\includegraphics[width=0.35\linewidth]{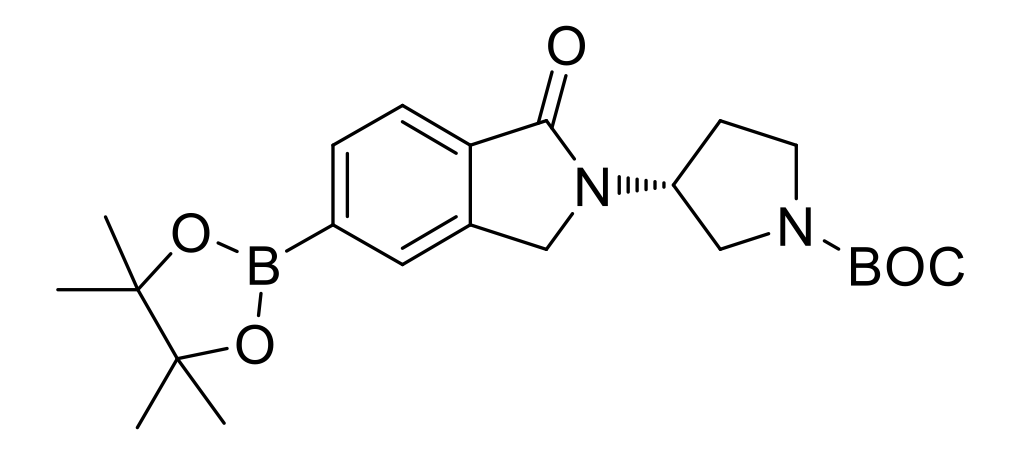}
\caption{Intermediate to be prepared in Step 2 for the synthesis of the \cmpFifteen}
\label{fig:pdl1_syn_1_2}
\end{figure}

Potassium acetate (541 mg, 5,51 mmol) was added to a stirring mixture of tert-butyl (R)-3-(5-bromo-1-oxoisoindolin-2-yl)pyrrolidine-1-carboxylate (700 mg, 1.84 mmol) and bis(pinacolato)diboron (1026 mg, 4.04 mmol) in 1,4-dioxane (10 ml). Nitrogen gas was bubbled through solution for 5 min, at which point
\iupac{[1,1'-bis(diphenylphosphino)ferrocene]dichloropalladium(II)} (134 mg, 0.184 mmol) was added. Microwave vial was capped, bubbled nitrogen for 2 min then heated to $100^{\circ} \mathrm{C}$ for 2 h. LC/MS analysis indicated product formation. Cooled to room temperature, diluted with water and extracted with ethyl acetate. Combined organic extracts were washed with brine, dried over magnesium sulfate, filtered and concentrated. Purified by CombiFlash (12 g SiO2 gold column) using a 0-20\% ethyl acetate/heptanes gradient to provide tert-butyl (R)-3-(1-oxo-5-(4,4,5,5-tetramethyl-1,3,2-dioxaborolan-2-yl)isoindolin-2-yl)pyrrolidine-1-carboxylate (700 mg, 89\% yield): $^1$H NMR (400 MHz, CDCl$_3$) $\delta$ 8.05--7.65 (m, 3H), 5.03 (p, J = 7.0 Hz, 1H), 4.38 (s, 2H), 3.94--3.15 (m, 4H), 2.34--2.03 (m, 2H), 1.37 (s, 9H) ppm; ({M} + 1) = 429.

\paragraph{Step 3: Preparation of \iupac{tert-butyl (R)-3-(5-(3-bromo-2-(difluoromethyl)phenyl)-1-oxoisoindolin-2-yl) pyrrolidine-1-carboxylate (Figure~\ref{fig:pdl1_syn_1_3})}} 
\begin{figure}[H]
\centering
\includegraphics[width=0.35\linewidth]{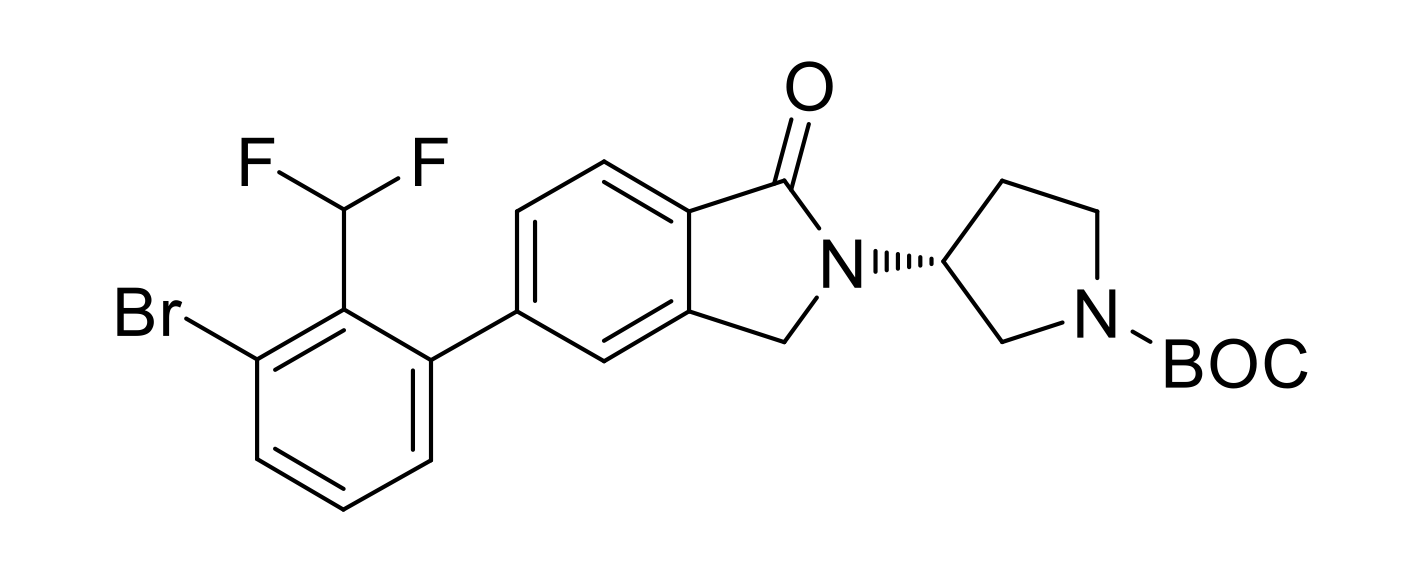}
\caption{Intermediate to be prepared in Step 3 for the synthesis of the \cmpFifteen}
\label{fig:pdl1_syn_1_3}
\end{figure}

To a vial charged with \iupac{tert-butyl (R)-3-(1-oxo-5-(4,4,5,5-tetramethyl-1,3,2-dioxaborolan-2-yl)isoindolin-2-yl)pyrrolidine-1-carboxylate} (50 mg, 0.117 mmol), 
\iupac{1,3-dibromo-2-(difluoromethyl)benzene} (33 mg, 0.117 mmol), and cesium carbonate (114 mg, 0.350 mmol) in 1,4-dioxane (4 ml) and water (1 ml) was added SPHOS PD G4 (9 mg, 0.012 mmol) and the vial was evacuated and purged with N2 (x3). The vial was sealed and the reaction was stirred at $90^{\circ} \mathrm{C}$ overnight. LC/MS analysis indicated the reaction was completed. Cooled to room temperature, diluted with water and extracted with dichloromethane. Combined organic extracts were washed with brine, dried over magnesium sulfate, filtered and concentrated. Purified on CombiFlash (12 g SiO2 gold column) using a 0-5\% methanol/dichloromethane gradient to provide \iupac{tert-butyl (R)-3-(5-(3-bromo-2-(difluoromethyl)phenyl)-1-oxoisoindolin-2-yl)pyrrolidine-1-carboxylate} (25 mg, 42\% yield): $^1$H NMR (400 MHz, CDCl$_3$) $\delta$ 7.90 (dd, J = 7.6, 1.0 Hz, 1H), 7.72 (dd, J = 8.0, 1.2 Hz, 1H), 7.43 (d, J = 8.5 Hz, 2H), 7.37 (dd, J = 8.4, 7.2 Hz, 1H), 7.25 (d, J = 7.4 Hz, 1H), 6.88 (t, J = 53.4 Hz, 1H), 5.07 (p, J = 6.9 Hz, 1H), 4.47 (s, 2H), 3.87--3.13 (m, 5H), 2.29 (dq, J = 13.2, 6.6 Hz, 1H), 2.23--2.06 (m, 1H), 1.48 (d, J = 5.9 Hz, 9H) ppm; ({M} + 1) = 509. 

\paragraph{Step 4: Preparation of \iupac{tert-butyl (R)-3-(5-(2-(difluoromethyl)-4'-fluoro-3'-methoxy-[1,1'-biphenyl]-3-yl)-1-oxoisoindolin-2-yl)pyrrolidine-1-carboxylate}  (Figure~\ref{fig:pdl1_syn_1_4})}

\begin{figure}[H]
\centering
\includegraphics[width=0.45\linewidth]{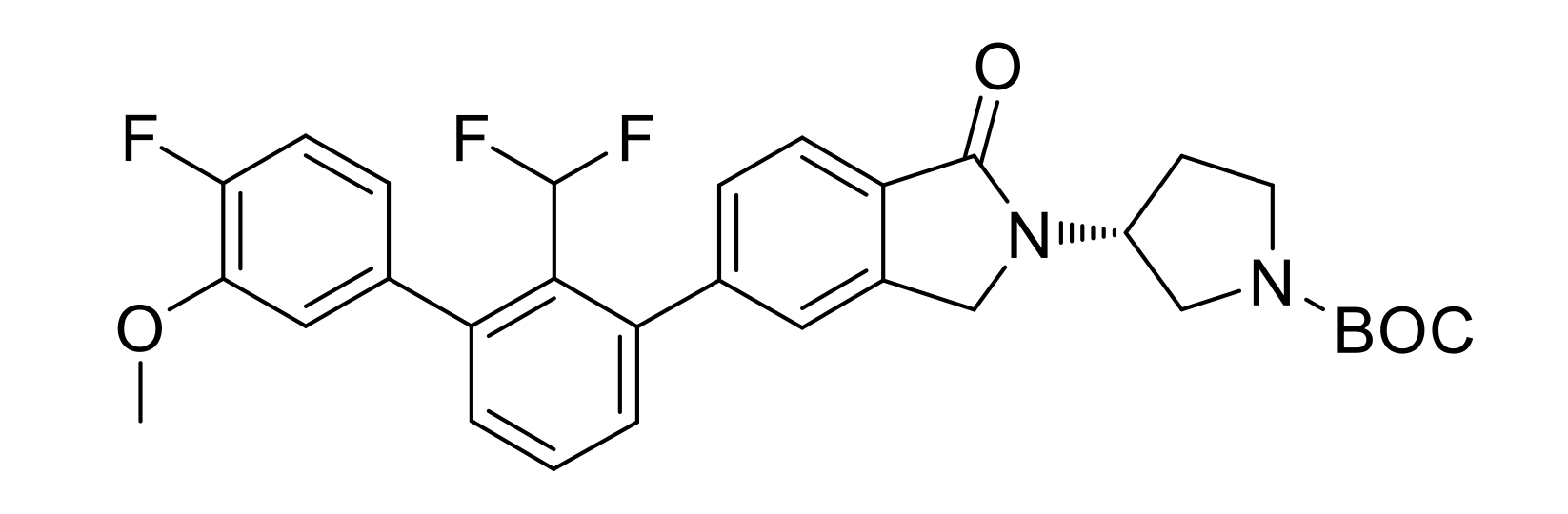}
\caption{Intermediate to be prepared in Step 4 for the synthesis of the \cmpFifteen}
\label{fig:pdl1_syn_1_4}
\end{figure}

To a vial charged with \iupac{tert-butyl (R)-3-(5-(3-bromo-2-(difluoromethyl)phenyl)-1-oxoisoindolin-2-yl)pyrrolidine-1-carboxylate} (25 mg, 0.049 mmol), \iupac{2-(4-fluoro-3-methoxy-phenyl)-4,4,5,5-tetramethyl-1,3,2-dioxaborolane} (1 equiv., 0.0493 mmol), and cesium carbonate (3 equiv., 0.1478 mmol) in 1,4-dioxane (2 mL, 23.4 mmol) and water (0.5 mL) was added SPHOS PD G4 (0.1 equiv., 0.0049 mmol) and the vial was evacuated and purged with N2 (x3). The vial was sealed and the reaction was stirred at $90^{\circ} \mathrm{C}$ overnight. LC/MS analysis indicated the reaction was completed. Cooled to room temperature, diluted with water and extracted with dichloromethane. Combined organic extracts were washed with brine, dried over magnesium sulfate, filtered and concentrated. Purified on CombiFlash (12 g SiO2 gold column) using a 0-5\% methanol/dichloromethane gradient to provide tert-butyl (3R)-3-[5-[2-(difluoromethyl)-3-(4-fluoro-3-methoxy-phenyl)phenyl]-1-oxo-isoindolin-2-yl]pyrrolidine-1-carboxylate (15 mg, 0.0271 mmol, 55\% yield): $^1$H NMR (400 MHz, CDCl$_3$) $\delta$ 7.96--7.85 (m, 1H), 7.59--7.48 (m, 3H), 7.39 (d, J = 7.6 Hz, 1H), 7.33 (dd, J = 7.6, 1.2 Hz, 1H), 7.16 (dd, J = 11.1, 8.3 Hz, 1H), 7.04 (dd, J = 8.1, 2.1 Hz, 1H), 6.94 (ddd, J = 8.2, 4.2, 2.1 Hz, 1H), 6.46 (t, J = 53.8 Hz, 1H), 5.07 (p, J = 6.7 Hz, 1H), 4.47 (d, J = 5.8 Hz, 2H), 3.93 (s, 3H), 3.81--3.28 (m, 4H), 2.28 (dt, J = 14.7, 6.5 Hz, 1H), 2.23--2.06 (m, 1H), 1.48 (t, J = 5.5 Hz, 9H) ppm; ({M} + 1) = 553.

\paragraph{Step 5: Preparation of \iupac{(R)-5-(2-(difluoromethyl)-4'-fluoro-3'-methoxy-[1,1'-biphenyl]-3-yl)-2-(pyrrolidin-3-yl)isoindolin-1-one} (Figure~\ref{fig:pdl1_syn_1_5})}

\begin{figure}[H]
\centering
\includegraphics[width=0.4\linewidth]{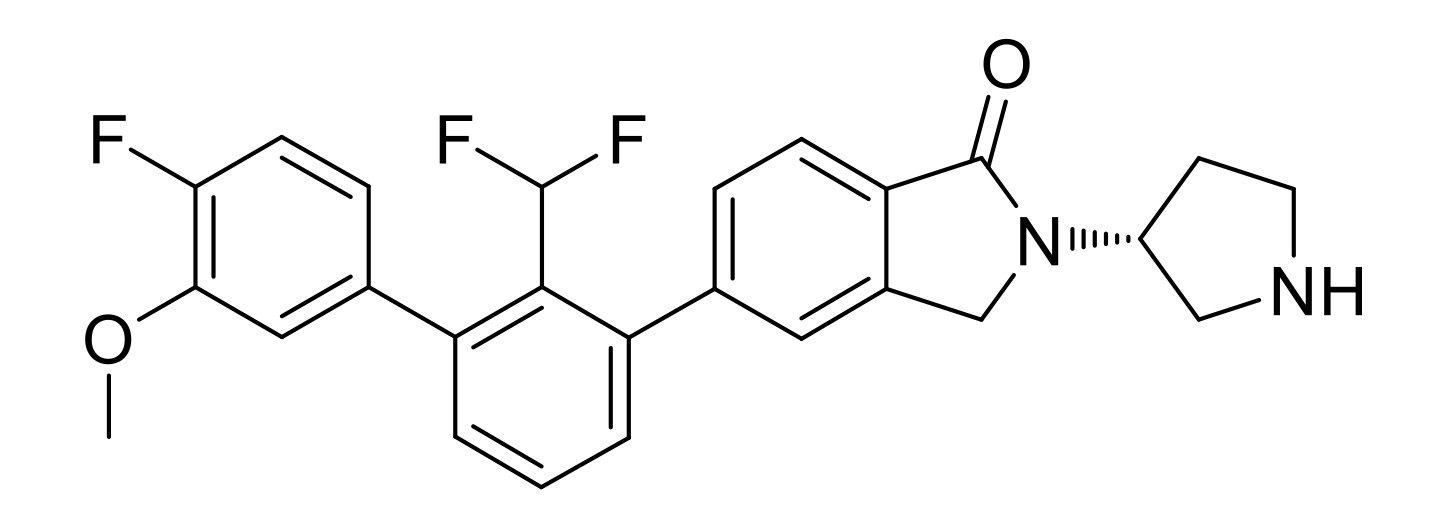}
\caption{Final compound (\cmpFifteen) to be synthesized {in Step 5}}
\label{fig:pdl1_syn_1_5}
\end{figure}

TFA (2 mL, 26.0 mmol) was added to a stirring solution of tert-butyl (3R)-3-[5-[2-(difluoromethyl)-3-(4-fluoro-3-methoxy-phenyl)phenyl]-1-oxo-isoindolin-2-yl]pyrrolidine-1-carboxylate (15 mg, 0.0271 mmol) in dichloromethane (4 mL, 62.40 mmol) at $0^{\circ} \mathrm{C}$. Stirred for 2 h with warming to room temperature. LC/MS analysis indicated the reaction was completed. Concentrated the reaction mixture. The residue was partitioned between dichloromethane and saturated sodium bicarbonate solution, then extracted with dichloromethane. Combined organic extracts were washed with brine, dried over magnesium sulfate, filtered and concentrated. Preparative RP chromatography (Interchim puriFlash XS 530, RediSep Gold C18 50 g column, 20\% acetonitrile/water/0.1\% formic acid to 50\% acetonitrile/0.1\% formic acid elute) provided 5-[2-(difluoromethyl)-3-(4-fluoro-3-methoxy-phenyl)phenyl]-2-[(3R)-pyrrolidin-3-yl]isoindolin-1-one;formic acid (10 mg, 74\% yield): $^1$H NMR (400 MHz, DMSO) $\delta$ 8.22 (s, 1H), 7.68 (d, J = 7.8 Hz, 1H), 7.61 (t, J = 7.7 Hz, 1H), 7.54 (s, 1H), 7.42 (t, J = 7.7 Hz, 2H), 7.33 (d, J = 7.6 Hz, 1H), 7.26 (dd, J = 11.6, 8.3 Hz, 1H), 7.12 (dd, J = 8.3, 2.1 Hz, 1H), 6.88 (ddd, J = 8.4, 4.3, 2.1 Hz, 1H), 6.50 (t, J = 53.3 Hz, 1H), 4.72 (q, J = 6.8 Hz, 1H), 4.52 (s, 2H), 3.81 (s, 3H), 3.20--2.78 (m, 4H), 1.95 (ddd, J = 60.4, 13.4, 6.2 Hz, 2H) ppm; ({M} + 1) = 453.

\subsubsection{Synthesis of \CmpEighteen: (S)-5-(2-(difluoromethyl)-3-(indolin-5-yl)phenyl)-2-(pyrrolidin-3-yl)isoindolin-1-one}
\label{supp:synthesis:cmp18}

The synthesis of \CmpEighteen was carried out in two steps.

\paragraph{Step 1: Preparation of \iupac{tert-butyl (S)-5-(3-(2-(1-(tert-butoxycarbonyl)pyrrolidin-3-yl)-1-oxoisoindolin-5-yl)-2-(difluoromethyl)phenyl)indoline-1-carboxylate} (Figure~\ref{fig:pdl1_syn_4_1})}

\begin{figure}[H]
\centering
\includegraphics[width=0.5\linewidth]{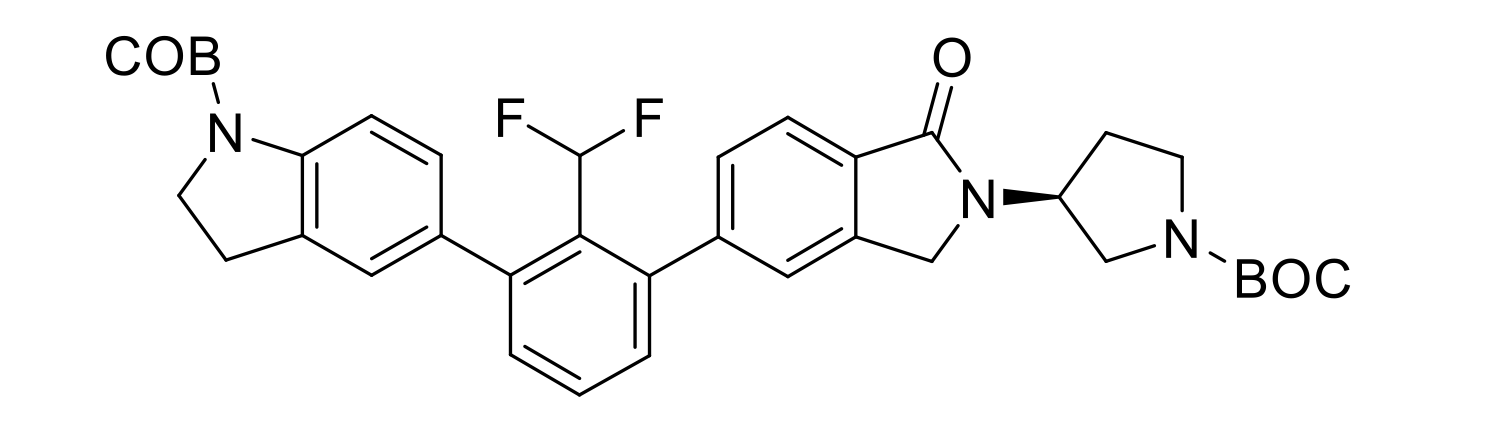}
\caption{Intermediate to be prepared in Step 1 for the synthesis of the \cmpEighteen}
\label{fig:pdl1_syn_4_1}
\end{figure}

To a vial charged with \iupac{tert-butyl (3S)-3-[5-[3-bromo-2-(difluoromethyl)phenyl]-1-oxo-isoindolin-2-yl]pyrrolidine-1-carboxylate} (128 mg, 0.25 mmol),
\iupac{1-(tert-butoxycarbonyl)-5-indolineboronic acid} (80 mg, 0.30 mmol), and cesium carbonate (247 mg, 0.75 mmol) in 1,4-dioxane (3 mL) and water (0.75 mL) was added SPHOS PD G4 (20 mg, 0.025 mmol) and the vial was evacuated and purged with N2 (x3). The vial was sealed and the reaction was stirred at $90^{\circ} \mathrm{C}$ overnight. LC/MS analysis indicated the reaction was completed. Cooled to room temperature, diluted with water and extracted with dichloromethane. Combined organic extracts were washed with brine, dried over magnesium sulfate, filtered and concentrated. Purified on CombiFlash (12 g SiO2 gold column) using a 0-5\% methanol/ dichloromethane gradient to provide \iupac{tert-butyl 5-[3-[2-[(3S)-1-tert-butoxycarbonylpyrrolidin-3-yl]-1-oxo-isoindolin-5-yl]-2-(difluoromethyl)phenyl]indoline-1-carboxylate} (128 mg, 78\% yield): $^1$H NMR (400 MHz, CDCl$_3$) $\delta$ 7.90 (d, J = 8.2 Hz, 1H), 7.59--7.48 (m, 3H), 7.37 (d, J = 7.7 Hz, 1H), 7.27 (m, 2H), 7.25--7.18 (m, 2H), 6.50 (t, J = 53.9 Hz, 1H), 5.07 (p, J = 6.8 Hz, 1H), 4.47 (s, 2H), 4.12--3.98 (m, 2H), 3.88--3.30 (m, 4H), 3.17 (t, J = 8.7 Hz, 2H), 2.42--2.06 (m, 2H), 1.59 (m, 18H) ppm; ({M} + 1) = 646.

\paragraph{Step 2: Preparation of (S)-5-(2-(difluoromethyl)-3-(indolin-5-yl)phenyl)-2-(pyrrolidin-3-yl)isoindolin-1-one (Figure~\ref{fig:pdl1_syn_4_2})}

\begin{figure}[H]
\centering
\includegraphics[width=0.45\linewidth]{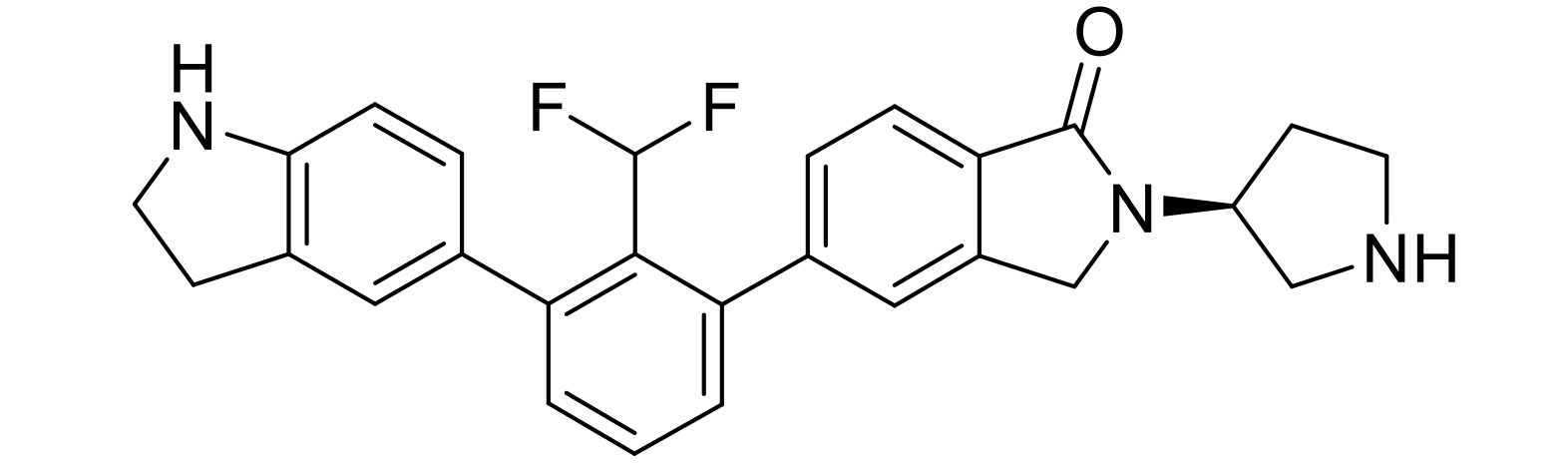}
\caption{Final compound (\cmpEighteen) to be synthesized {in Step 2}}
\label{fig:pdl1_syn_4_2}
\end{figure}

TFA (4 mL) was added to a stirring mixture of tert-butyl \iupac{5-[3-[2-[(3S)-1-tert-47ybutoxycarbonylpyrrolidin-3-yl]-1-oxo-isoindolin-5-yl]-2-(difluoromethyl)phenyl]indoline-1-carboxylate} (128 mg, 0.198 mmol) in dichloromethane (8 mL) at $0^{\circ} \mathrm{C}$. Reaction was stirred for 2 h at room temperature. LC/MS analysis indicated the reaction was completed. Reaction mixture was concentrated. Residue was partitioned between NaHCO3 and dichloromethane. Combined organic extracts were dried over magnesium sulfate, filtered and concentrated. Purified by preparative RP chromatography (Interchim puriFlash XS 530, RediSep Gold C18 50 g column, 10\% acetonitrile/water/0.1\% formic acid to 50\% acetonitrile/0.1\% formic acid elute) to provide 5-[2-(difluoromethyl)-3-indolin-5-yl-phenyl]-2-[(3S)-pyrrolidin-3-yl]isoindolin-1-one (75 mg, 77\% yield): $^1$H NMR (400 MHz, DMSO) $\delta$ 8.36 (s, 1H), 7.73 (d, J = 7.8 Hz, 1H), 7.68--7.56 (m, 2H), 7.54--7.48 (m, 1H), 7.39 (d, J = 7.6 Hz, 1H), 7.27 (dd, J = 7.6, 1.3 Hz, 1H), 7.07 (d, J = 1.8 Hz, 1H), 6.94 (dd, J = 7.9, 1.9 Hz, 1H), 6.79--6.30 (m, 2H), 4.83 (m, 1H), 4.60 (d, J = 2.1 Hz, 2H), 3.34--2.81 (m, 6H), 2.25--1.84 (m, 4H) ppm; ({M} + 1) = 446.

\subsubsection{Synthesis of \cmpSeventeen: 5,5'-(2-(difluoromethyl)-1,3-phenylene)bis(2-((S)-pyrrolidin-3-yl)isoindolin-1-one)}
\label{supp:synthesis:cmp17}

The synthesis of \cmpSeventeen was carried out in two steps.

\paragraph{Step 1: Preparation of di-tert-butyl 3,3'-((2-(difluoromethyl)-1,3-phenylene)bis(1-oxoisoindoline-5,2-diyl)) (3S,3'S)-bis(pyrrolidine-1-carboxylate) (Figure~\ref{fig:pdl1_syn_3_1})}


\begin{figure}[H]
\centering
\includegraphics[width=0.60\linewidth]{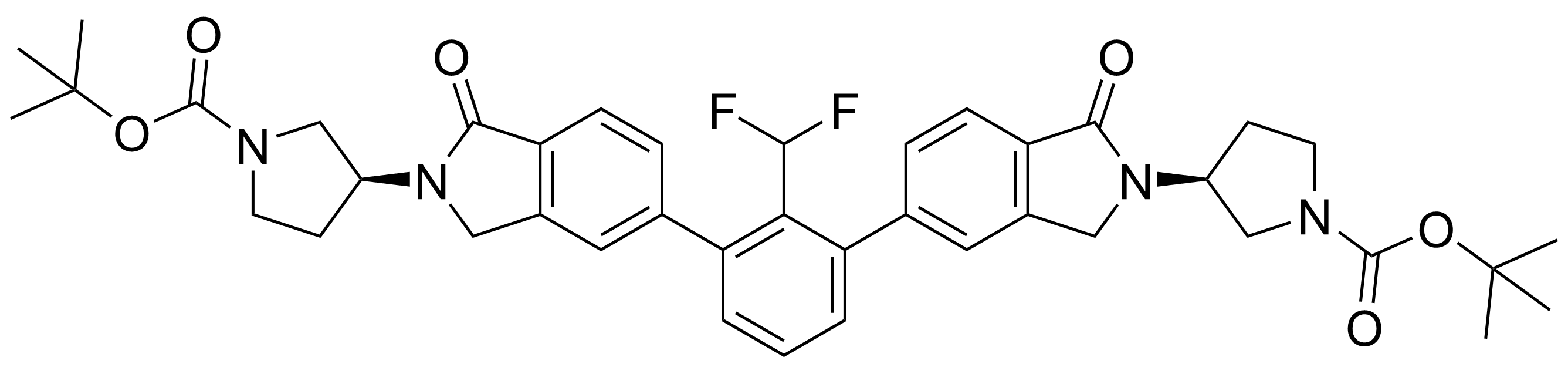}
\caption{Intermediate to be prepared in Step 1 for the synthesis of the \cmpSeventeen}
\label{fig:pdl1_syn_3_1}
\end{figure}

To a microwave vial charged with tert-butyl 3-[(2S)-1-oxo-5-(4,4,5,5-tetramethyl-1,3,2-dioxaborolan-2-yl)isoindolin-2-yl]pyrrolidine-1-carboxylate (593 mg, 1.39 mmol), 1,3-dibromo-2-(difluoromethyl)benzene (330 mg, 1.15 mmol), and cesium carbonate (3 equiv., 3.46 mmol) in 1,4-dioxane (8 mL) and water (2 mL) was added SPHOS PD G4 (92 mg, 0.12 mmol) and the vial was evacuated and purged with N2 (x3). The vial was sealed and the reaction was stirred at $90^{\circ} \mathrm{C}$ overnight. LC/MS analysis indicated the reaction was completed. Cooled to room temperature, diluted with water and extracted with dichloromethane. Combined organic extracts were washed with brine, dried over magnesium sulfate, filtered and concentrated. Purified on CombiFlash (24 g SiO2 gold column) using a 0-100\% ethyl acetate/heptanes gradient to provide tert-butyl 3-[(2S)-5-[3-[(2S)-2-(1-tert-butoxycarbonylpyrrolidin-3-yl)-1-oxo-isoindolin-5-yl]-2-(difluoromethyl)phenyl]-1-oxo-isoindolin-2-yl]pyrrolidine-1-carboxylate (350 mg, 42\% yield) as a solid: $^1$H NMR (400 MHz, CDCl$_3$) $\delta$ 7.93 (d, J = 8.2 Hz, 2H), 7.69--7.47 (m, 5H), 7.38 (d, J = 7.7 Hz, 2H), 6.43 (t, J = 53.7 Hz, 1H), 5.07 (p, J = 6.8 Hz, 2H), 4.49 (s, 4H), 3.87--3.29 (m, 8H), 2.29 (ddt, J = 12.6, 7.5, 3.7 Hz, 2H), 2.14 (dq, J = 12.7, 7.5 Hz, 2H), 1.48 (d, J = 6.3 Hz, 18H) ppm; ({M} + 1) = 729.

\paragraph{Step 2: Preparation of 5,5'-(2-(difluoromethyl)-1,3-phenylene)bis(2-((S)-pyrrolidin-3-yl)isoindolin-1-one) (Figure~\ref{fig:pdl1_syn_3_2})}

\begin{figure}[H]
\centering
\includegraphics[width=0.5\linewidth]{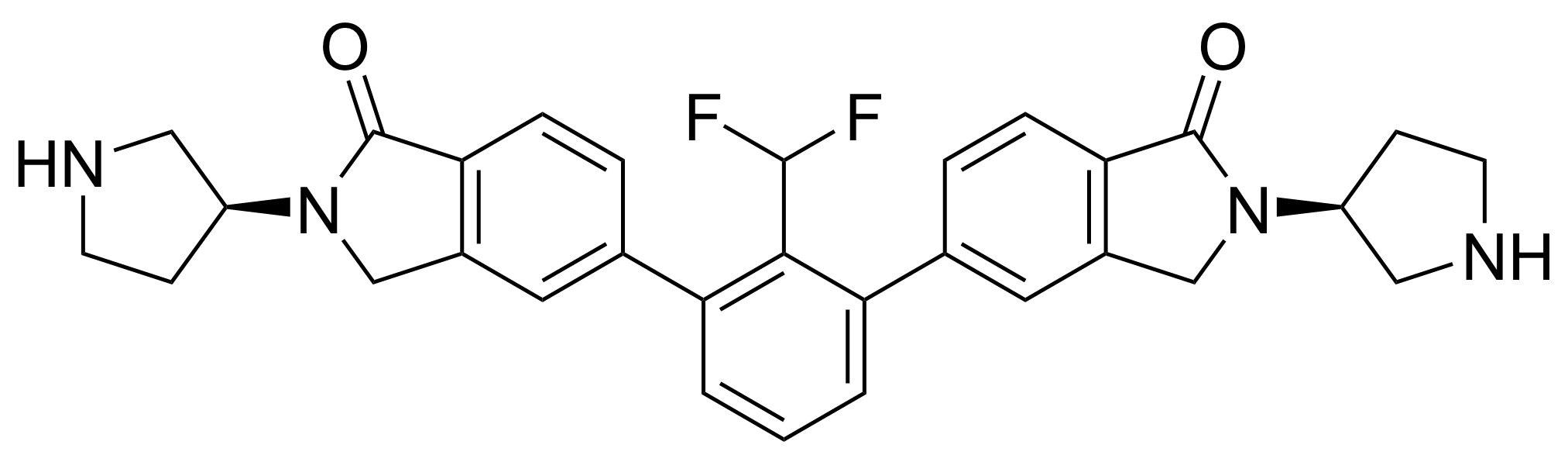}
\caption{Final compound (\cmpSeventeen) to be synthesized {in Step 2}}
\label{fig:pdl1_syn_3_2}
\end{figure}

TFA (4 mL, 51.9 mmol) was added to a stirring mixture of \iupac{tert-butyl 3-[(2S)-5-[3-[(2S)-2-(1-tert-butoxycarbonylpyrrolidin-3-yl)-1-oxo-isoindolin-5-yl]-2-(difluoromethyl)phenyl]-1-oxo-isoindolin-2-yl]pyrrolidine-1-carboxylate} (350 mg, 0.480 mmol) in dichloromethane (8 mL) at $0^{\circ} \mathrm{C}$. The resulting mixture was stirred for 2 h at room temperature. LC/MS analysis indicated the reaction was complete. Concentrated reaction mixture. The residue was partitioned between saturated NaHCO3 and dichloromethane. LC/MS analysis indicated product was in water layer. Water layer was lyophilized. Solid was suspended in 20\% methanol/dichloromethane solution and filtered to remove solid material. Filtrate was concentrated. Purified residue by preparative RP chromatography (Interchim puriFlash XS 530, RediSep Gold C18 50 g column, 15\% acetonitrile/water/0.1\% formic acid to 50\% acetonitrile/0.1\% formic acid elute) to provide 5-[2-(difluoromethyl)-3-[1-oxo-2-[(3R)-pyrrolidin-3-yl]isoindolin-5-yl]phenyl]-2-[(3R)-pyrrolidin-3-yl]isoindolin-1-one (25 mg, 10\% yield): $^1$H NMR (400 MHz, DMSO) $\delta$ 8.93 (m, 4H), 7.81 (d, J = 7.8 Hz, 2H), 7.74 (t, J = 7.7 Hz, 1H), 7.66 (s, 2H), 7.54 (dd, J = 7.8, 1.5 Hz, 2H), 7.46 (d, J = 7.7 Hz, 2H), 6.56 (t, J = 53.4 Hz, 1H), 4.89 (p, J = 7.4 Hz, 2H), 4.64 (s, 4H), 3.35 (ddd, J = 43.1, 12.1, 6.1 Hz, 6H), 2.35--2.08 (m, 4H) ppm; ({M} + 1) = 529.

\subsection{Synthesis of \CSFoneR Ligands}

\subsubsection{Synthesis of \cmpTen: 4-[4-[2-(benzylamino)-2-oxo-ethyl]-2-methyl-phenyl]benzoic acid}
\label{supp:synthesis:cmp9}

The synthesis of \cmpTen was carried out in three steps.

\paragraph{Step 1: Preparation of N-benzyl-2-(4-bromo-3-methyl-phenyl)acetamide (Figure~\ref{fig:csf1r_syn_1_1})}

\begin{figure}[H]
\centering
\includegraphics[width=0.3\linewidth]{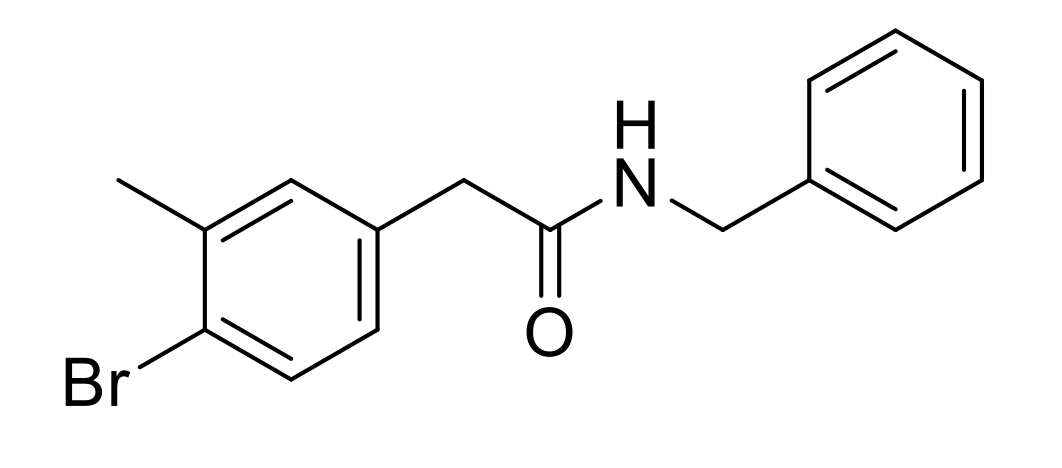}
\caption{Intermediate to be prepared in Step 1 for the synthesis of the \cmpTen}
\label{fig:csf1r_syn_1_1}
\end{figure}

To a 20 mL reaction vial equipped with a magnetic stir bar were added 
2-(4-bromo-3-methyl-phenyl)acetic acid ($0.50 \mathrm{~g}$, $2.18 \mathrm{~mmol}$), benzylamine ($0.26 \mathrm{~g}$, $2.40 \mathrm{~mmol}$), and pyridine ($0.40 \mathrm{~mL}$, $4.95 \mathrm{~mmol}$). 
The mixture was treated with 1-propanephosphonic anhydride solution in acetonitrile ($3.0 \mathrm{~mL}$, $4.14 \mathrm{~mmol}$) and was allowed to stir at room temperature. After 16 h, LC/MS analysis revealed that the reaction was complete. The crude reaction mixture was subjected to purification without prior workup. Preparative RP chromatography (Interchim puriFlash XS 530, RediSep Gold $\mathrm{C}_{18}$ $50 \mathrm{~g}$ column, 10\% acetonitrile/water/0.1\% formic acid to 100\% acetonitrile/0.1\% formic acid elute) provided two fractions. The fractions were combined and diluted with saturated potassium carbonate solution ($20 \mathrm{~mL}$)/ethyl acetate ($20 \mathrm{~mL}$). The phases were separated, and the organic phase was dried over magnesium sulfate, filtered, and concentrated to provide $N$-benzyl-2-(4-bromo-3-methyl-phenyl)acetamide ($0.15 \mathrm{~g}$, $0.47 \mathrm{~mmol}$, $22 \%$ yield) as a white solid: $^1$H NMR (400 MHz, DMSO-\emph{d}$_6$) $\delta$ 8.56 (t, J = 5.9 Hz, 1H), 7.50 (d, J = 8.1 Hz, 1H), 7.35--7.27 (m, 2H), 7.27--7.20 (m, 4H), 7.04 (dd, J = 8.1, 2.3 Hz, 1H), 4.26 (d, J = 5.9 Hz, 2H), 3.43 (s, 2H), 2.31 (s, 3H) ppm; (M + 1) = 318.

\paragraph{Step 2: Preparation of methyl 4-[4-[2-(benzylamino)-2-oxo-ethyl]-2-methyl-phenyl]benzoate (Figure~\ref{fig:csf1r_syn_1_2})}

\begin{figure}[H]
\centering
\includegraphics[width=0.45\linewidth]{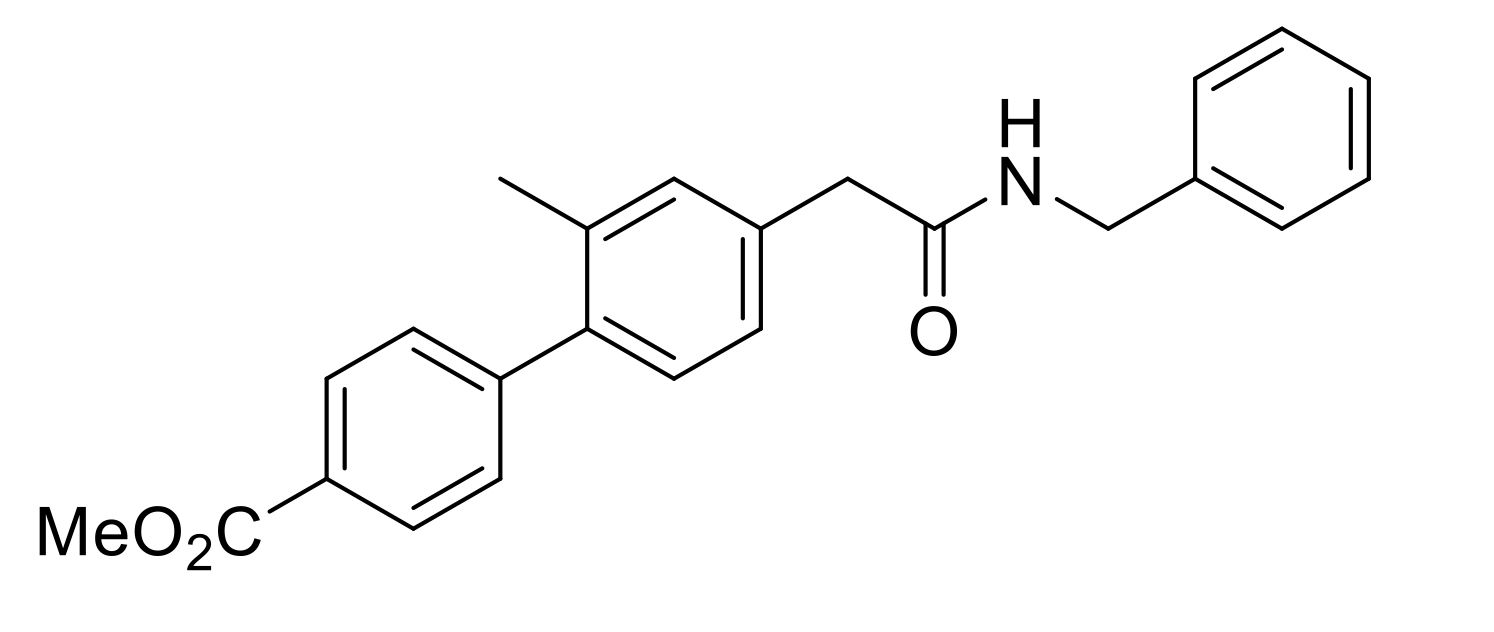}
\caption{Intermediate to be prepared in Step 2 for the synthesis of the \cmpTen}
\label{fig:csf1r_syn_1_2}
\end{figure}

To a 20 mL reaction vial equipped with a magnetic stir bar were added $N$-benzyl-2-(4-bromo-3-methyl-phenyl)acetamide ($0.15 \mathrm{~g}$, $0.47 \mathrm{~mmol}$), (4-methoxycarbonylphenyl)boronic acid ($0.11 \mathrm{~g}$, $0.59 \mathrm{~mmol}$), 
\iupac{[1,1'-bis(diphenylphosphino)ferrocene]dichloropalladium(II)}
($0.035 \mathrm{~g}$, $0.048 \mathrm{~mmol}$), sodium carbonate ($0.10 \mathrm{~g}$, $0.94 \mathrm{~mmol}$), 1,4-dioxane (10 mL), and water (2 mL).
The vessel was sealed, and the contents were degassed under vacuum/back filled with $\mathrm{N}_2(\times 3)$. The mixture was heated to $85^{\circ} \mathrm{C}$ in a heating block. After 30 min, LC/MS analysis revealed that the reaction was complete. The mixture was allowed to cool to room temperature and was diluted with water $(30 \mathrm{~mL}) /$ethyl acetate $(30 \mathrm{~mL})$. The phases were separated, and the organic phase was dried over magnesium sulfate, filtered, and concentrated to provide a brown solid. Preparative RP chromatography (Interchim puriFlash XS 530, RediSep Gold $\mathrm{C}_{18}$ $50 \mathrm{~g}$ column, 10\% acetonitrile/water/0.1\% formic acid to 100\% acetonitrile/0.1\% formic acid elute) provided one fraction. The fraction was diluted with saturated potassium carbonate solution $(20 \mathrm{~mL}) /$ethyl acetate $(20 \mathrm{~mL})$. The phases were separated, and the organic phase was dried over magnesium sulfate, filtered, and concentrated to provide a yellow oil. Further chromatographic purification (CombiFlash, $24 \mathrm{~g}$ $\mathrm{SiO}_2$ gold column, 20-70\% ethyl acetate/heptane elute, combined fractions 6-8) afforded methyl 4-[4-[2-(benzylamino)-2-oxo-ethyl]-2-methylphenyl]benzoate ($0.060 \mathrm{~g}$, $0.16 \mathrm{~mmol}$, $34 \%$ yield) as a white solid: 
$^1$H NMR (400 MHz, DMSO-\emph{d}$_6$) $\delta$ 8.60 (t, J = 5.9 Hz, 1H), 8.05--7.99 (m, 2H), 7.52--7.47 (m, 2H), 7.36--7.29 (m, 2H), 7.29--7.14 (m, 6H), 4.29 (d, J = 5.9 Hz, 2H), 3.88 (s, 3H), 3.50 (s, 2H), 2.22 (s, 3H) ppm; ({M} + 1) = 374.


\paragraph{Step 3: Preparation of 4-[4-[2-(benzylamino)-2-oxo-ethyl]-2-methyl-phenyl]benzoic acid (Figure~\ref{fig:csf1r_syn_1_3})}

\begin{figure}[H]
\centering
\includegraphics[width=0.45\linewidth]{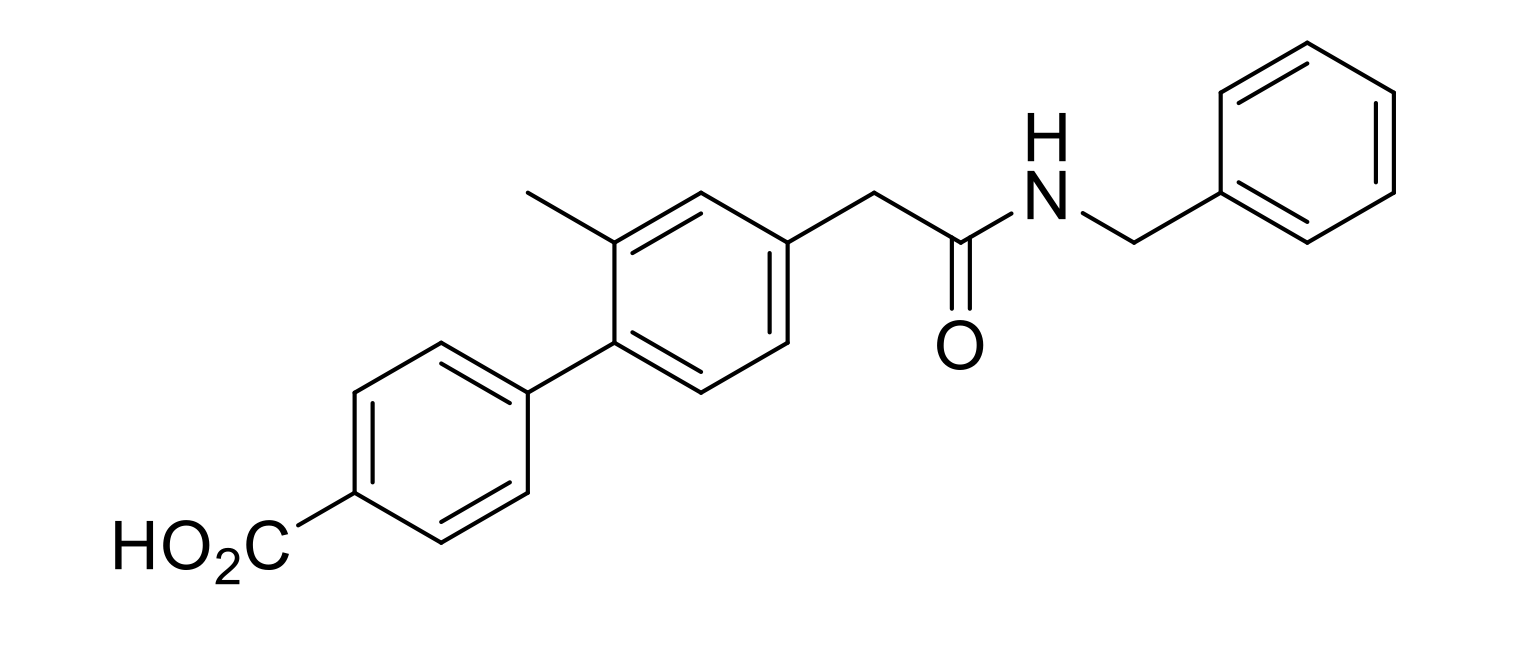}
\caption{Final compound (\cmpTen) to be synthesized in Step 3}
\label{fig:csf1r_syn_1_3}
\end{figure}

To a 100 mL recovery flask equipped with a magnetic stir bar were added methyl 4-[4-[2-(benzylamino)-2-oxo-ethyl]-2-methyl-phenyl]benzoate ($0.060 \mathrm{~g}$, $0.16 \mathrm{~mmol}$), methanol (10 mL), and water $(2 \mathrm{~mL})$. The suspension was treated with lithium hydroxide monohydrate ($0.10 \mathrm{~g}$, $2.38 \mathrm{~mmol}$) and was briefly warmed with a heat gun. The resulting homogeneous solution was allowed to stir. After 30 min, LC/MS analysis revealed that most of the starting material had been consumed. The mixture was concentrated, and the residue was partitioned between water (15 mL)/diethyl ether (15 mL). The phases were separated, and the organic phase was discarded. The aqueous phase was acidified with 1.0N hydrochloric acid solution ($\sim$$10\mathrm{~mL}$), and the resulting mixture was extracted with ethyl acetate ($20 \mathrm{~mL}$). The phases were separated, and the organic phase was dried over magnesium sulfate, filtered, and concentrated to provide 4-[4-[2-(benzylamino)-2-oxo-ethyl]-2-methylphenyl]benzoic acid ($0.042 \mathrm{~g}$, $0.12 \mathrm{~mmol}$, $73 \%$ yield) as a white solid: $^1$H NMR (400 MHz, DMSO-\emph{d}$_6$) $\delta$ 12.98 (s, 1H), 8.60 (t, J = 5.9 Hz, 1H), 8.04--7.95 (m, 2H), 7.50--7.44 (m, 2H), 7.36--7.29 (m, 2H), 7.29--7.15 (m, 6H), 4.29 (d, J = 5.9 Hz, 2H), 3.50 (s, 2H), 2.22 (s, 3H) ppm; ({M} $-$ 1) = 358.

\subsubsection{Synthesis of \cmpEleven: N-(4-methoxy-2-methylphenyl)-2-methyl-3-(pyridin-3-ylmethyl)benzamide}
\label{supp:synthesis:cmp11}

The synthesis of \cmpEleven was carried out in three steps.

\paragraph{Step 1: Preparation of methyl 2-methyl-3-(pyridin-3-ylmethyl)benzoate (Figure~\ref{fig:csf1r_syn_2_1})}

\begin{figure}[H]
\centering
\includegraphics[width=0.3\linewidth]{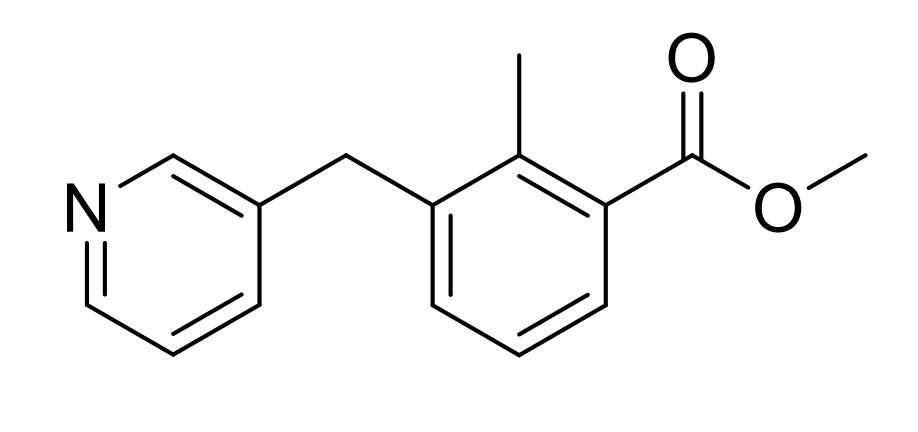}
\caption{Intermediate to be prepared in Step 1 for the synthesis of the \cmpEleven}
\label{fig:csf1r_syn_2_1}
\end{figure}

To a 100 mL recovery flask equipped with a magnetic stir bar, reflux condenser, and $\mathrm{N}_2$ inlet were added methyl 3-(bromomethyl)-2-methyl-benzoate ($0.64 \mathrm {~mg}$, $2.62 \mathrm{~mmol}$), 3-pyridylboronic acid $(0.52 \mathrm{~g}$, $4.19 \mathrm{~mmol})$, {[}1,\iupac{1'bis(diphenylphosp-hino)ferrocene{]}dichloropalladium(II)} ($0.20 \mathrm{~g}$, $0.27 \mathrm{~mmol}$), 2.0 M sodium carbonate solution ($8.0 \mathrm{~mL}$, $16.0 \mathrm{~mmol}$), and 1,4-dioxane ($25 \mathrm{~mL}$). The mixture was degassed under vacuum/back filled with $\mathrm{N}_2(\times 3)$. The mixture was heated to $90^{\circ} \mathrm{C}$ in a heating block. After 16 h, LC/MS analysis revealed that some of the desired material had formed. The mixture was allowed to cool to room temperature and was diluted with water $(50 \mathrm{~mL}) /$ethyl acetate $(50 \mathrm{~mL})$. The mixture was filtered through Celite to remove insoluble material, and the phases of the filtrate were separated. The aqueous phase was extracted with ethyl acetate $(40 \mathrm{~mL})$, and the combined organic phases were dried over magnesium sulfate, filtered, and concentrated to provide a brown oil. Preparative RP chromatography (Interchim XS 530, RediSep Gold $\mathrm{C}_{18}$ $50 \mathrm{~g}$ column, 10\% acetonitrile/water/$0.1 \%$ formic acid to $100 \%$ acetonitrile/$0.1 \%$ formic acid elute) provided four fractions. The fractions were combined and diluted with saturated potassium carbonate solution $(40 \mathrm{~mL}) /$ethyl acetate $(40 \mathrm{~mL})$. The phases were separated, and the organic phase was dried over magnesium sulfate, filtered, and concentrated to provide a brown oil. Further chromatographic purification (CombiFlash, $24 \mathrm{~g}$ $\mathrm{SiO}_2$ gold column, 30-90$\%$ ethyl acetate/heptane elute, combined fractions 1-4) afforded methyl 2-methyl-3-(3-pyridylmethyl)benzoate (0.15 g, $0.60 \mathrm{~mmol}, 23 \%$ yield) as a colorless oil: $^1$H NMR (400 MHz, CDCl$_3$) $\delta$ 8.46 (dd, J = 4.9, 1.6 Hz, 2H), 7.70 (dd, J = 7.6, 1.7 Hz, 1H), 7.36--7.33 (m, 1H), 7.28--7.25 (m, 1H), 7.23 (d, $J = 7.6$ Hz, 1H), 7.22--7.17 (m, 1H), 4.06 (s, 2H), 3.89 (s, 3H), 2.41 (s, 3H) ppm; ({M} + 1) = 242.

\paragraph{Step 2: Preparation of 2-methyl-3-(3-pyridylmethyl)benzoic acid (Figure~\ref{fig:csf1r_syn_2_2})}

\begin{figure}[H]
\centering
\includegraphics[width=0.3\linewidth]{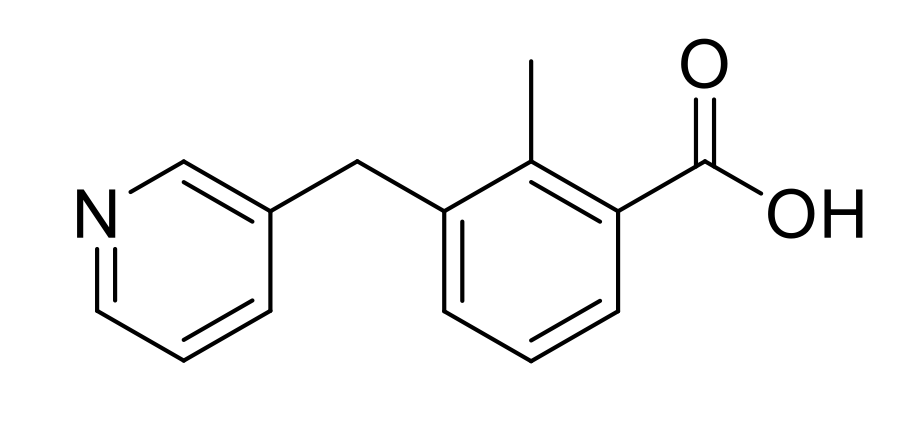}
\caption{Intermediate to be prepared in Step 2 for the synthesis of the \cmpEleven}
\label{fig:csf1r_syn_2_2}
\end{figure}

To a 100 mL recovery flask equipped with a magnetic stir bar and reflux condenser were added methyl 2-methyl-3-(3-pyridylmethyl)benzoate ($0.15 \mathrm{~g}$, $0.60 \mathrm{~mmol}$), methanol (10 mL), and water ($1 \mathrm{~mL}$). The mixture was treated with lithium hydroxide monohydrate (0.25 $\mathrm{g}$, $5.96 \mathrm{~mmol}$) and was heated to $75^{\circ} \mathrm{C}$ in a heating block. After $1 \mathrm{~h}$, $\mathrm{LC} / \mathrm{MS}$ analysis revealed that the reaction was complete. The mixture was concentrated, and the residue acidified to $\mathrm{pH} \sim 5$ with 1.0N hydrochloric acid solution $(\sim 3 \mathrm{~mL})$. The mixture was extracted with ethyl acetate $(3 \times 20 \mathrm{~mL})$. The combined organic phases were dried over magnesium sulfate, filtered, and concentrated to provide 2-methyl-3-(3-pyridylmethyl)benzoic acid ($0.95 \mathrm{~g}$, $0.42 \mathrm{~mmol}$, $70 \%$ yield) as a white solid: $^1$H NMR (400 MHz, DMSO-\emph{d}$_6$) $\delta$ 12.96 (br s, 1H), 8.52--8.33 (m, 2H), 7.58 (dd, J = 7.7, 1.5 Hz, 1H), 7.49 (dt, J = 7.9, 2.0 Hz, 1H), 7.36--7.29 (m, 2H), 7.24 (t, J = 7.7 Hz, 1H), 4.07 (s, 2H), 2.35 (s, 3H) ppm; ({M} $-$ 1) = 226.

\paragraph{Step 3: Preparation of N-(4-methoxy-2-methylphenyl)-2-methyl-3-(pyridin-3-ylmethyl)benzamide (Figure~\ref{fig:csf1r_syn_2_3})}

\begin{figure}[H]
\centering
\includegraphics[width=0.4\linewidth]{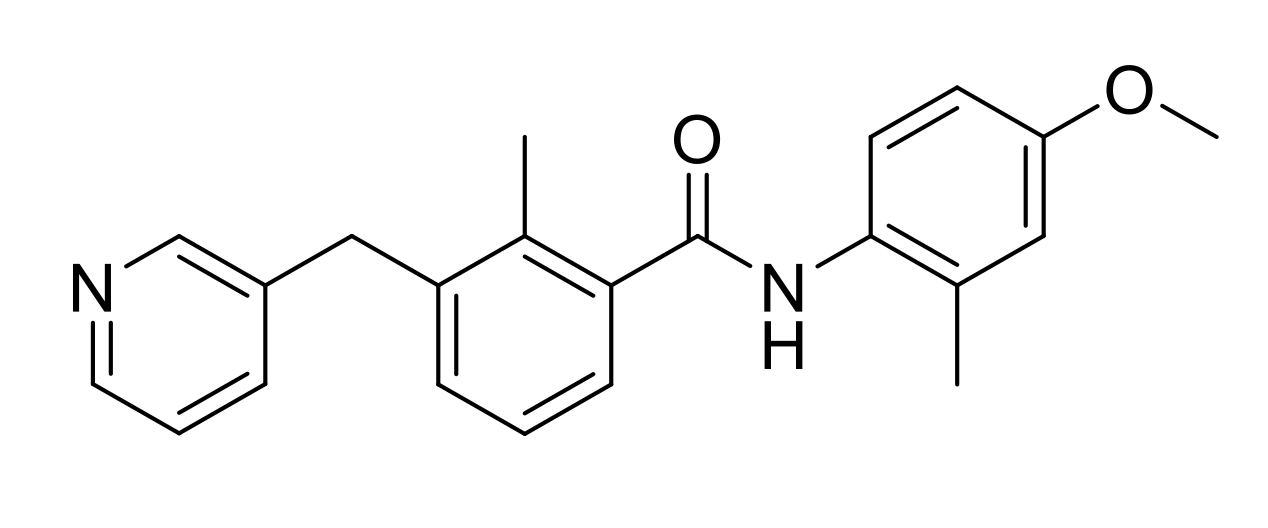}
\caption{Final compound (\cmpEleven) to be synthesized in Step 3}
\label{fig:csf1r_syn_2_3}
\end{figure}

To a 20 mL reaction vial equipped with a magnetic stir bar were added 2-methyl-3-(3pyridylmethyl)benzoic acid ($0.070 \mathrm{~g}$, $0.31 \mathrm{~mmol}$), 4-methoxy-2-methyl-aniline ($0.50 \mathrm{~g}$, $0.36$ mmol), and pyridine ($0.100 \mathrm{~mL}$, $1.24 \mathrm{~mmol}$). The mixture was treated with 1-propanephosphonic anhydride solution in acetonitrile ($1.0 \mathrm{~mL}$, $1.38 \mathrm{~mmol}$) and was allowed to stir at room temperature. After 15 min, LC/MS analysis revealed that the reaction was complete. The crude reaction mixture was subjected to purification without prior workup. Preparative RP chromatography (Interchim XS 530, RediSep Gold $\mathrm{C}_{18}$ $50 \mathrm{~g}$ column, 10\% acetonitrile/water/0.1\% formic acid to 100\% acetonitrile/0.1\% formic acid elute) provided one fraction. The fraction was diluted with saturated potassium carbonate solution $(20 \mathrm{~mL}) /$ethyl acetate ($20 \mathrm{~mL}$). The phases were separated, and the organic phase was dried over magnesium sulfate, filtered, and concentrated to provide N-(4-methoxy-2-methyl-phenyl)-2-methyl-3-(3-pyridylmethyl)benzamide ($0.025 \mathrm{~g}$, $0.072 \mathrm{~mmol}$, $23 \%$ yield) as a white solid: $^1$H NMR (400 MHz, CDCl$_3$) $\delta$ 8.49--8.44 (m, 2H), 7.70--7.64 (m, 1H), 7.42 (dt, J = 8.0, 2.2 Hz, 2H), 7.28--7.16 (m, 5H), 6.80 (d, J = 7.3 Hz, 2H), 4.05 (s, 2H), 3.80 (s, 3H), 2.39 (s, 3H), 2.29 (s, 3H) ppm; ({M} + 1) = 347.


\section{Biological Testing Assays for Case Study Targets}
\label{supp:assay}
\subsection{\CSFoneR Kinase Assay}

Recombinant human \CSFoneR inhibition was assessed with an activity assay in 8~mM MOPS pH 7.0, 0.2~mM EDTA, 250~$\mu$M KKKSPGEYVNIEFG substrate, 10~mM Mg Acetate with 33P ATP at Km. 
After incubation for 40 minutes at room temperature, the reaction was stopped by the addition of phosphoric acid to a concentration of 0.5\%. 
An aliquot of the reaction was then spotted onto a filter and washed four times in 0.425\% phosphoric acid and once in methanol prior to drying and scintillation counting.  Initial tests of compounds were conducted at 10 micromolar concentrations.  
IC$_{50}$ values were determined from dose-response curves. 

\subsection{\PDLone Binding Assay}

Surface Plasmon Resonance (SPR) experiments were performed on a Biacore 8K+ instrument using a series S SA (Streptavidin) sensor chip (Cytiva \#29104992) at $20^{\circ} \mathrm{C}$.
All analyte and ligand samples were prepared in TBS-P running buffer (20~mM Tris-HCl, pH 7.5, 150~mM NaCl, 0.02\%(v/v) surfactant P20, and 3\%(v/v) DMSO). 
Active ligand surfaces (approximately 2,000--2,500 RU) were generated by capturing biotinylated human \PDLone extracellular domain with a C-terminal Avi-tag (ACRO Biosystems \#PD1-H82E5) onto the streptavidin-coated sensor chip, while reference flow cells contained unmodified streptavidin surfaces. 
Following ligand capture, residual streptavidin binding sites were blocked with biocytin (Sigma-Aldrich \#B4261) to minimize non-specific interactions. 
Compounds were dispensed at concentrations ranging from 0.015 to 20~$\mu$M using an Echo Acoustic Liquid Handler and binding to \PDLone was assessed using a multi-cycle setup with 60 s association and 300 s dissociation phases. 
Experimental data were analyzed using Biacore Insight Evaluation Software and equilibrium dissociation constants (K$_\text{D}$) were determined by fitting to a steady-state affinity model.

\section{IC$_{50}$ Curves for Tested Molecules}
\label{supp:ic50}

\begin{figure}[htbp]

\newcommand{\curvefig}[3][0pt]{%
\begin{subfigure}[t]{0.32\textwidth}
    \centering
    \begingroup
    \setbox0=\hbox{%
        \scalebox{.85}[1.0]{%
            \includegraphics[width=.9\linewidth,angle=270]{#2}%
        }%
    }%

    \phantom{\box0}%
    \llap{%
        \raisebox{#1}[0pt][0pt]{%
            \scalebox{.85}[1.0]{%
                \includegraphics[width=.9\linewidth,angle=270]{#2}%
            }%
        }%
    }%
    \endgroup

    \vspace{-20pt}
    \caption*{\small\textbf{#3}}
\end{subfigure}%
}

\curvefig[-1.8pt]{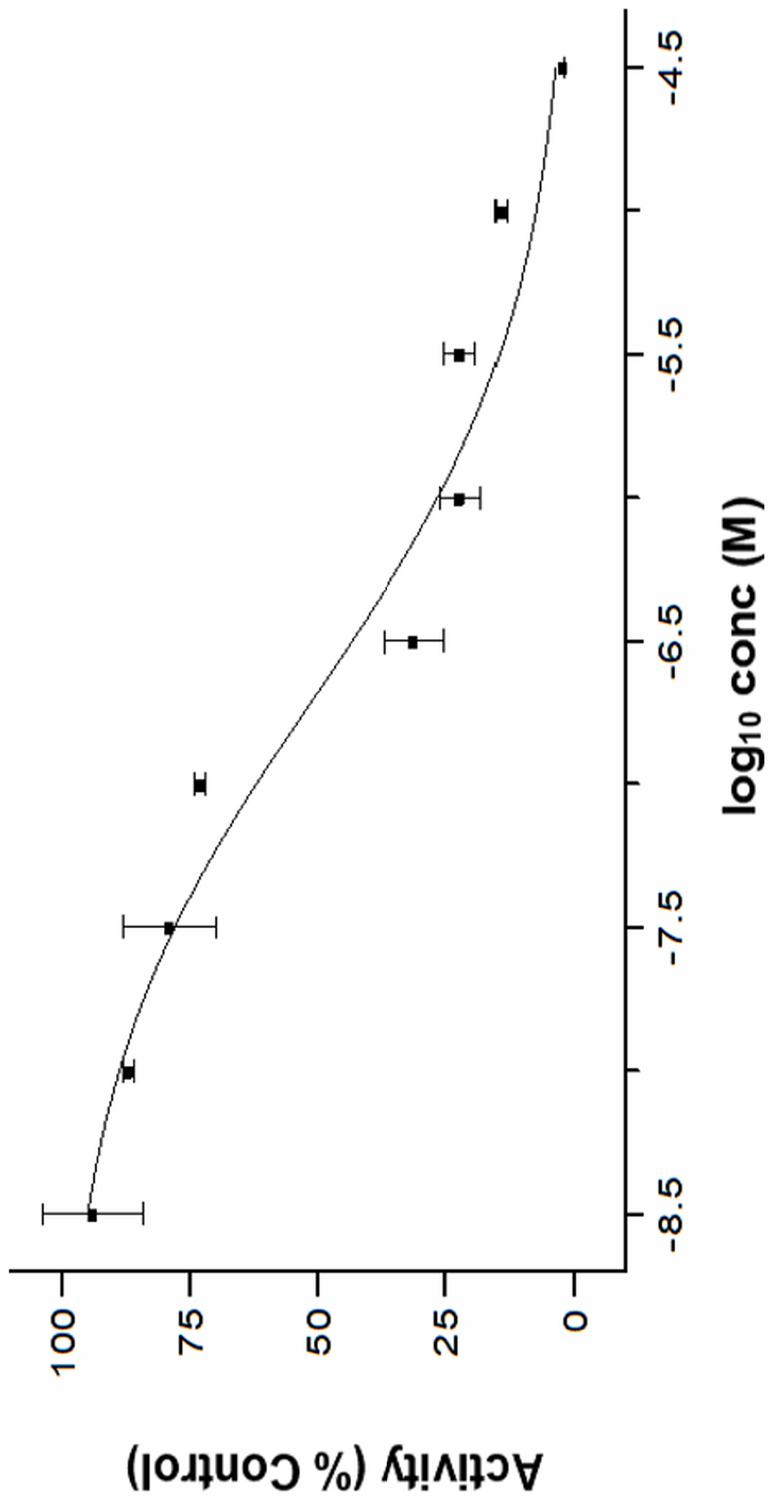}{CL-1}
\hfill
\curvefig{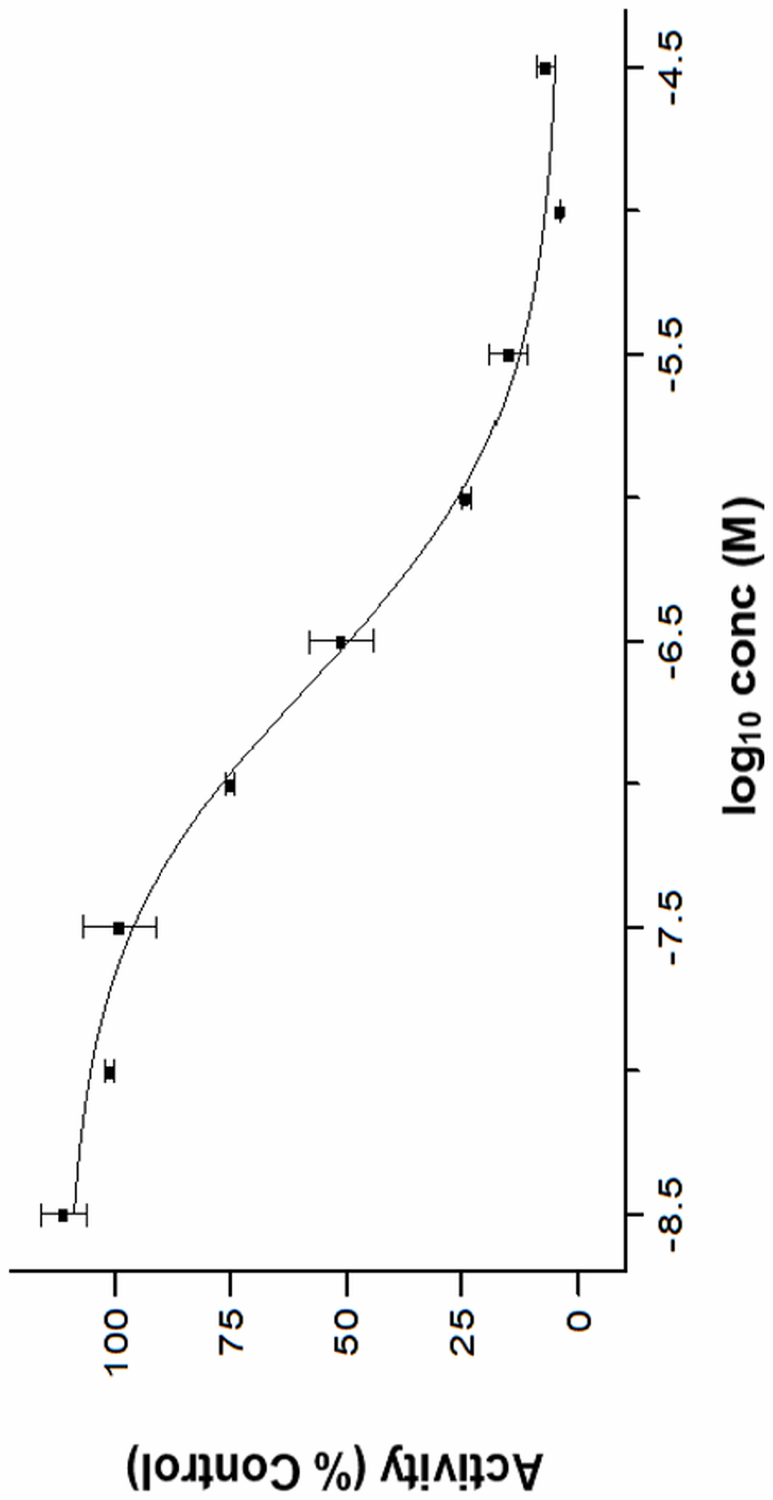}{CL-2}
\hfill
\curvefig[2.8pt]{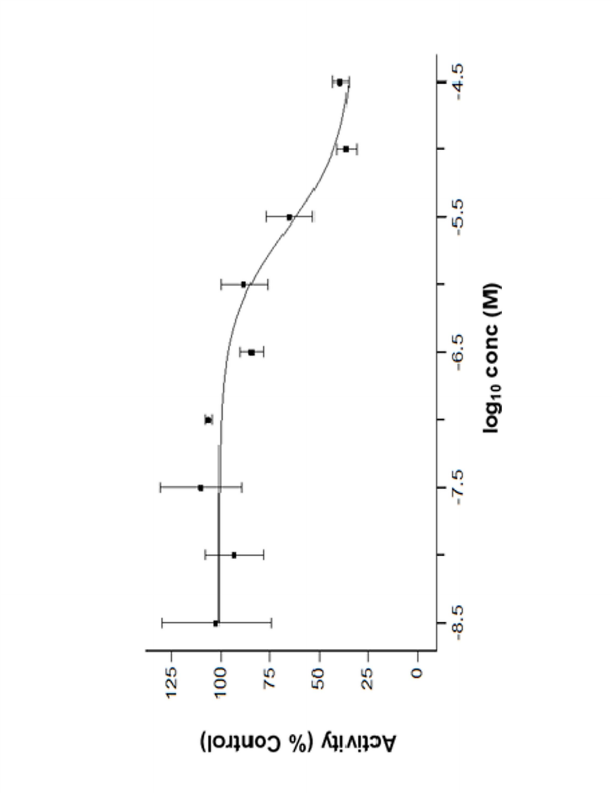}{CL-3}

\vspace{0.5em}

\curvefig[-1.8pt]{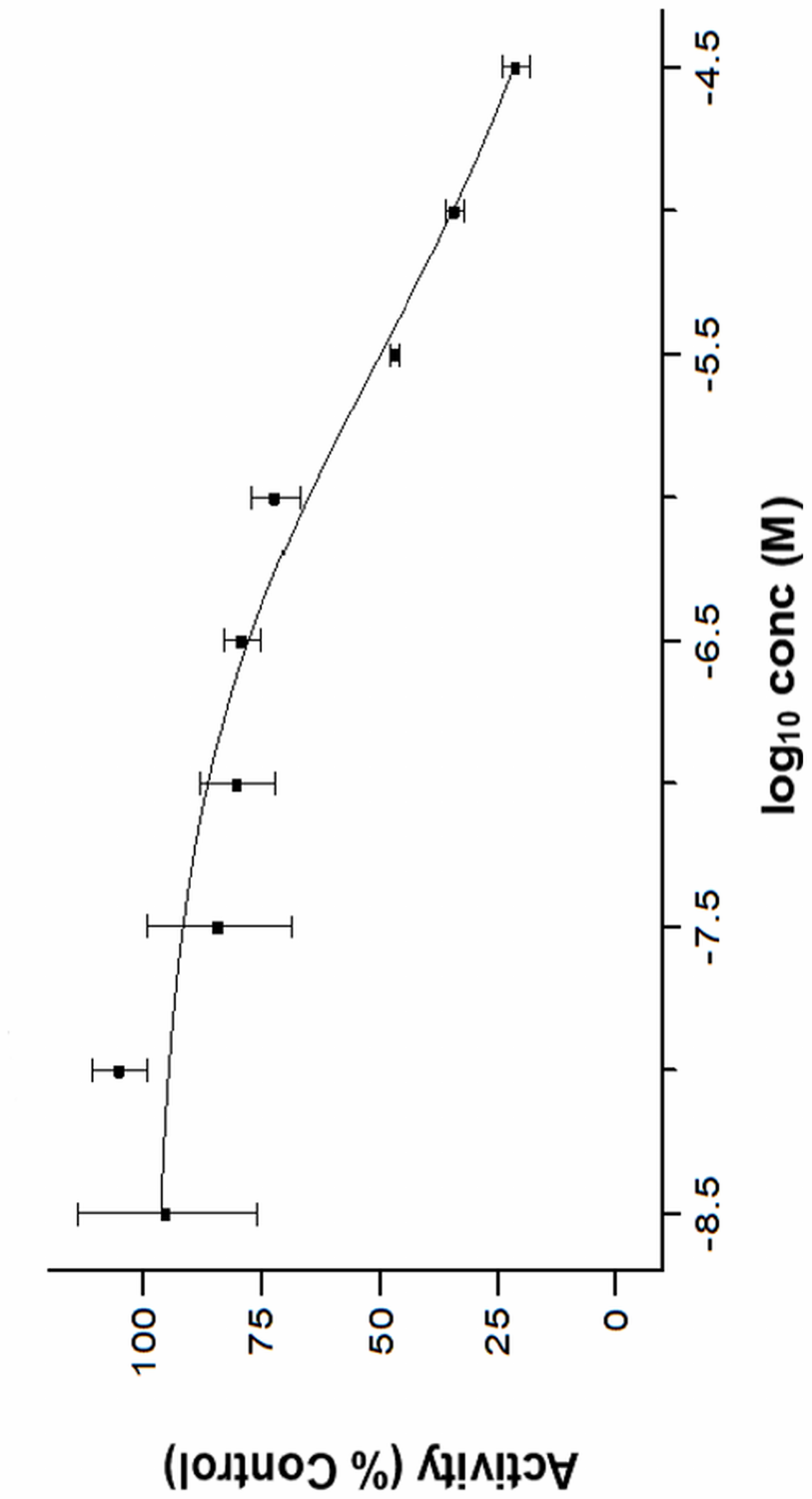}{CL-4}
\hfill
\curvefig{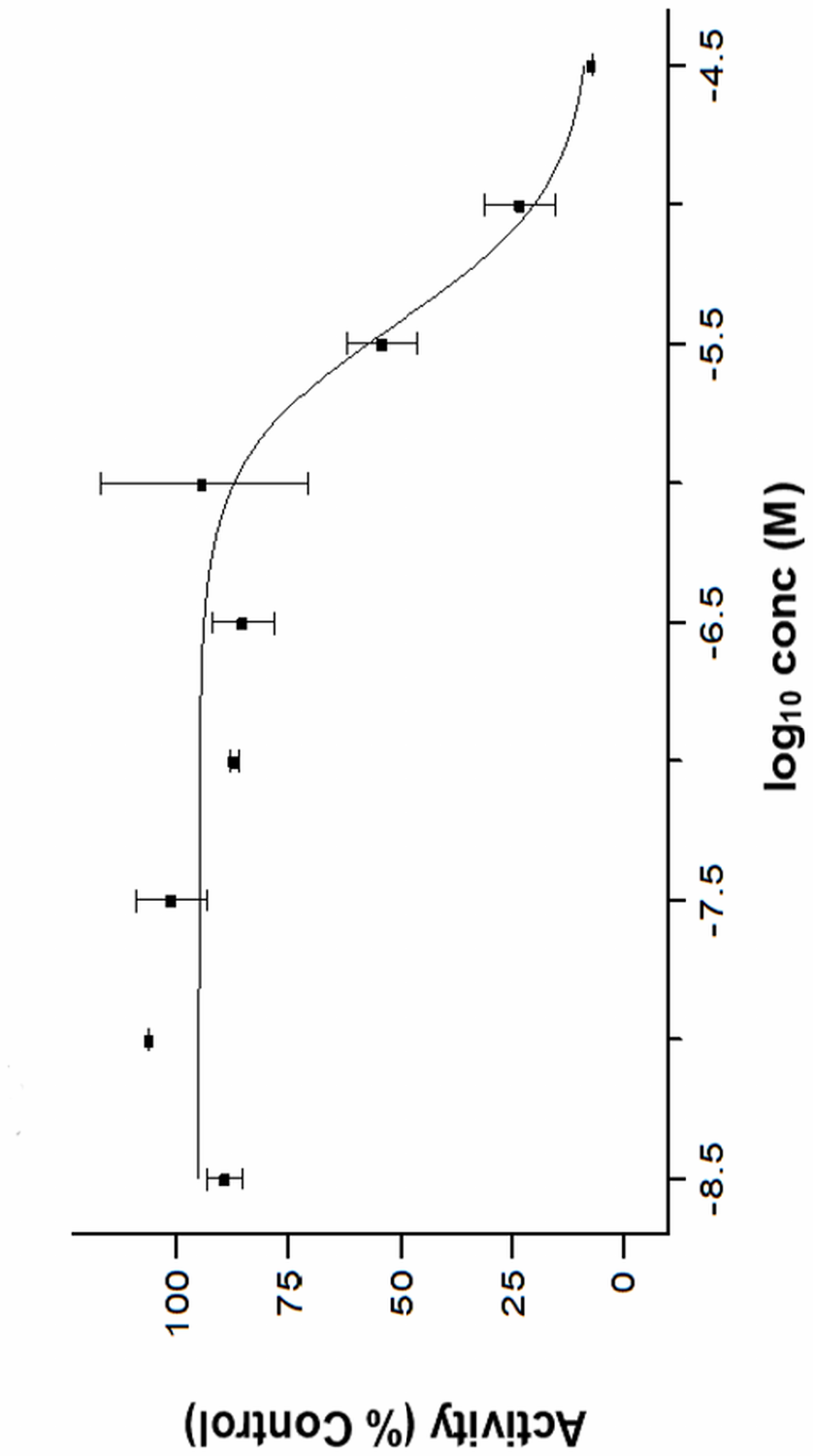}{CL-5}
\hfill
\curvefig[0.2pt]{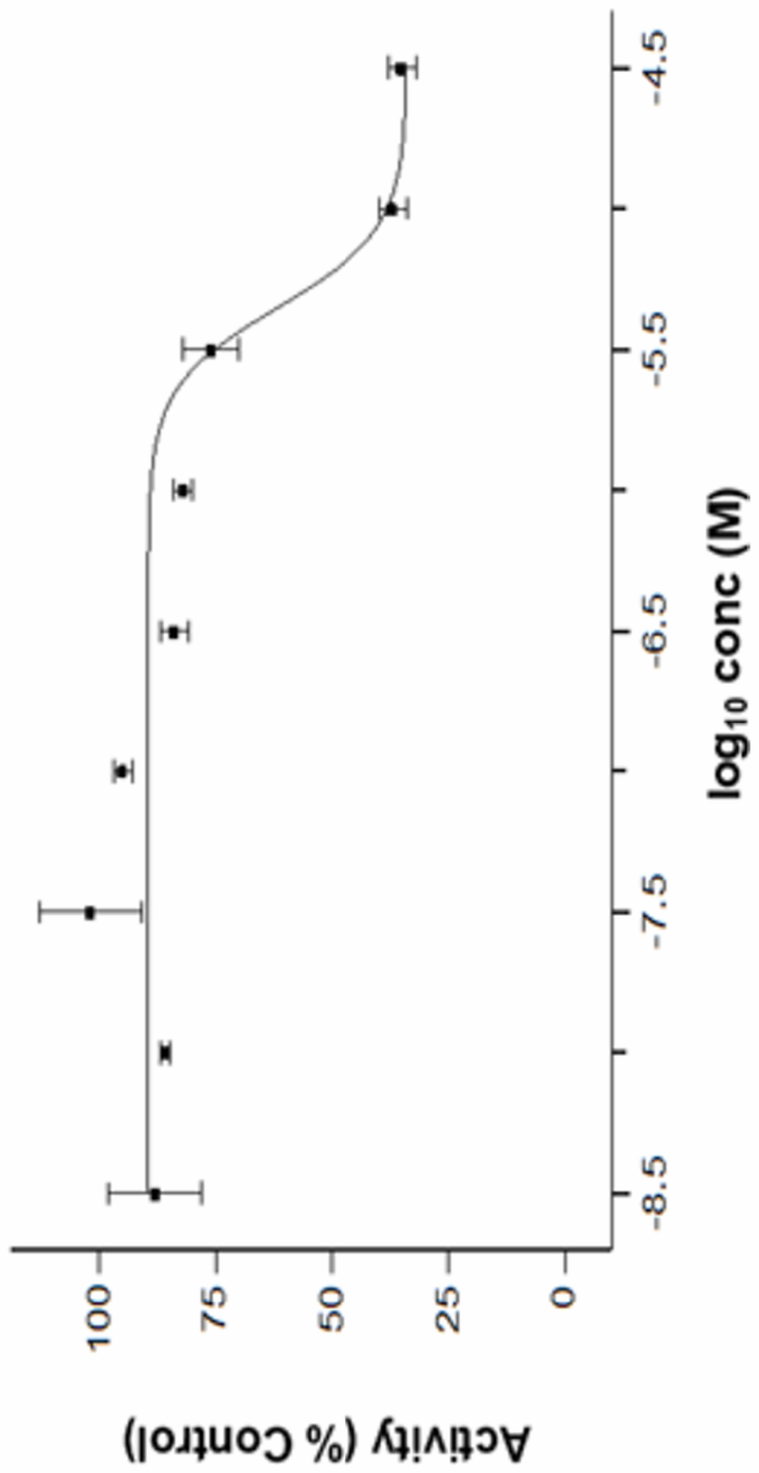}{CL-6}
\hfill
\curvefig{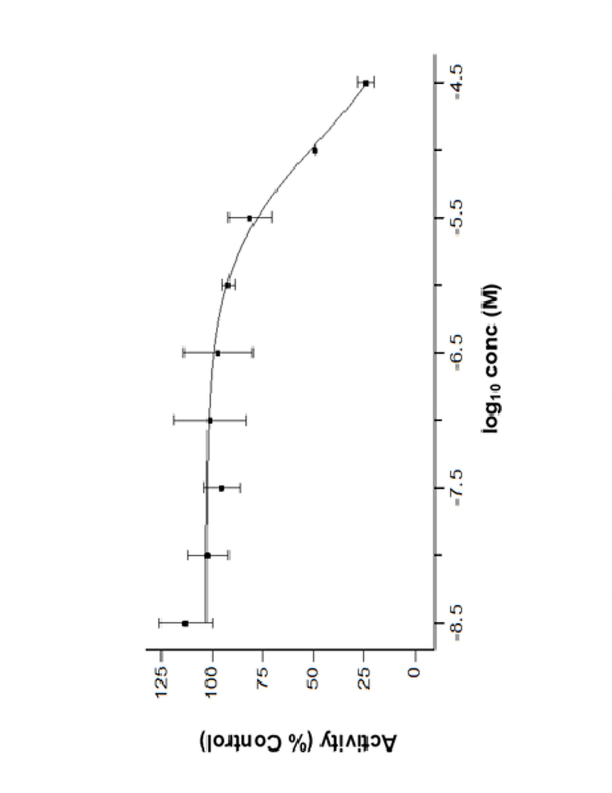}{CL-7}
\hfill
\curvefig{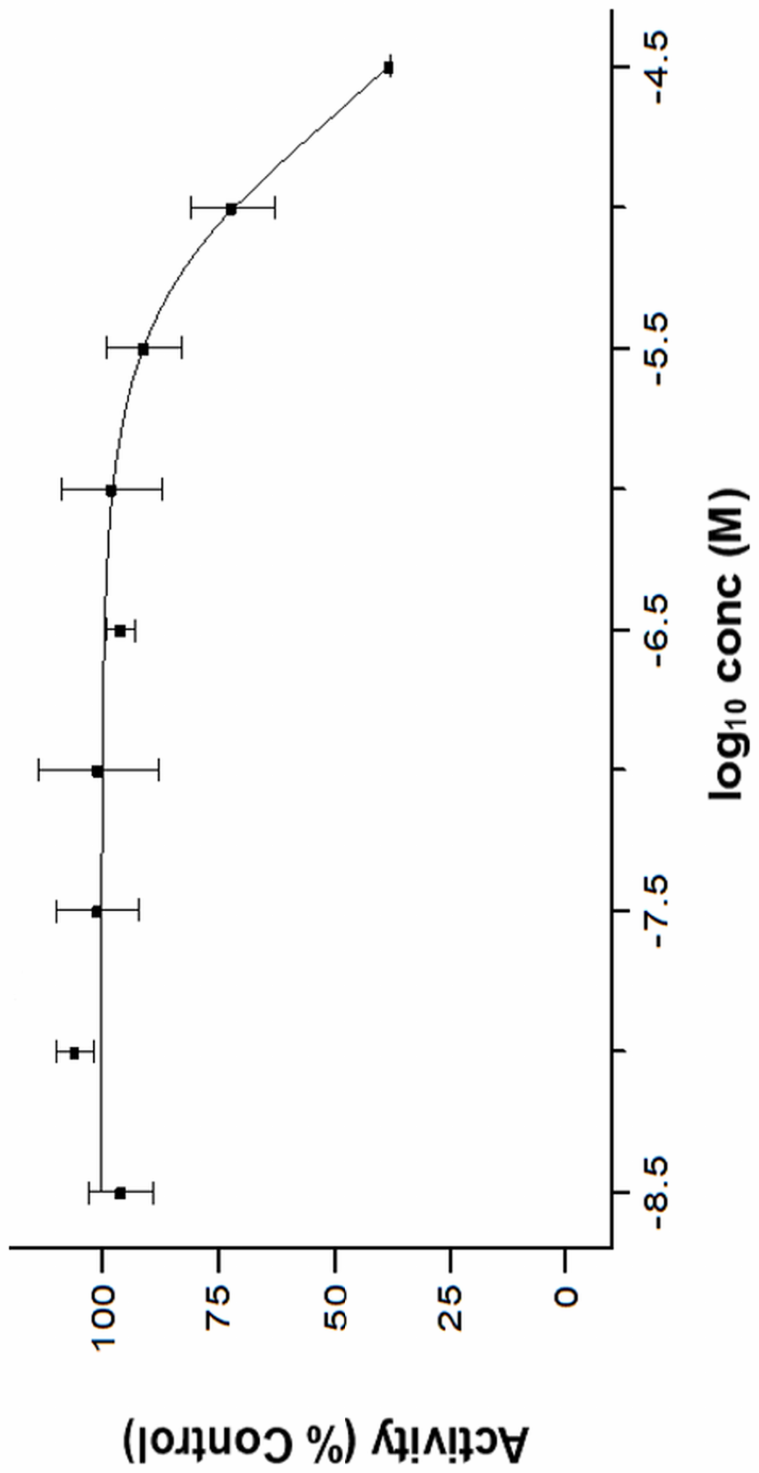}{CL-8}
\hfill
\curvefig{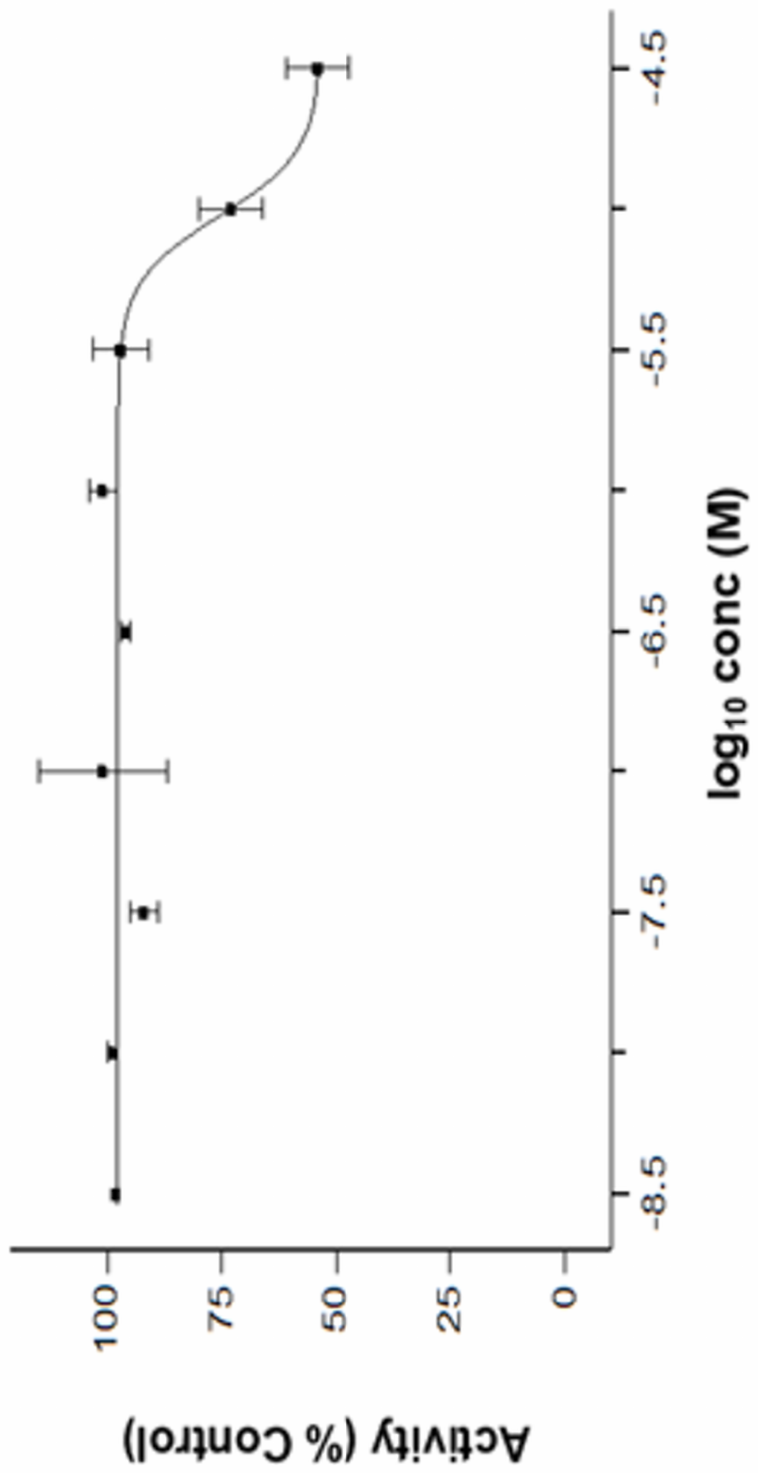}{CL-9}

\caption{
Dose--response curves for tested \CSFoneR ligands.
}
\label{fig:csf1r_ic50_curves}
\end{figure}

\end{document}